\documentclass{article}

\usepackage{PRIMEarxiv,cite}

\usepackage[utf8]{inputenc} % allow utf-8 input
\usepackage[T1]{fontenc}    % use 8-bit T1 fonts
\usepackage{url}            % simple URL typesetting
\usepackage{booktabs}       % professional-quality tables
\usepackage{amsfonts}       % blackboard math symbols
\usepackage{nicefrac}       % compact symbols for 1/2, etc.
\usepackage{microtype}      % microtypography
\usepackage{lipsum}
\usepackage{amsmath}
\usepackage{amsthm}
\usepackage{amssymb}   
\usepackage{subcaption}

\usepackage{algorithm}
\usepackage{algpseudocode}

\usepackage{siunitx }
\usepackage{bigints}
\usepackage{bm}
\usepackage{diagbox}
\usepackage{makecell}
\usepackage{multirow}
\usepackage{caption}
\usepackage{subcaption}
\usepackage{graphicx}

\usepackage{float}

\usepackage{array}

\usepackage{xcolor}
\usepackage{mathtools}
\usepackage{enumitem}

\usepackage{hyperref}
\hypersetup{
	colorlinks=true,
	citecolor=black,
	linkcolor=black,
	filecolor=black,
	urlcolor=black,
}

\usepackage{pifont}

\numberwithin{equation}{section}

\title{Separated-Variable Spectral Neural Networks: A Physics-Informed Learning Approach for High-Frequency PDEs}

\author{Xiong Xiong ${}^{1,4}$, Zhuo Zhang ${}^{5,6}$, Rongchun Hu ${}^{2,4}$\thanks{Corresponding author: rongchun\_hu@pun.edu.cn}, Chen Gao ${}^{2,4}$,Zichen Deng ${}^{1,2,3,4}$\thanks{Corresponding author: dweifan@nwpu.edu.cn}\\
	1. School of Mathematics and Statistics, 2. Department of Engineering Mechanics,\\
	3. Department of Aeronautical Engineering, 4. MIIT Key Laboratory of Dynamics and Control of Complex Systems,\\
	Northwestern Polytechnical University, Xi'an, 710072, China;\\
	5. College of Computer Science and Technology, \\
	National University of Defense Technology, Changsha 410073, Hunan, P. R. China;\\
	6. Department of Aerospace Engineering, Seoul National University, Seoul, Korea}

\begin{document}
	\maketitle
	
	\begin{abstract}
		Solving high-frequency oscillatory partial differential equations (PDEs) is a critical challenge in scientific computing, with applications in fluid mechanics, quantum mechanics, and electromagnetic wave propagation. Traditional physics-informed neural networks (PINNs) suffer from spectral bias, limiting their ability to capture high-frequency solution components.
		We introduce Separated-Variable Spectral Neural Networks (SV-SNN), a novel framework that addresses these limitations by integrating separation of variables with adaptive spectral methods. Our approach features three key innovations: (1) decomposition of multivariate functions into univariate function products, enabling independent spatial and temporal networks; (2) adaptive Fourier spectral features with learnable frequency parameters for high-frequency capture; and (3) theoretical framework based on singular value decomposition to quantify spectral bias.
		Comprehensive evaluation on benchmark problems including Heat equation, Helmholtz equation, 
		Poisson equations and Navier-Stokes equations demonstrates that SV-SNN achieves 1-3 orders of magnitude improvement 
		in accuracy while reducing parameter count by over 90\% and training time by 60×. 
		These results establish SV-SNN as an effective solution to the spectral bias problem in neural PDE solving.
		The implementation will be made publicly available upon acceptance at \url{https://github.com/xgxgnpu/SV-SNN}.
	\end{abstract}
	
	
	\section{Introduction}
	
	In recent years, deep learning technologies have attracted significant attention in scientific computing, particularly with the emergence of Physics-Informed Neural Networks (PINNs), which have established new approaches for solving partial differential equations \cite{raissi2019physics,karniadakis2021physics,cuomo2022scientific}. Unlike traditional supervised learning approaches that learn patterns from labeled data, PINNs directly solve PDEs in an unsupervised or semi-supervised manner by fully exploiting the mesh-free characteristics and function approximation capabilities of deep neural networks. This approach effectively addresses the mesh generation and computational complexity challenges encountered by traditional numerical methods in complex geometric domains and high-dimensional problems \cite{lu2021deepxde,jagtap2020extended}. PINNs have been successfully applied across various scientific and engineering domains, including fluid mechanics \cite{cai2021physics,pathak2023fourcastnet}, solid mechanics \cite{he2024multi}, and electromagnetics \cite{chen2025fe,huang2024physics}, demonstrating substantial application potential. Compared to traditional finite element and finite difference methods, PINNs offer unique advantages: they are mesh-free, differentiable, and capable of handling complex geometries \cite{haghighat2021physics,mishra2024numerical}.
	
	However, when applied to high-frequency partial differential equations—including large wavenumber Helmholtz equations, high-frequency wave equations, and equations with rapidly oscillating solutions—existing neural network methods encounter severe challenges. These equations are fundamental to numerous applications in quantum mechanics, electromagnetic wave propagation, seismic wave simulation, and acoustic engineering \cite{song2022versatile,smith2024deep,riedlinger2025physics}. The presence of high-frequency characteristics causes solution functions to exhibit complex multi-scale oscillatory patterns. Traditional numerical methods typically require extremely fine meshes to ensure accuracy, imposing the stringent requirement of "multiple grid points per wavelength," which results in exponential growth in computational cost and substantial increases in storage requirements \cite{cui2024neural,wang2024nsno}. For instance, in large wavenumber Helmholtz equations, when the wavenumber $\kappa$ exceeds $100\pi$, traditional methods require grid point numbers that grow quadratically, rendering three-dimensional problems computationally intractable \cite{sete2022helmholtz,laloy2018training}.
	
	Traditional physics-informed neural networks exhibit fundamental limitations when handling high-frequency problems, with the core issue being the inherent \textbf{spectral bias} phenomenon in neural networks. Rahaman et al. \cite{rahaman2019spectral} first systematically demonstrated that deep ReLU networks naturally prioritize learning low-frequency components while neglecting high-frequency details—a bias that proves particularly detrimental in high-frequency PDE solving. Wang et al. \cite{wang2021eigenvector} further advanced this understanding from a Neural Tangent Kernel (NTK) theory perspective, proposing the "NTK eigenvector bias" theory and elucidating the mechanism by which networks tend to learn functions along the principal eigenvector directions of their limiting NTK. Xu et al. \cite{xu2020frequency} introduced the Frequency Principle (F-Principle), revealing the intrinsic pattern of deep neural networks fitting target functions from low to high frequencies, which contrasts sharply with the high-frequency priority convergence of traditional numerical methods. A recent comprehensive review \cite{xu2025understanding} systematically summarizes the research progress in this field, identifying the spectral bias problem as the core bottleneck constraining neural network methods' application to high-frequency problems. Additionally, the rapid accumulation of numerical errors in gradient computation caused by high-frequency oscillations and the extreme complexity of loss function landscapes further exacerbate optimization difficulties \cite{wang2021understanding,krishnapriyan2021characterizing}.
	
	To address these challenges, researchers have proposed various enhancement strategies aimed at overcoming spectral bias. In feature engineering, Tancik et al. \cite{tancik2020fourier} introduced random Fourier feature methods by incorporating high-frequency encoding in input layers to enhance networks' frequency expression capabilities; Wang et al. \cite{wang2024diminishing} developed adaptive spatial encoding methods that utilize loss feedback to gradually expose frequency information of input coordinates; Wang et al. \cite{wang2021eigenvector} constructed spatiotemporal random Fourier feature architectures, achieving multi-scale problem handling through coordinate embedding layers. In network architecture innovation, Ng et al. \cite{ng2024spectrum} proposed Spectrum-Informed Multi-Stage Neural Networks (SI-MSNN) that achieve machine precision-level ($O(10^{-16})$) approximation accuracy through multi-stage residual learning, albeit with high computational cost; Liu et al. \cite{liu2024binary} designed Binary Structure Physics-Informed Neural Networks (BsPINN) that optimize network structure through sparse connections and channel splitting, demonstrating excellent performance in learning rapidly changing solutions; Huang et al. \cite{huang2025frequency} recently proposed frequency-adaptive multi-scale deep neural networks that reduce dependence on downscaling mapping parameters through adaptive parameter adjustment. In operator learning, Li et al. \cite{li2020fourier} pioneered Fourier Neural Operators (FNO), which perform convolution operations in the frequency domain to learn solution operators for parameterized PDE families. Although these methods have achieved considerable success in their respective application domains, they represent essentially \textbf{incremental improvements} to traditional PINNs and fail to fundamentally resolve the core difficulties associated with high-frequency problems.
	
	More critically, existing methods generally lack rigorous theoretical analysis tools and guiding principles. Most architectural designs rely on empirical knowledge and intuition, lacking effective analysis methods for neural network parameter spaces \cite{xu2025understanding}. Existing NTK theory primarily focuses on training dynamics but lacks comprehensive analysis of parameter structure and effective dimensions \cite{canatar2021spectral,fridovich2021spectral}. Furthermore, the absence of mathematical tools to quantify spectral bias impact results in network design and optimization lacking theoretical guidance \cite{hong2022activation,basdevant1986spectral}. This theoretical foundation weakness severely constrains further development in this field, necessitating new theoretical frameworks to guide fundamental architectural innovation.
	
	Inspired by the proven success of classical separation of variables and spectral methods in numerical analysis, 
	we contend that the key to solving high-frequency PDEs lies in fully exploiting the intrinsic mathematical structure 
	of the problems. Separation of variables reduces computational complexity by decomposing multidimensional functions 
	into products of one-dimensional functions, serving as a classical and effective approach for solving linear 
	PDEs \cite{brandstetter2023message}. Spectral methods leverage the excellent approximation properties of 
	global basis functions (such as Fourier series and Chebyshev polynomials) to achieve high-precision approximation 
	with fewer degrees of freedom \cite{trefethen2000spectral}. The core advantages of these two methods 
	directly address the weaknesses of traditional PINNs: separation of variables reduces dimensional complexity, 
	while spectral methods are inherently suited for high-frequency representation. 
	Based on this insight, we propose \textbf{Separated-Variable Spectral Neural Networks} (SV-SNN), a novel neural network framework specifically designed for high-frequency partial differential equations.
	
	Our main contributions can be summarized in five key aspects: \textbf{(1) Fundamental reconstruction of mathematical architecture}: We present the first systematic integration of separation of variables with spectral methods into neural network architecture, representing multivariate functions as linear combinations of products of univariate functions. We design both spatial spectral neural networks and spatiotemporal spectral neural networks to effectively decouple spatial and spatiotemporal complexities, achieving a fundamental breakthrough at the architectural level compared to incremental improvements of existing methods. \textbf{(2) Adaptive Fourier spectral feature representation}: We employ Fourier spectral feature networks with learnable frequency parameters to replace traditional fully-connected neural networks, enabling natural adaptation to high-frequency information representation through adaptive mechanisms and achieving truly adaptive high-frequency capture compared to fixed feature mapping and simple parameter adjustment. \textbf{(3) Three-level frequency sampling strategy}: We design hierarchical sampling mechanisms including basic frequency layer, characteristic frequency layer, and high-frequency compensation layer based on problem characteristic frequencies, surpassing existing single sampling strategies to achieve systematic coverage of the full frequency spectrum. \textbf{(4) Hybrid differentiation method}: We utilize analytical differentiation for spatial derivatives and automatic differentiation for temporal derivatives, effectively reducing numerical error accumulation in derivative computation during high-frequency calculations. \textbf{(5) SVD effective rank theoretical framework}: We introduce the concept of effective rank through singular value decomposition of Jacobian matrices for analyzing effective parameter space dimensions and spectral bias during gradient descent, while defining parameter space collapse phenomena, thereby providing new theoretical analysis tools for network architecture design and training optimization.
	
	\section{Proposed Method}
	
	\subsection{Variable Separation}
	
	Variable separation is a classical technique for solving partial differential equations. Its fundamental principle involves decomposing multivariate functions into products of univariate functions, thereby transforming complex high-dimensional problems into several relatively simple one-dimensional problems. This method offers significant advantages when solving linear partial differential equations by effectively reducing the dimensional complexity of the problems.
	
	For partial differential equations defined on high-dimensional spatiotemporal domain $\Omega_T = \Omega \times [0,T]$, where $\Omega \subset \mathbb{R}^d$ is a $d$-dimensional spatial domain, variable separation assumes the solution function can be written as a product of spatial and temporal functions:
	\begin{equation}
	u(\bm{x}, t) = X(\bm{x}) \cdot T(t)
	\end{equation}
	where $\bm{x} = (x_1, x_2, \ldots, x_d) \in \Omega$, $X(\bm{x})$ depends only on spatial variables $\bm{x}$, and $T(t)$ depends only on temporal variable $t$.
	
	In more general cases, solution functions usually cannot be represented by a single product form, but need to be expressed as linear superposition of multiple modes:
	\begin{equation}
	u(\bm{x}, t) = \sum_{n=1}^{N} c_n X_n(\bm{x}) T_n(t)
	\end{equation}
	where $\{X_n(\bm{x})\}_{n=1}^N$ are spatial mode functions, $\{T_n(t)\}_{n=1}^N$ are temporal mode functions, and $\{c_n\}_{n=1}^N$ are mode coefficients.
	
	For high-dimensional spatial problems, spatial variables can be further separated. For example, for $d$-dimensional problems:
	\begin{equation}
	u(\bm{x}, t) = \sum_{n=1}^{N} c_n \prod_{j=1}^{d} X_n^{(j)}(x_j) \cdot T_n(t)
	\end{equation}
	where $X_n^{(j)}(x_j)$ represents the component function of the $n$-th mode in the $j$-th coordinate direction.
	
	Variable separation offers significant advantages: in dimensional reduction, it decomposes $(d+1)$-dimensional spatiotemporal problems into $d$ one-dimensional spatial problems and one one-dimensional temporal problem, substantially improving computational efficiency and making each separated sub-problem relatively simple and tractable; mode decomposition provides clear physical intuition and interpretability for the solution structure. However, traditional variable separation also exhibits obvious limitations: in applicability, it is primarily restricted to linear partial differential equations and specific boundary conditions; geometrically, it typically requires regular geometric domains with stringent requirements for boundary condition forms; it encounters difficulties when handling strongly nonlinear problems, necessitating additional treatment strategies. To fully exploit the advantages of variable separation, we subsequently integrate variable separation with the mesh-free characteristics and flexibility of neural networks, designing an innovative neural network architecture that aims to preserve the advantages of variable separation while overcoming its inherent limitations.
	
	\subsection{Spectral Methods}
	
	Spectral methods are classical high-precision approaches in numerical analysis for solving partial differential equations. Their fundamental principle involves approximating solution functions using linear combinations of global basis functions. For functions $u(\bm{x})$ defined on a $d$-dimensional region $\Omega \subset \mathbb{R}^d$, spectral methods expand them as:
	\begin{equation}
	u(\bm{x}) \approx \sum_{\bm{k} \in \mathcal{K}} c_{\bm{k}} \phi_{\bm{k}}(\bm{x})
	\end{equation}
	where $\bm{k} = (k_1, k_2, \ldots, k_d)$ is a multi-dimensional index, $\{\phi_{\bm{k}}(\bm{x})\}_{\bm{k} \in \mathcal{K}}$ is a preselected basis function system (such as Fourier series, orthogonal polynomial bases including Legendre polynomials, Chebyshev polynomials, etc.), and $\{c_{\bm{k}}\}_{\bm{k} \in \mathcal{K}}$ are expansion coefficients to be determined.
	
	Here, taking traditional two-dimensional Fourier spectral methods as an example, for function $u(x, y)$ defined on rectangular domain $\Omega = [0, L_1] \times [0, L_2]$, using two-dimensional trigonometric basis expansion, expressed in complex form as:
	\begin{equation}
	u(x, y) = \sum_{m=-M}^{M} \sum_{n=-N}^{N} \hat{u}_{mn} e^{i\left(\frac{2\pi m x}{L_1} + \frac{2\pi n y}{L_2}\right)}
	\end{equation}
	where $\hat{u}_{mn}$ are complex Fourier coefficients, and base frequencies are $w_{1} = 2\pi/L_1$ and $w_{2} = 2\pi/L_2$.
	
	Although spectral methods exhibit exponential convergence for smooth solutions, their fixed basis function characteristics limit their applicability in complex geometric domains. Additionally, the number of parameters in traditional multidimensional spectral methods grows exponentially with dimension, causing severe curse of dimensionality problems.
	
	\subsection{Separated-Variable Spectral Neural Networks}
	
	Inspired by separation of variables and spectral methods, we propose separated-variable spectral neural networks that extend fixed-frequency Fourier basis functions to learnable adaptive spectral features while incorporating variable separation concepts. This approach leverages both the high-frequency representation capability of Fourier spectral methods and the mesh-free characteristics of neural networks, endowing SV-SNN with the following advantages: through learnable frequency parameters, networks can automatically discover and adapt to problem characteristic frequencies without manual presetting of fixed basis functions, making them particularly suitable for handling complex problems with multi-scale and high-frequency characteristics. Compared to traditional multidimensional spectral methods, the variable separation design reduces parameter complexity from $O(K^d)$ to $O(d \cdot K)$, substantially reducing parameter count. Furthermore, this method adheres to the mathematical structure of classical variable separation, maintaining a degree of physical interpretability. Additionally, it is independent of specific mesh structures, can handle different computational domains and complex boundary conditions, thereby overcoming traditional Fourier spectral methods' limitations to regular geometric domains while inheriting spectral methods' high-frequency representation capability, making it particularly well-suited for handling high-frequency partial differential equation problems with oscillatory characteristics.
	
\begin{figure}[htbp]
		\centering
		\includegraphics[width=1.0\textwidth]{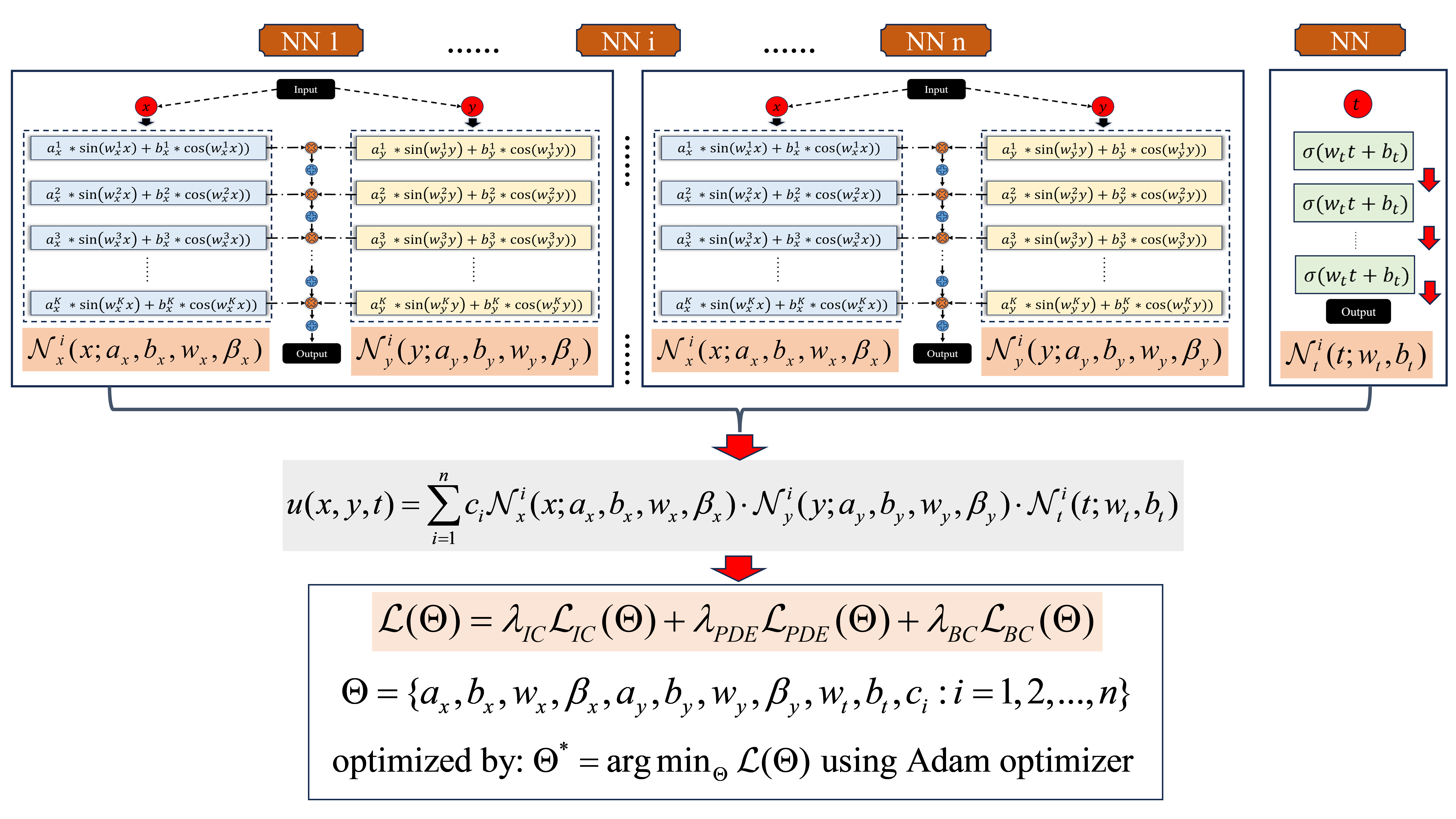}
		\caption{Separated-Variable Spectral Neural Networks (SV-SNN) architecture diagram: This architecture adopts variable separation design, representing solution functions as linear superposition of multiple modes $u(x,y,t) = \sum_{i=1}^{n} c_i \mathcal{N}_x^i \cdot \mathcal{N}_y^i \cdot \mathcal{N}_t^i$, where each spatial mode $\mathcal{N}_x^i$ and $\mathcal{N}_y^i$ uses adaptive Fourier spectral feature networks (we will design frequency initialization sampling schemes for spectral feature networks based on problem characteristic frequencies), containing learnable frequency parameters $w_x, w_y$, amplitude parameters $a_x, b_x, a_y, b_y$ and bias terms $\beta_x, \beta_y$. Temporal mode $\mathcal{N}_t^i$ uses fully-connected neural networks. Networks are trained through physics-informed loss functions, constraining initial conditions, PDE residuals, and boundary conditions respectively, using Adam optimizer for parameter updates. This design effectively combines the dimensional reduction advantages of classical variable separation with the high-frequency representation capability of spectral methods, and embeds the mesh-free characteristics and flexibility of neural networks. Analytical computation of spatial derivatives can avoid numerical error accumulation from automatic differentiation.}
		\label{fig:sv-snn}
	\end{figure}

	The proposed separated-variable spectral neural network architecture is shown in Figure~\ref{fig:sv-snn}. This network architecture combines classical variable separation ideas with adaptive Fourier spectral features, maintaining both the mathematical elegance of traditional methods and the flexibility and expressiveness of neural networks.
	
	\subsubsection{Spatial Spectral Neural Networks}
	
	For elliptic partial differential equations, we adopt a spatial spectral neural network architecture, representing solution functions as:
	\begin{equation}
	u(\bm{x}; \Theta) = \sum_{n=1}^{N} c_n \mathcal{N}^n(\bm{x}; \Theta_n^{(s)}) =\sum_{n=1}^{N} c_n \Phi_n(\bm{x}; \Theta_n^{(s)})
	\end{equation}
	where $\mathcal{N}^n(\bm{x}; \Theta_n^{(s)})$ is the neural network corresponding to the $n$-th spatial mode, specifically represented as $\Phi_n(\bm{x} ;\Theta_n^{(s)})$ using Fourier spectral features, $\Theta_n^{(s)}$ are parameters of the $n$-th spatial mode, $c_n$ are mode coefficients, and $\Theta_n^{(s)}, c_n$ are all learnable parameters in the network, with $N$ being the number of network modes.
	
	For $d$-dimensional spatial problems $\bm{x} = (x_1, x_2, \ldots, x_d) \in \Omega \subset \mathbb{R}^d$, we adopt a complete separation function product strategy, representing high-dimensional Fourier features as:
	\begin{equation}
	\Phi_n(\bm{x}; \Theta_n^{(s)}) = \prod_{j=1}^{d} \Phi_n^{(j)}(x_j; \Theta_n^{(s,j)})
	\end{equation}
	where each directional one-dimensional spectral feature is defined as:
	\begin{equation}
	\Phi_n^{(j)}(x_j; \Theta_n^{(s,j)}) = \sum_{k=1}^{K_j} [a_{n,k}^{(j)} \sin(w_{n,k}^{(j)} x_j) + b_{n,k}^{(j)} \cos(w_{n,k}^{(j)} x_j)] + \beta_n^{(j)}
	\end{equation}
	
	Here $\Theta_n^{(s,j)} = \{a_{n,k}^{(j)}, b_{n,k}^{(j)}, w_{n,k}^{(j)}, \beta_n^{(j)}\}$, containing Fourier coefficients, frequencies, and bias terms for the $n$-th spatial mode in the $j$-th coordinate direction.
	
	\subsubsection{Spatiotemporal Spectral Neural Networks}
	
	For hyperbolic or parabolic partial differential equations, we adopt a spatiotemporal spectral neural network modal decomposition architecture:
	\begin{equation}
	u(\bm{x}, t; \Theta) = \sum_{n=1}^{N} c_n \mathcal{N}^n_{\bm{x}}(\bm{x}; \Theta_n^{(s)}) \mathcal{N}^n_t(t; \Theta_n^{(t)}) = \sum_{n=1}^{N} c_n \Phi_n(\bm{x}; \Theta_n^{(s)}) T_n(t; \Theta_n^{(t)})
	\end{equation}
	where $\mathcal{N}^n_{\bm{x}}(\bm{x}; \Theta_n^{(s)})$ is the neural network corresponding to the $n$-th spatial mode, specifically represented as $\Phi_n(\bm{x}; \Theta_n^{(s)})$ using Fourier spectral features, $\Theta_n^{(s)}$ are parameters of the $n$-th spatial mode; $\mathcal{N}^n_t(t; \Theta_n^{(t)})$ is the neural network corresponding to the $n$-th temporal mode, specifically represented as $T_n(t; \Theta_n^{(t)})$, $\Theta_n^{(t)}$ are parameters of the $n$-th temporal mode; $c_n$ are mode coefficients, $\Theta_n^{(s)}$, $\Theta_n^{(t)}$, $c_n$ are all learnable parameters in the network.
	
	Each spatial mode adopts a complete separation function product structure:
	\begin{equation}
	\Phi_n(\bm{x}; \Theta_n^{(s)}) = \prod_{j=1}^{d} \Phi_n^{(j)}(x_j; \Theta_n^{(s,j)})
	\end{equation}
	
	Temporal mode networks adopt shallow fully-connected structures:
	\begin{equation}
	T_n(t; \Theta_n^{(t)}) = f_n^{(NN)}(t) = \sigma_L(W_L \sigma_{L-1}(W_{L-1} \cdots \sigma_1(W_1 t + b_1) \cdots) + b_L)
	\end{equation}
	where $\sigma_l$ is the activation function of the $l$-th layer, selected according to temporal evolution characteristics of the problem.
	
	\subsubsection{Multi-level Frequency Sampling Strategy}
	
	To better represent high-frequency characteristics of solutions, we need to design specialized frequency sampling methods based on problem characteristics. In actual partial differential equation solving, characteristic frequencies $\omega_{\text{char}}^{(j)}$ for each dimension can be defined through physical analysis. For the $j$-th dimension, they are determined through: Fourier transform of initial conditions in the $j$-th dimension to extract dominant frequencies $w_{\text{IC}}^{(j)}$, analysis of boundary condition spectral characteristics in the $j$-th dimension to obtain boundary frequencies $w_{\text{BC}}^{(j)}$, spectral decomposition of forcing terms in the $j$-th dimension to obtain forcing frequencies $w_{\text{force}}^{(j)}$, or eigenvalue analysis of differential operators in the $j$-th dimension to obtain system natural frequencies $w_{\text{sys}}^{(j)}$. Additionally, characteristic frequency coefficients $w_c$ in equations can be considered. Comprehensive analysis yields characteristic frequency for the $j$-th dimension:
	\begin{equation}
	w_{\text{char}}^{(j)} = \max\left\{w_{\text{IC}}^{(j)}, w_{\text{BC}}^{(j)}, w_{\text{force}}^{(j)}, w_{\text{sys}}^{(j)},w_{\text{c}}^{(j)}\right\}
	\end{equation}
	
	Initialization strategy for frequency parameters $\{w_k^{(j)}\}$ is also very important for network performance. We design a three-level sampling strategy based on problem characteristics. For the $j$-th dimension:
	\begin{align}
	w_k^{(j)} = \begin{cases}
	\text{Linear distribution}[w_{\min}^{(j)}, w_{\text{char}}^{(j)}], & k \in [1, K_j/4] \text{ (Basic frequencies)} \\
	\text{Gaussian distribution}\mathcal{N}(w_{\text{char}}^{(j)}, (\sigma^{(j)})^2), & k \in [K_j/4+1, 3K_j/4] \text{ (Characteristic frequencies)} \\
	\text{Uniform distribution}[w_{\text{char}}^{(j)}, w_{\max}^{(j)}], & k \in [3K_j/4+1, K_j] \text{ (High-frequency components)}
	\end{cases}
	\end{align}
	
	This three-level frequency sampling strategy can simultaneously cover multiple levels of frequencies including low, medium, and high frequencies, making SV-SNN more capable of representing different frequency characteristics of solutions, better overcoming spectral bias phenomena, and improving training efficiency and solution accuracy for high-frequency characteristics.
	
	\subsection{Hybrid Differentiation Strategy}
	
	Traditional physics-informed neural networks rely on automatic differentiation to compute derivatives, which can easily produce numerical error accumulation. One important advantage of our SV-SNN framework is the ability to analytically compute spatial partial derivatives.
	
	For $d$-dimensional Fourier features $\Phi(\bm{x}) = \prod_{j=1}^{d} \Phi^{(j)}(x_j)$, arbitrary-order mixed partial derivatives have analytical expressions. According to the Leibniz product rule:
	\begin{equation}
	\frac{\partial^{|\bm{p}|} \Phi}{\partial \bm{x}^{\bm{p}}}(\bm{x}) = \prod_{j=1}^{d} \frac{\partial^{p_j} \Phi^{(j)}}{\partial x_j^{p_j}}(x_j)
	\end{equation}
	where $\bm{p} = (p_1, p_2, \ldots, p_d)$ are partial derivative orders for each dimension, $|\bm{p}| = \sum_{j=1}^d p_j$.
	
	In spatiotemporal spectral neural network architecture, we adopt a hybrid differential computation strategy to fully leverage the advantages of Fourier spectral features. Hybrid differentiation means analytical derivatives + automatic differentiation. Specifically, spatial partial derivatives are directly computed through analytical expressions of Fourier features, achieving machine precision; temporal derivatives use automatic differentiation directly on time-related neural networks, with high computational efficiency due to small network scale. This hybrid differentiation strategy has significant advantages: first, analytical computation of spatial partial derivatives fundamentally avoids numerical error accumulation; second, analytical forms of spatial Fourier modes avoid gradient vanishing or explosion problems that automatic differentiation might encounter in high-frequency oscillatory functions, improving numerical stability, and hybrid differentiation has high computational efficiency.
	
	\section{Physics-Informed Spectral Neural Networks}
	
	\subsection{Loss Function Construction}
	
	We use SV-SNN combined with physical constraints, called Physics-Informed Spectral Neural Networks. The network's predicted solution $u^{\Theta}$ approximates the solution $u$ of partial differential equations, where $\Theta \in \mathbb{R}^p$ are network parameters. Partial differential equations can generally be expressed as:
	\begin{align}
	\mathcal{F}[u](\bm{x},t) &= 0, \quad (\bm{x},t) \in \Omega \times (0,T] \\
	\mathcal{I}[u](\bm{x},0) &= g_0(\bm{x}), \quad \bm{x} \in \Omega \\
	\mathcal{B}[u](\bm{x},t) &= g_B(\bm{x},t), \quad (\bm{x},t) \in \partial\Omega \times [0,T]
	\end{align}
	where $\mathcal{F}$, $\mathcal{I}$, and $\mathcal{B}$ represent differential operator, initial condition operator, and boundary condition operator respectively, $\Omega \subset \mathbb{R}^d$ represents spatial domain, $\partial\Omega$ represents boundary, $T$ represents time endpoint, $g_0(\bm{x})$ is given initial condition function, and $g_B(\bm{x},t)$ is given boundary condition function.
	
	The training objective of physics-informed spectral neural networks contains three weighted loss components:
	\begin{equation}
	\label{eq:total_loss}
	\mathcal{L}(\Theta) = \lambda_{\text{IC}} \mathcal{L}_{\text{IC}}(\Theta) + \lambda_{\text{PDE}} \mathcal{L}_{\text{PDE}}(\Theta) + \lambda_{\text{BC}} \mathcal{L}_{\text{BC}}(\Theta)
	\end{equation}
	
	Specific loss terms are:
	\begin{align}
	\mathcal{L}_{\text{IC}}(\Theta) &= \frac{1}{N_{\text{IC}}} \sum_{i=1}^{N_{\text{IC}}} \left|\mathcal{I}[u^{\Theta}](\bm{x}_{\text{IC}}^i, 0) - g_0(\bm{x}_{\text{IC}}^i)\right|^2 \\
	\mathcal{L}_{\text{PDE}}(\Theta) &= \frac{1}{N_{\text{PDE}}} \sum_{i=1}^{N_{\text{PDE}}} \left|\mathcal{F}[u^{\Theta}](\bm{x}_{\text{PDE}}^i, t_{\text{PDE}}^i)\right|^2 \\
	\mathcal{L}_{\text{BC}}(\Theta) &= \frac{1}{N_{\text{BC}}} \sum_{i=1}^{N_{\text{BC}}} \left|\mathcal{B}[u^{\Theta}](\bm{x}_{\text{BC}}^i, t_{\text{BC}}^i) - g_B(\bm{x}_{\text{BC}}^i, t_{\text{BC}}^i)\right|^2
	\end{align}
	
	where PDE residual loss utilizes analytical derivative computation, all spatial derivative terms are obtained through analytical expressions of Fourier features, avoiding numerical error accumulation from automatic differentiation.
	
	To completely describe the core algorithmic process of the proposed method, we present the specific training algorithm for SV-SNN 
	in Appendix~\ref{sv_snn_training}

	\subsection{Network Training Dynamics Analysis}
	
	To deeply understand the training difficulty mechanisms of PINNs, we establish a theoretical analysis framework starting from Jacobian matrices. By analyzing the singular value decay characteristics of Jacobian matrices, the intrinsic relationship between effective rank and parameter space collapse in gradient descent, and the connection between singular value decay and spectral bias, we reveal the mathematical essence of PINN training difficulties.
	
	Consider a physics-informed neural network $u^{\Theta}$ with parameters $\Theta \in \mathbb{R}^p$. For each component $\mathcal{O} \in \{\mathcal{I}, \mathcal{F}, \mathcal{B}\}$ of the loss function, define the corresponding Jacobian matrix:
	\begin{equation}
	\mathbf{J}_{\mathcal{O}} = \begin{bmatrix}
	\frac{\partial \mathcal{O}[u^{\Theta}](\bm{x}_{\mathcal{O}}^1, t_{\mathcal{O}}^1)}{\partial \theta_1} & \frac{\partial \mathcal{O}[u^{\Theta}](\bm{x}_{\mathcal{O}}^1, t_{\mathcal{O}}^1)}{\partial \theta_2} & \cdots & \frac{\partial \mathcal{O}[u^{\Theta}](\bm{x}_{\mathcal{O}}^1, t_{\mathcal{O}}^1)}{\partial \theta_p} \\
	\frac{\partial \mathcal{O}[u^{\Theta}](\bm{x}_{\mathcal{O}}^2, t_{\mathcal{O}}^2)}{\partial \theta_1} & \frac{\partial \mathcal{O}[u^{\Theta}](\bm{x}_{\mathcal{O}}^2, t_{\mathcal{O}}^2)}{\partial \theta_2} & \cdots & \frac{\partial \mathcal{O}[u^{\Theta}](\bm{x}_{\mathcal{O}}^2, t_{\mathcal{O}}^2)}{\partial \theta_p} \\
	\vdots & \vdots & \ddots & \vdots \\
	\frac{\partial \mathcal{O}[u^{\Theta}](\bm{x}_{\mathcal{O}}^{N_{\mathcal{O}}}, t_{\mathcal{O}}^{N_{\mathcal{O}}})}{\partial \theta_1} & \frac{\partial \mathcal{O}[u^{\Theta}](\bm{x}_{\mathcal{O}}^{N_{\mathcal{O}}}, t_{\mathcal{O}}^{N_{\mathcal{O}}})}{\partial \theta_2} & \cdots & \frac{\partial \mathcal{O}[u^{\Theta}](\bm{x}_{\mathcal{O}}^{N_{\mathcal{O}}}, t_{\mathcal{O}}^{N_{\mathcal{O}}})}{\partial \theta_p}
	\end{bmatrix} \in \mathbb{R}^{N_{\mathcal{O}} \times p}
	\end{equation}
	
	Perform singular value decomposition on $\mathbf{J}_{\mathcal{O}}$:
	\begin{equation}
	\mathbf{J}_{\mathcal{O}} = \mathbf{U}_{\mathcal{O}} \boldsymbol{\Sigma}_{\mathcal{O}} \mathbf{V}_{\mathcal{O}}^T
	\end{equation}
	
	where $\boldsymbol{\Sigma}_{\mathcal{O}} = \text{diag}(\sigma_1^{\mathcal{O}}, \sigma_2^{\mathcal{O}}, \ldots, \sigma_{r_{\mathcal{O}}}^{\mathcal{O}})$ contains singular values arranged in decreasing order $\sigma_1^{\mathcal{O}} \geq \sigma_2^{\mathcal{O}} \geq \ldots \geq \sigma_{r_{\mathcal{O}}}^{\mathcal{O}} > 0$, $r_{\mathcal{O}} = \text{rank}(\mathbf{J}_{\mathcal{O}})$ is the algebraic rank of the matrix.
	
	\textbf{Effective Rank:} In actual deep networks, singular values of Jacobian matrices exhibit rapid decay characteristics. To quantify this decay degree, we define effective rank based on cumulative energy:
	\begin{equation}
	r_{\mathcal{O}}^{\text{eff}}(\eta) = \min\left\{k : \frac{\sum_{i=1}^{k} \left(\sigma_i^{\mathcal{O}}\right)^2}{\sum_{i=1}^{r_{\mathcal{O}}} \left(\sigma_i^{\mathcal{O}}\right)^2} \geq \eta\right\}
	\label{eq:effective_rank}
    \end{equation}
	
	where $\eta \in (0,1)$ is the energy threshold (we take $\eta = 0.99$ in this paper). This formula essentially seeks the minimum number of principal components that can explain more than 99\% of total energy (i.e., total sum of squared singular values), thereby reflecting the effective dimensions where Jacobian matrices truly play a role.
	
	\textbf{Parameter Space Collapse:} When singular values of Jacobian matrices undergo rapid decay, meaning the matrix's effective rank is very small, the effective rank $r_{\mathcal{O}}^{\text{eff}}(\eta) \ll r_{\mathcal{O}}$ defined by equation \eqref{eq:effective_rank}. This rapid decay of singular values directly leads to drastic collapse of effective dimensions in parameter space. Specifically, during gradient update processes, only the first $r_{\mathcal{O}}^{\text{eff}}(\eta)$ principal singular value directions can produce effective parameter updates, while updates in the remaining $r_{\mathcal{O}} - r_{\mathcal{O}}^{\text{eff}}(\eta)$ directions are severely suppressed, causing networks to be unable to fully utilize their parameter space's expressive capability.
	
	The gradient of loss function $\mathcal{L}_{\mathcal{O}}(\Theta)$ is:
	\begin{equation}
	\nabla_{\Theta} \mathcal{L}_{\mathcal{O}}(\Theta) = \frac{2}{N_{\mathcal{O}}} \mathbf{J}_{\mathcal{O}}^T \boldsymbol{r}_{\mathcal{O}}
	\end{equation}
	
	where $\boldsymbol{r}_{\mathcal{O}} = [\mathcal{O}[u^{\Theta}](\bm{x}_{\mathcal{O}}^1, t_{\mathcal{O}}^1), \ldots, \mathcal{O}[u^{\Theta}](\bm{x}_{\mathcal{O}}^{N_{\mathcal{O}}}, t_{\mathcal{O}}^{N_{\mathcal{O}}})]^T$ is the residual vector.
	
	Using singular value decomposition, the gradient can be expressed as:
	\begin{equation}
	\nabla_{\Theta} \mathcal{L}_{\mathcal{O}}(\Theta) = \frac{2}{N_{\mathcal{O}}} \sum_{i=1}^{r_{\mathcal{O}}} \sigma_i^{\mathcal{O}} c_i^{\mathcal{O}} \mathbf{v}_i^{\mathcal{O}}
	\end{equation}
	
	where $c_i^{\mathcal{O}} = (\mathbf{u}_i^{\mathcal{O}})^T \boldsymbol{r}_{\mathcal{O}}$ are projection coefficients of residuals onto left singular vectors, and $\mathbf{v}_i^{\mathcal{O}}$ are right singular vectors.
	
	The parameter update equation is:
	\begin{equation}
	\Delta\Theta = -\alpha \nabla_{\Theta} \mathcal{L}_{\mathcal{O}}(\Theta) = -\frac{2\alpha}{N_{\mathcal{O}}} \sum_{i=1}^{r_{\mathcal{O}}} \sigma_i^{\mathcal{O}} c_i^{\mathcal{O}} \mathbf{v}_i^{\mathcal{O}}
	\end{equation}
	
	where $\alpha$ is the learning rate.
	
	When effective rank $r_{\mathcal{O}}^{\text{eff}}(\eta) \ll r_{\mathcal{O}}$, for singular value directions $i > r_{\mathcal{O}}^{\text{eff}}(\eta)$, since $\sigma_i^{\mathcal{O}}$ is very small, corresponding effective updates are extremely weak, gradient updates in most parameter directions are severely suppressed, network parameter utilization efficiency is greatly reduced, and network expressive capability is severely weakened.
	
	\paragraph{Spectral Bias Analysis:} Construct comprehensive Jacobian matrix $\mathbf{J} = [\mathbf{J}_{\mathcal{I}}^T, \mathbf{J}_{\mathcal{F}}^T, \mathbf{J}_{\mathcal{B}}^T]^T \in \mathbb{R}^{N \times p}$, where $N = N_{\mathcal{I}} + N_{\mathcal{F}} + N_{\mathcal{B}}$. The neural tangent kernel matrix is defined as:
	\begin{equation}
	\mathbf{K} = \mathbf{J}\mathbf{J}^T = \begin{bmatrix}
	\mathbf{K}_{\mathcal{I}\mathcal{I}} & \mathbf{K}_{\mathcal{I}\mathcal{F}} & \mathbf{K}_{\mathcal{I}\mathcal{B}} \\
	\mathbf{K}_{\mathcal{F}\mathcal{I}} & \mathbf{K}_{\mathcal{F}\mathcal{F}} & \mathbf{K}_{\mathcal{F}\mathcal{B}} \\
	\mathbf{K}_{\mathcal{B}\mathcal{I}} & \mathbf{K}_{\mathcal{B}\mathcal{F}} & \mathbf{K}_{\mathcal{B}\mathcal{B}}
	\end{bmatrix}
	\end{equation}
	
	There exists a key mathematical relationship between Jacobian matrices and neural tangent kernel matrices. Eigenvalues of diagonal blocks $\mathbf{K}_{\mathcal{O}\mathcal{O}} = \mathbf{J}_{\mathcal{O}} \mathbf{J}_{\mathcal{O}}^T$ and singular values of Jacobian matrices satisfy:
	\begin{equation}
	\lambda_i^{\mathcal{O}} = \left(\sigma_i^{\mathcal{O}}\right)^2, \quad i = 1, 2, \ldots, r_{\mathcal{O}}
	\end{equation}
	
	This relationship reveals that transformation from Jacobian matrices to neural tangent kernel matrices causes quadratic amplification of condition numbers:
	\begin{equation}
	\kappa(\mathbf{K}_{\mathcal{O}\mathcal{O}}) = \frac{\lambda_1^{\mathcal{O}}}{\lambda_{r_{\mathcal{O}}}^{\mathcal{O}}} = \frac{\left(\sigma_1^{\mathcal{O}}\right)^2}{\left(\sigma_{r_{\mathcal{O}}}^{\mathcal{O}}\right)^2} = \left[\kappa(\mathbf{J}_{\mathcal{O}})\right]^2
	\end{equation}
	
	This quadratic amplification effect greatly deteriorates numerical stability, which we consider as one of the key mechanisms for PINN training difficulties.
	
	Under neural tangent kernel approximation, training dynamics follow linear differential equations:
	\begin{equation}
	\frac{d\boldsymbol{r}(t)}{dt} = -\mathbf{K}(0) \boldsymbol{r}(t)
	\end{equation}
	
	where $\boldsymbol{r}(t) = [\boldsymbol{r}_{\mathcal{I}}(t)^T, \boldsymbol{r}_{\mathcal{F}}(t)^T, \boldsymbol{r}_{\mathcal{B}}(t)^T]^T$ is the residual vector.
	
	Performing spectral decomposition on neural tangent kernel matrix $\mathbf{K}(0) = \mathbf{Q}\boldsymbol{\Lambda}\mathbf{Q}^T$, where $\boldsymbol{\Lambda} = \text{diag}(\lambda_1, \lambda_2, \ldots, \lambda_N)$ contains eigenvalues $\lambda_1 \geq \lambda_2 \geq \ldots \geq \lambda_N \geq 0$.
	
	In the eigenbasis, training dynamics can be decoupled as:
	\begin{equation}
	\frac{d\tilde{r}_i(t)}{dt} = -\lambda_i \tilde{r}_i(t)
	\end{equation}
	
	where $\tilde{\boldsymbol{r}}(t) = \mathbf{Q}^T\boldsymbol{r}(t)$ is the residual representation in eigenspace. The solution is:
	\begin{equation}
	\tilde{r}_i(t) = \tilde{r}_i(0) e^{-\lambda_i t}
	\end{equation}
	
	This analytical form reveals the profound connection between singular value characteristics of Jacobian matrices and spectral bias phenomena. The singular value distribution of Jacobian matrix $\mathbf{J}_{\mathcal{O}}$ directly determines the eigenvalue distribution of neural tangent kernel matrix. When singular values of Jacobian matrices decay rapidly in high-frequency directions, i.e., $\sigma_{\text{high}}^{\mathcal{O}} \ll \sigma_{\text{low}}^{\mathcal{O}}$, corresponding neural tangent kernel eigenvalues decay even more dramatically, satisfying $\frac{\lambda_{\text{high}}}{\lambda_{\text{low}}} = \frac{(\sigma_{\text{high}}^{\mathcal{O}})^2}{(\sigma_{\text{low}}^{\mathcal{O}})^2} = \left(\frac{\sigma_{\text{high}}^{\mathcal{O}}}{\sigma_{\text{low}}^{\mathcal{O}}}\right)^2 \ll 1$. This quadratic amplified singular value decay propagation chain directly leads to frequency-dependent learning rate differences. Existing research shows that high-frequency features generally correspond to small eigenvalues of neural tangent kernel matrices, i.e., corresponding to small singular values of Jacobian matrices, making convergence speed of high-frequency error components extremely slow. Their convergence rate $\propto \lambda_{\text{high}} = (\sigma_{\text{high}}^{\mathcal{O}})^2 \ll (\sigma_{\text{low}}^{\mathcal{O}})^2 = \lambda_{\text{low}}$ is much smaller than low-frequency convergence rate.
	
	For Jacobian matrix $\mathbf{J}_{\mathcal{O}}$, we speculate that the physical mechanism of singular value decay in high-frequency directions may originate from inherent spectral bias of neural networks (standard fully-connected networks naturally bias toward learning low-frequency functions), numerical error accumulation in automatic differentiation (errors in high-order derivative computation weaken high-frequency signals), and expressive capability limitations of activation function smoothness (limiting high-frequency mode representation). These factors jointly cause singular values of $\mathbf{J}_{\mathcal{O}}$ to decay sharply in high-frequency directions, forming vicious cycles of numerical condition number deterioration. Combined with singular value decomposition of Jacobian matrices and neural tangent kernel theory, spectral bias phenomena can be characterized as: $\frac{\text{High-frequency error decay time}}{\text{Low-frequency error decay time}} = \frac{1/\lambda_{\text{high}}}{1/\lambda_{\text{low}}} = \left[\frac{\sigma_{\text{low}}^{\mathcal{O}}}{\sigma_{\text{high}}^{\mathcal{O}}}\right]^2$. This indicates that to alleviate spectral bias problems, the key lies in improving singular value distribution of Jacobian matrices, particularly preventing excessive decay of singular values in high-frequency directions.
	
	In summary, singular value decomposition analysis of Jacobian matrices reveals two fundamental mechanisms of PINN training difficulties: (1) Rapid decay of singular values leads to insufficient effective rank, causing parameter space collapse, limiting effective directions of gradient descent and parameter space utilization efficiency; (2) Singular value decay in high-frequency directions directly leads to spectral bias phenomena through quadratic amplification effects, making high-frequency error convergence extremely slow. These theoretical insights provide important guidance for designing new and more effective physics-informed neural network architectures.
	
	\begin{table}[H]
	\centering
	\caption{Comparison of spectral methods, PINN, and separated-variable spectral neural networks}
	\label{tab:spectral_comparison}
	\resizebox{1.0\textwidth}{!}{%
	\begin{tabular}{lccc}
	\toprule
	\textbf{Feature} & \textbf{Spectral Methods} & \textbf{PINN} & \textbf{SV-SNN} \\
	\midrule
	\textbf{Basis Function Type} & Fixed orthogonal basis functions & Deep neural networks & Learnable Fourier spectral features \\
	& (Fourier, Chebyshev) & $u^{\Theta}(\boldsymbol{x},t)$ & $X^{(F)}(x) = \sum_k [a_k\sin(w_k x) + b_k\cos(w_k x)]$ \\
	\midrule
	\textbf{Frequency Parameters} & Fixed frequencies & Implicit frequency learning & Adaptive frequencies\\
	& $w_k = k \cdot w_0$ & Through weights indirectly & Learnable frequencies $\omega_k$ \\
	\midrule
	\textbf{Modal Architecture} & Fixed basis function linear combination & Linear and nonlinear transformations & Multi-modal network cumulative summation \\
	& $u = \sum_k c_k \phi_k(x)$ & $u^{\Theta}(\boldsymbol{x},t; \Theta)$ & $u^{\Theta} = \sum_n c_n X_n^{(F)}(x) T_n^{(N)}(t)$ \\
	\midrule
	\textbf{Computational Domain} & Fourier space & Physical space & Physical space \\
	& Transform-based solution & Direct spatial-temporal computation & Direct spatial-temporal computation \\
	\midrule
	\textbf{Geometric Adaptability} & Regular domains & Arbitrary complex domains & Arbitrary complex domains \\
	& Requires coordinate transformation & Mesh-free collocation method & Mesh-free collocation method \\
	\midrule
	\textbf{Time Evolution} & Explicit/implicit time stepping & Continuous spatiotemporal coupling & Continuous spatiotemporal decoupling \\
	& Requires time discretization & $u^{\Theta}(\boldsymbol{x},t)$ unified modeling & Spatiotemporal separation $u^{\Theta}(\boldsymbol{x},t) = \sum c_n X_n^{(F)}(\boldsymbol{x})T_n^{(N)}(t)$ \\
	\midrule
	\textbf{Derivative Computation} & Analytical derivatives & Automatic differentiation & Analytical derivatives + automatic differentiation \\
	& Spectral differentiation matrices & Numerical error accumulation & Hybrid differentiation strategy \\
	\midrule
	\textbf{High-frequency Expression} & Fourier mode high-frequency representation & Serious spectral bias & Adaptive high-frequency representation \\
	& Fixed frequency limitations & Biased toward low-frequency learning & Characteristic frequency sampling \\
	\bottomrule
	\end{tabular}}
	\end{table}
	
	\section{Experimental Results}
	
    	This section validates the effectiveness of SV-SNN through a series of representative benchmark problems. The experimental design encompasses the main types of partial differential equations and challenging scenarios:
    (1) One-dimensional heat conduction equations ($\kappa = 20\pi, 100\pi, 500\pi$) to verify the spatiotemporal separation architecture's capability in handling high-frequency spatiotemporal dynamics;
    (2) Two-dimensional nonlinear elliptic equations to test nonlinear term handling capability;
    (3) Poisson equations to evaluate complex boundary adaptability;
    (4) High-frequency Poisson equations ($\mu = 15$) to examine high-frequency oscillation capture capability;
    (5) Taylor-Green vortex and double-cylinder Navier-Stokes equations to verify solving performance for complex strongly nonlinear fluid mechanics problems and complex boundary conditions.
    The experiments comprehensively evaluate SV-SNN's advantages over traditional PINNs across three dimensions: parameter efficiency, solution accuracy, and convergence speed.

	\begin{table}[htbp]
		\centering
		\caption{SV-SNN performance summary across various test cases: We conducted 10 experiments under different random seeds and present the average results}
		\label{tab:svsnn_performance_summary}
		\resizebox{0.7\textwidth}{!}{%
		\begin{tabular}{lccc}
		\toprule
		\textbf{Case} & \textbf{Epochs} & \textbf{ReL2E} & \textbf{Training Time/s} \\
		\midrule
		Heat Conduction Eq. ($\kappa = 20\pi$) & 5,000 & 3.81$\times$10$^{-4}$ & 146.33 \\
		Heat Conduction Eq. ($\kappa = 100\pi$) & 5,000 & 6.87$\times$10$^{-3}$ & 148.29 \\
		Heat Conduction Eq. ($\kappa = 500\pi$) & 5,000 & 3.92$\times$10$^{-2}$ & 147.54 \\
		Helmholtz Eq. ($\kappa = 24\pi$) & 5,000 & 1.62$\times$10$^{-2}$ & 205.76 \\
		Helmholtz Eq. (Single Cylinder) & 5,000 & 2.71$\times$10$^{-2}$ & 224.68 \\
		Helmholtz Eq. ($\kappa = 48\pi$) & 5,000 & 4.49$\times$10$^{-2}$ & 210.38 \\
		Nonlinear Elliptic Eq. & 5,000 & 4.25$\times$10$^{-3}$ & 120.84 \\
		Poisson Eq. (Complex Boundary) & 5,000 & 4.28$\times$10$^{-2}$ & 132.61 \\
		Poisson Eq. (Complex Source) & 40,000 & 4.89$\times$10$^{-3}$ & 498.78 \\
		Taylor-Green Vortex & 5,000 & 3.56$\times$10$^{-3}$ & 591.23 \\
		Steady Navier-Stokes (Double Cylinder) & 15,000 & 5.78$\times$10$^{-4}$ & 610.09 \\
		\bottomrule
		\end{tabular}}
	\end{table}

	To effectively evaluate algorithm performance, we establish a performance evaluation system: For parameter efficiency, we adopt the total number of trainable parameters as a key indicator for measuring model complexity; for solution accuracy, we use relative $l_2$ error
	\begin{equation}
	\text{ReL2E} = \frac{\|u^{\Theta} - u_{\text{exact}}\|_{L^2}}{\|u_{\text{exact}}\|_{L^2}}
	\end{equation}
	and maximum absolute pointwise error
	\begin{equation}
	\text{MAPE} = \max_{x \in \Omega} |u^{\Theta}(x) - u_{\text{exact}}(x)|
	\end{equation}
	to quantify deviations between predicted solutions and exact solutions; for convergence speed, we can evaluate algorithm convergence speed and training stability by analyzing training loss and test error, as well as dynamic changes in loss and test error during training, where training loss function comprehensively considers weighted combinations of PDE residual loss, boundary condition loss, and initial condition loss. Additionally, singular value decay curves and effective rank $r_{\mathcal{O}}^{\text{eff}}$ of Jacobian matrices can help us evaluate whether algorithms experience parameter space collapse and spectral bias problems during training, and help analyze algorithm convergence speed.
	
	In each test case, we set loss weight coefficients as: $\lambda_{\mathcal{I}} = \lambda_{\mathcal{B}} = \lambda_{\mathcal{F}} = 1$. During training, we adopt the Adam optimizer with initial learning rate set to $\alpha = 1 \times 10^{-3}$, using cosine annealing learning rate strategy with decay factor 0.99, decaying every 500 training epochs. All experiments are conducted on a single NVIDIA RTX 4090 GPU, with experimental code based on the PyTorch framework.

	\subsection{Heat Conduction Equations}
	
	\subsubsection{$\kappa$ = 20$\pi$}
	
	We first select the one-dimensional heat conduction equation with initial condition frequency $\kappa=20\pi$, defined on spatiotemporal region $\Omega_T = [-1,1] \times [0,1]$, with governing equation:
	\begin{equation}
	\frac{\partial u}{\partial t} - \alpha \frac{\partial^2 u}{\partial x^2} = 0
	\end{equation}
	where diffusion coefficient $\alpha = \frac{1}{(20\pi)^2} \approx 2.53 \times 10^{-4}$. The exact solution for this test case is $u_{\text{exact}}(x,t) = e^{-t} \sin(20\pi x)$.
	Initial condition is a high-frequency sine function:
	\begin{equation}
	u(x,0) = \sin(20\pi x)
	\end{equation}
	Boundary conditions are homogeneous Dirichlet conditions:
	\begin{equation}
	u(-1,t) = u(1,t) = 0
	\end{equation}

	Based on the proposed SV-SNN architecture, we adopt spatiotemporal separation architecture:
	\begin{equation}
	u^{\Theta}(x,t) = \sum_{n=1}^{N} c_n X_n^{(F)}(x) T_n^{(n)}(t)
	\end{equation}
	Number of network modes $N=10$, temporal network part $T_n^{(n)}(t)$ uses small-scale fully-connected networks, spatial network part $X_n^{(F)}(x)$ uses adaptive Fourier spectral feature networks:
	\begin{equation}
	X_n^{(F)}(x) = \sum_{k=1}^{K} [a_{n,k} \sin(w_{n,k} x) + b_{n,k} \cos(w_{n,k} x)] + \beta_n
	\end{equation}

	Since initial condition frequency is $20\pi$, we define characteristic frequency $w_{char}=20\pi$ for the heat conduction equation, sampling frequency number $K=40$, adopting three-level frequency sampling strategy:
	25\% low-frequency components linearly distributed in $[1,20\pi]$ range, 50\% characteristic frequencies using $\mathcal{N}(20\pi, 20^2)$ Gaussian distribution covering characteristic frequencies,
	25\% high-frequency components uniformly distributed in $[20\pi,40\pi]$ range. Temporal network $T_n^{(n)}(t)$ adopts 4-layer fully-connected network with 10 neurons per layer,
	using $\tanh$ activation function to adapt to exponential decay characteristics, SV-SNN total parameter count is 3730.
	During training, initial condition points, boundary condition points, and PDE collocation points all use Latin Hypercube Sampling.
	We adopt Adam optimizer during training with initial learning rate $\alpha = 1 \times 10^{-3}$,
	using cosine annealing learning rate strategy with decay factor 0.99, decaying every 500 training epochs, total training iterations 5,000.

	\begin{table}[htbp]
		\centering
		\caption{Performance comparison between SV-SNN and PINN on heat conduction equation ($\kappa = 20\pi$) problem}
		\label{tab:rank_comparison_heat20pi}
		\resizebox{0.8\textwidth}{!}{%
		\begin{tabular}{lcccccccccc}
		\toprule
		Method & \makecell{Total\\Parameters} & \makecell{$r_{\mathcal{I}}^{\text{eff}}$} & IC Loss & \makecell{$r_{\mathcal{B}}^{\text{eff}}$} & BC Loss & \makecell{$r_{\mathcal{F}}^{\text{eff}}$} & PDE Loss & \makecell{ReL2E} & \makecell{MAPE} \\
		\midrule
		SV-SNN & 3,730 & 74 & 2.72$\times$10$^{-8}$ & 3 & 1.85$\times$10$^{-7}$ & 61 & 5.02$\times$10$^{-7}$ & 2.56$\times$10$^{-4}$ & 3.90$\times$10$^{-4}$ \\
		PINN & 40,801 & 3 & 4.97$\times$10$^{-1}$ & 2 & 3.21$\times$10$^{-4}$ & 7 & 2.66$\times$10$^{-4}$ & 9.99$\times$10$^{-1}$ & 9.87$\times$10$^{-1}$ \\
		\bottomrule
		\end{tabular}}
	\end{table}

	The comparison results in Table~\ref{tab:rank_comparison_heat20pi} demonstrate significant advantages of SV-SNN over traditional PINN methods across multiple key dimensions.
	\textbf{Parameter Efficiency}: SV-SNN achieves high-precision solving using only 3,730 parameters, reducing 91% compared to PINN's 40,801 parameters. This significant improvement in parameter efficiency is mainly due to the spatiotemporal separation architecture design, avoiding high-dimensional parameter redundancy in traditional neural networks, and efficient utilization of adaptive Fourier spectral feature networks for frequency domain representation. More importantly, effective rank analysis of Jacobian matrices reveals essential differences in parameter utilization efficiency: SV-SNN's effective rank under initial condition constraints $r_{\mathcal{I}}^{\text{eff}} = 74$ far exceeds PINN's $r_{\mathcal{I}}^{\text{eff}} = 3$, and effective rank under PDE constraints $r_{\mathcal{F}}^{\text{eff}} = 61$ significantly higher than PINN's $r_{\mathcal{F}}^{\text{eff}} = 7$, indicating SV-SNN successfully avoids parameter space collapse problems and achieves full effective utilization of parameter space. \textbf{Convergence Speed}: From loss function values, SV-SNN exhibits excellent convergence performance across all constraint conditions, with initial condition loss reaching ultra-low level of $2.72 \times 10^{-8}$, boundary condition loss of $1.85 \times 10^{-7}$, and PDE loss of $5.02 \times 10^{-7}$, all far superior to traditional PINN methods. Particularly in initial condition constraints, PINN's loss value reaches $4.97 \times 10^{-1}$, indicating serious fitting difficulties when handling high-frequency characteristics of initial conditions. High effective rank maintenance enables SV-SNN to utilize more gradient directions for optimization, achieving faster convergence speed and more stable training process. \textbf{Prediction Accuracy}: SV-SNN achieves significant breakthroughs in final solution accuracy, with relative L2 error reaching $2.56 \times 10^{-4}$ and maximum pointwise absolute error of $3.90 \times 10^{-4}$, while PINN's corresponding errors reach $9.99 \times 10^{-1}$ and $9.87 \times 10^{-1}$ respectively, representing accuracy improvement over 3 orders of magnitude. This enormous accuracy difference directly reflects fundamental performance differences between the two methods when handling high-frequency partial differential equations: SV-SNN effectively captures high-frequency oscillation patterns of solutions through separation of variables architecture and adaptive spectral features, while traditional PINN struggles to learn complex high-frequency characteristics due to parameter space collapse caused by insufficient effective rank.

\begin{figure}[htbp]
		\centering
		\begin{subfigure}{\textwidth}
			\centering
			\includegraphics[width=0.8\textwidth]{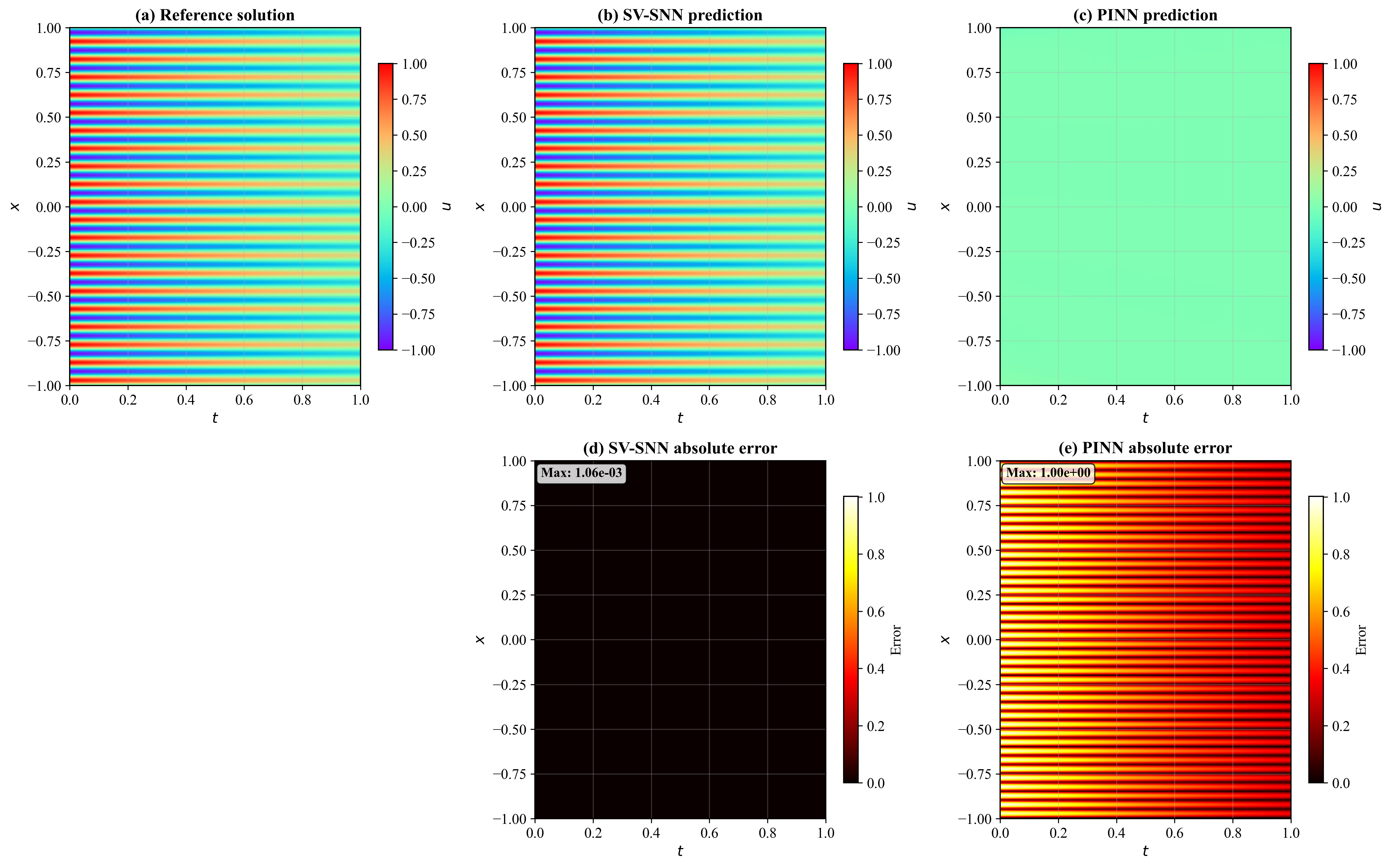}
			\caption{Prediction results and pointwise error distribution of SV-SNN and PINN}
			\label{fig:heat20pi_solutions_errors}
		\end{subfigure}
		
		\vspace{0.5cm}
		
		\begin{subfigure}{\textwidth}
			\centering
			\includegraphics[width=0.8\textwidth]{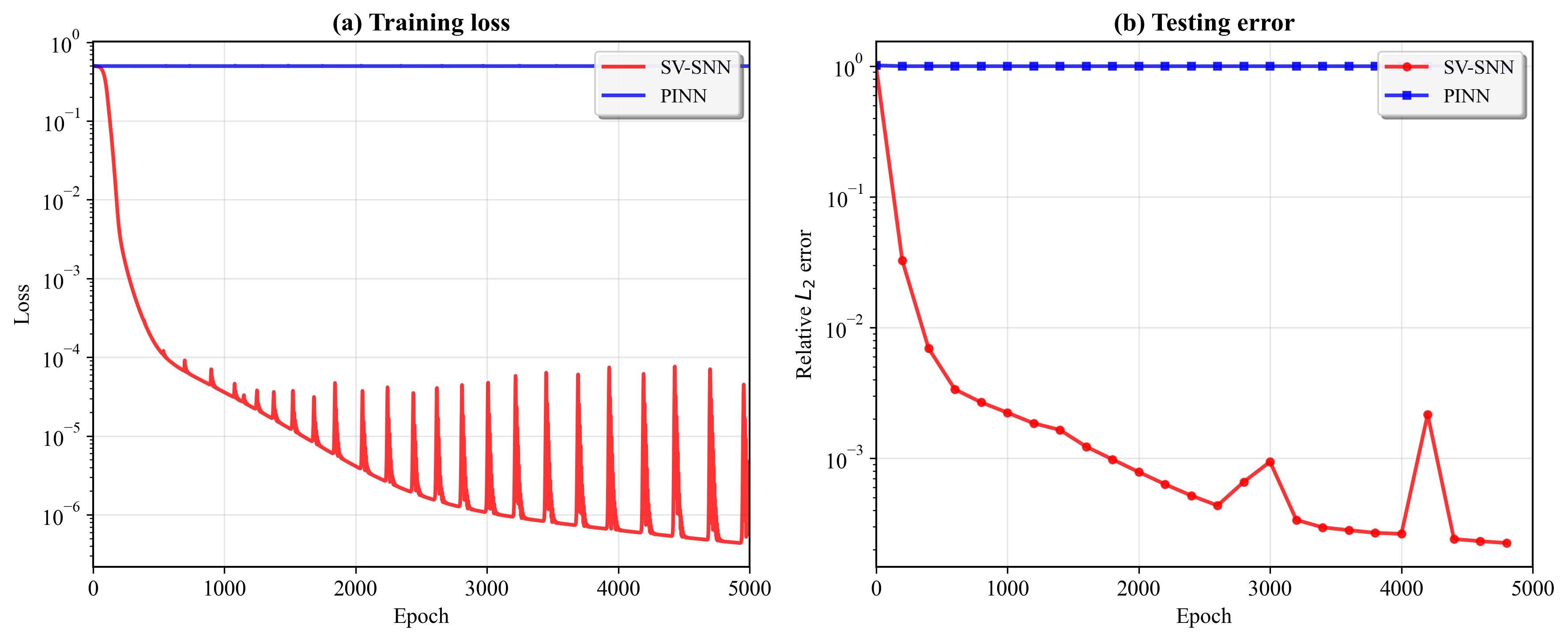}
			\caption{Training dynamics of SV-SNN and PINN, including training loss and test error}
			\label{fig:heat20pi_training_dynamics}
		\end{subfigure}
		
		\vspace{0.5cm}
		
		\begin{subfigure}{\textwidth}
			\centering
			\includegraphics[width=0.8\textwidth]{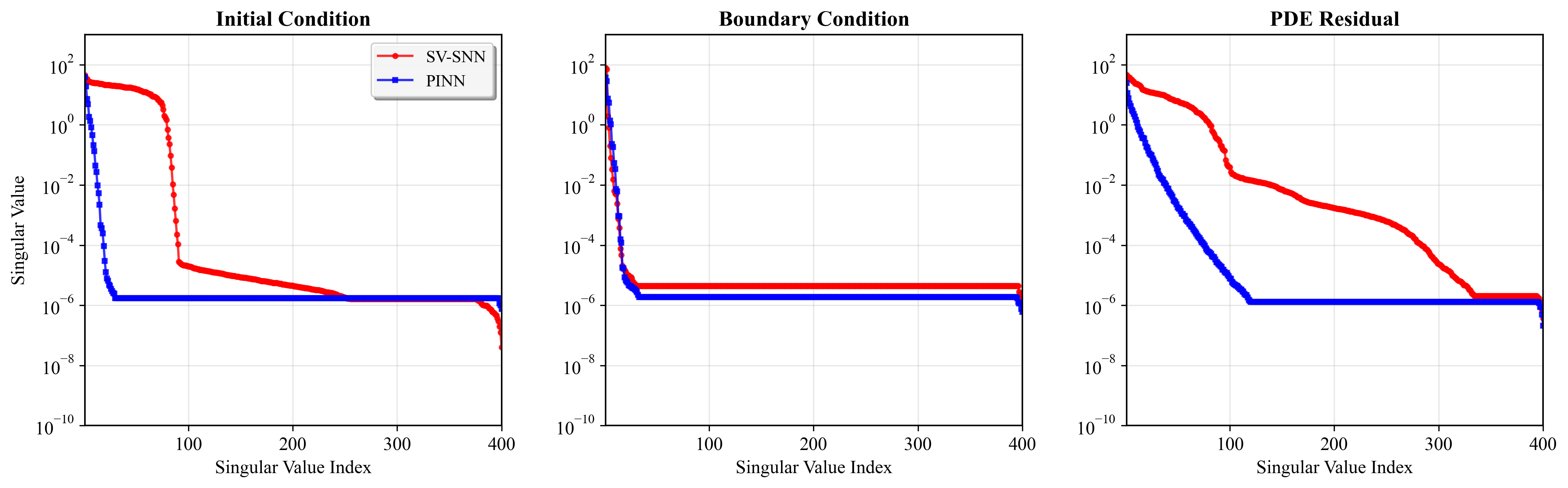}
			\caption{Singular value distributions of Jacobian matrices for SV-SNN and PINN}
			\label{fig:heat20pi_svd_distributions}
		\end{subfigure}

		\caption{Heat conduction equation ($\kappa = 20\pi$): Prediction performance, training dynamics, and singular value distributions of SV-SNN and PINN}
		\label{fig:heat20pi_combined}
	\end{figure}

	Figure~\ref{fig:heat20pi_solutions_errors} shows that SV-SNN can accurately capture high-frequency oscillation characteristics of solutions, with predicted solutions matching reference solutions well, while PINN's predicted solutions clearly deviate from true solutions. Figure~\ref{fig:heat20pi_training_dynamics} indicates SV-SNN exhibits fast and stable convergence behavior from early training stages, with loss functions rapidly decreasing to extremely low levels and remaining stable, test errors continuously decreasing during training and finally converging to $10^{-4}$ order of magnitude; while PINN shows obvious convergence difficulties during training, with test errors consistently maintaining high levels. The fundamental reason for this convergence difference is that SV-SNN provides more effective gradient optimization directions by maintaining high effective rank, while PINN suffers from insufficient gradient information due to parameter space collapse. Figure~\ref{fig:heat20pi_svd_distributions} intuitively validates the theoretical framework of effective rank analysis. SV-SNN's Jacobian matrices maintain rich singular value distributions under various constraint conditions, indicating full utilization of parameter space, particularly under initial condition and PDE constraints, where numerous singular values remain at high levels, forming "heavy-tail" distribution characteristics; in contrast, PINN's singular value distribution exhibits obvious "rapid decay" patterns, with most singular values approaching zero and only a few principal singular values playing roles. More importantly, SV-SNN's singular values are consistently significantly larger than PINN's, indicating SV-SNN's convergence speed for both low and high frequencies is significantly higher than PINN's.

	\subsubsection{$\kappa$ = 100$\pi$}
	
	To further verify SV-SNN's high-frequency processing capability, we consider initial condition frequency $\kappa = 100\pi$, where diffusion coefficient $\alpha = \frac{1}{(100\pi)^2}$, initial condition $u(x,0) = \sin(100\pi x)$, boundary conditions $u(\pm 1,t) = 0$. Analytical solution is $u_{\text{exact}}(x,t) = e^{-t} \sin(100\pi x)$.

	Using $N=4$ spatiotemporal separation modes, each spatial mode contains $K=50$ adaptive Fourier spectral features:
	\begin{equation}
	u^{\Theta}(x,t) = \sum_{n=1}^{N} c_n \left[\sum_{k=1}^{K} [a_{n,k} \cos(\omega_{n,k} x) + b_{n,k} \sin(\omega_{n,k} x)] + \beta_n\right] T_n(t)
	\end{equation}

	Based on initial condition characteristic frequency $100\pi$, we adopt three-level sampling for frequency sampling: 25% low-frequency components linearly distributed in $[1,100\pi]$ range, 50% characteristic frequencies using $\mathcal{N}(100\pi, 50^2)$ Gaussian distribution covering characteristic frequencies, 25% high-frequency components uniformly distributed in $[100\pi,200\pi]$ range. Temporal network $T_n^{(n)}(t)$ adopts 4-layer fully-connected network with 10 neurons per layer, total network parameters 1612, using Adam optimizer, training 2,000 epochs.
	
	\begin{table}[htbp]
		\centering
		\caption{Performance comparison between SV-SNN and PINN on Heat100$\pi$ problem}
		\label{tab:rank_comparison_heat100pi}
		\resizebox{0.8\textwidth}{!}{%
		\begin{tabular}{lcccccccccc}
		\toprule
		Method & \makecell{Total\\Parameters} & \makecell{$r_{\mathcal{I}}^{\text{eff}}$} & IC Loss & \makecell{$r_{\mathcal{B}}^{\text{eff}}$} & BC Loss & \makecell{$r_{\mathcal{F}}^{\text{eff}}$} & PDE Loss & \makecell{ReL2E} & \makecell{MAPE} \\
		\midrule
		SV-SNN & 1,612 & 145 & 3.04$\times$10$^{-6}$ & 2 & 2.13$\times$10$^{-8}$ & 111 & 5.44$\times$10$^{-6}$ & 5.82$\times$10$^{-3}$ & 1.57$\times$10$^{-2}$ \\
		PINN & 58,561 & 2 & 4.98$\times$10$^{-1}$ & 2 & 1.59$\times$10$^{-6}$ & 5 & 1.07$\times$10$^{-6}$ & 1.00 & 9.88$\times$10$^{-1}$ \\
		\bottomrule
		\end{tabular}}
	\end{table}

	Experimental results in Table~\ref{tab:rank_comparison_heat100pi} fully validate SV-SNN's excellent performance in solving $\kappa = 100\pi$ high-frequency heat conduction problems. In computational resource consumption, SV-SNN achieves high-quality numerical simulation using only 1,612 learnable parameters, compared to PINN requiring 58,561 parameters, improving parameter usage efficiency by 97%. This significant advantage benefits from organic combination of spatiotemporal separation representation structure and adaptive spectral feature learning mechanisms.
	
\begin{figure}[htbp]
		\centering
		\begin{subfigure}{\textwidth}
			\centering
			\includegraphics[width=0.8\textwidth]{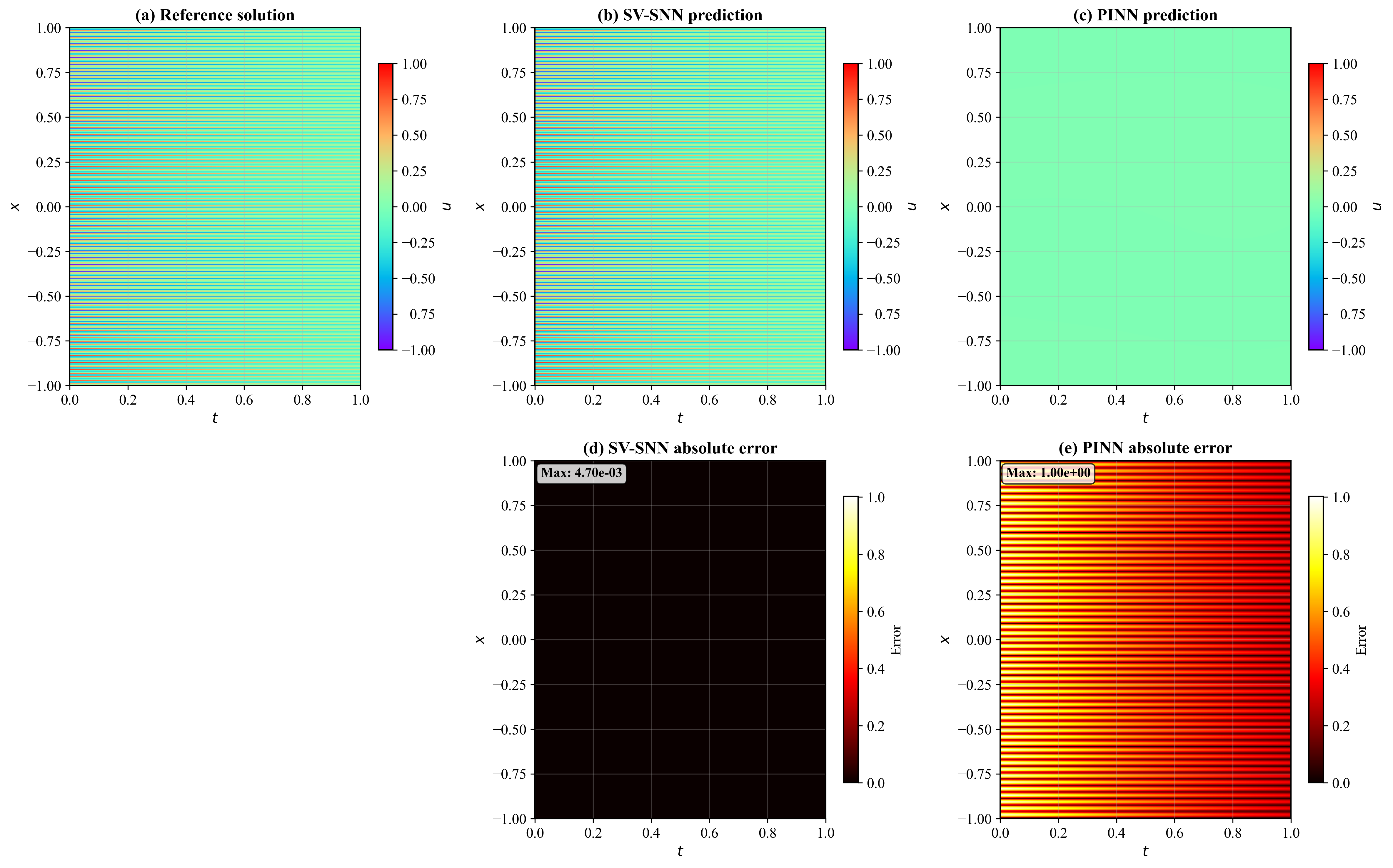}
			\caption{Prediction results and pointwise error distribution of SV-SNN and PINN}
			\label{fig:heat100pi_solutions_errors}
		\end{subfigure}
		
		\vspace{0.5cm}
		
		\begin{subfigure}{\textwidth}
			\centering
			\includegraphics[width=0.8\textwidth]{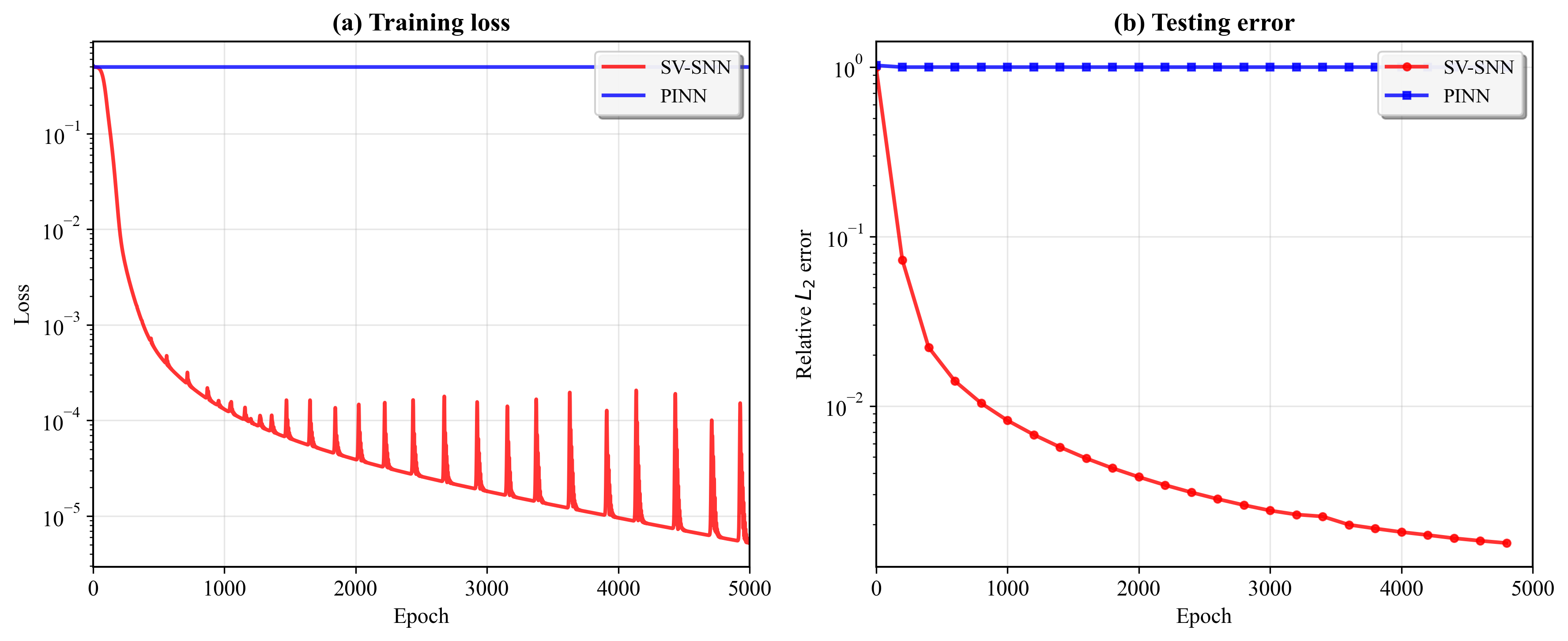}
			\caption{Training dynamics of SV-SNN and PINN, including training loss and test error}
			\label{fig:heat100pi_training_dynamics}
		\end{subfigure}
		
		\vspace{0.5cm}
		
		\begin{subfigure}{\textwidth}
			\centering
			\includegraphics[width=0.8\textwidth]{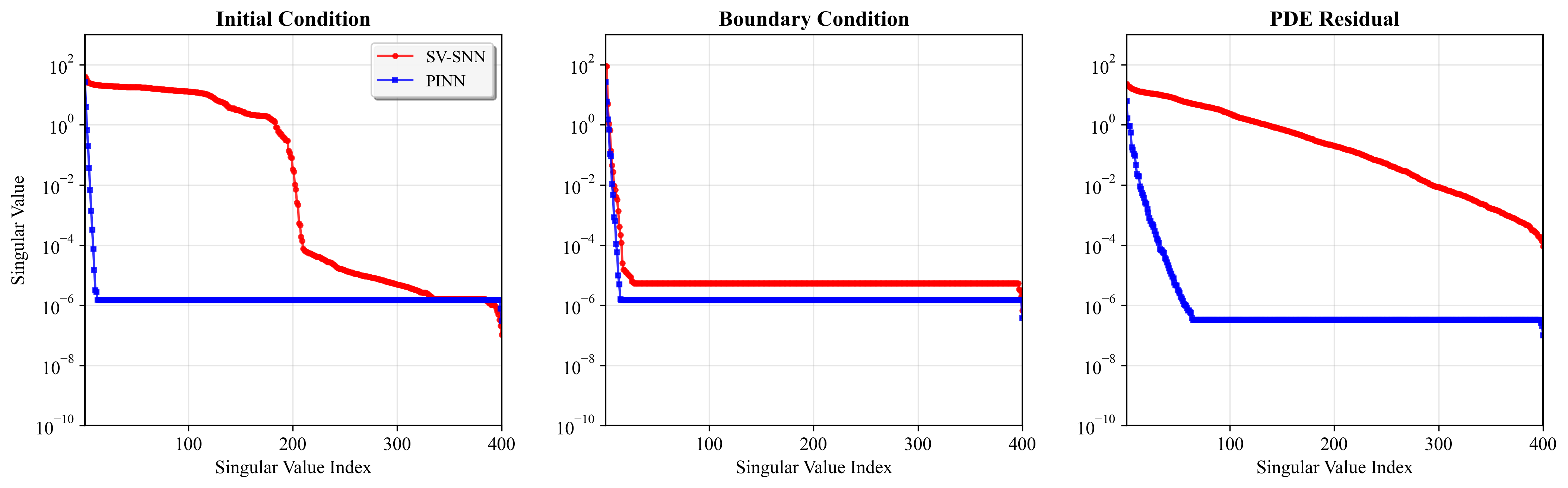}
			\caption{Singular value distributions of Jacobian matrices for SV-SNN and PINN}
			\label{fig:heat100pi_singular_value_distributions}
		\end{subfigure}
		
		\caption{Heat conduction equation ($\kappa = 100\pi$): Prediction performance, training dynamics, and singular value distributions of SV-SNN and PINN}
		\label{fig:heat100pi_combined}
	\end{figure}

	Through effective rank theoretical analysis of Jacobian matrices, intrinsic mechanism differences between the two methods are quantitatively characterized: SV-SNN achieves rich effective dimensions of $r_{\mathcal{I}}^{\text{eff}} = 145$ under initial condition constraints, greatly exceeding PINN's $r_{\mathcal{I}}^{\text{eff}} = 2$, and effective rank under physical equation constraints $r_{\mathcal{F}}^{\text{eff}} = 111$ also significantly superior to PINN's $r_{\mathcal{F}}^{\text{eff}} = 5$. This comparison result indicates that even facing $100\pi$ high-frequency challenges, SV-SNN can still effectively avoid parameter degradation phenomena and fully utilize network parameters' representation potential. Figure~\ref{fig:heat100pi_singular_value_distributions} provides intuitive validation for effective rank theory. SV-SNN maintains rich and balanced singular value distributions under various physical constraints, especially under initial condition and differential equation constraints, where numerous singular values maintain considerable numerical levels, while PINN's singular value spectrum exhibits typical "rapid decay" patterns with most singular values quickly decreasing to near-zero states. This distribution characteristic intuitively explains PINN's performance degradation mechanism in high-frequency scenarios.
	
	In numerical convergence performance, SV-SNN demonstrates good convergence speed and training stability, with initial condition loss decreasing to $3.04 \times 10^{-6}$, boundary condition loss reaching $2.13 \times 10^{-8}$, and differential equation loss of $5.44 \times 10^{-6}$, all indicators significantly superior to PINN. Particularly noteworthy is that PINN's initial condition fitting loss reaches $4.98 \times 10^{-1}$, fully exposing its essential defects when handling $100\pi$ high-frequency initial value problems. Figure~\ref{fig:heat100pi_training_dynamics} demonstrates essential differences in training processes between the two algorithms: SV-SNN exhibits rapid and smooth convergence trends from early training stages, with test errors monotonically decreasing and finally stabilizing at $10^{-3}$ order of magnitude, while PINN shows obvious convergence obstacles throughout training cycles, with test errors lingering at undesirable levels near 1 for extended periods.
	
	In solution approximation accuracy, SV-SNN achieves relative L2 error of $5.82 \times 10^{-3}$ and maximum absolute error of $1.57 \times 10^{-2}$, while PINN's corresponding error indicators are $1.00$ and $9.88 \times 10^{-1}$ respectively, representing accuracy improvement of two orders of magnitude. Figure~\ref{fig:heat100pi_solutions_errors} intuitively demonstrates SV-SNN's accurate reproduction capability for $100\pi$ high-frequency solution structures, with numerical solutions achieving high consistency with analytical solutions and error distributions exhibiting uniform low-magnitude characteristics. Relatively, PINN's prediction results show serious deviations from true solutions, with error distributions clearly indicating its severely insufficient processing capability for ultra-high-frequency characteristics.

	\subsubsection{$\kappa$ = 500$\pi$}
	
	To verify the method's performance in ultra-high-frequency heat conduction problems, we consider initial condition frequency $\kappa = 500\pi$. The problem is defined on $\Omega_T = [-1,1] \times [0,1]$:
	\begin{equation}
	\frac{\partial u}{\partial t} - \frac{1}{(500\pi)^2} \frac{\partial^2 u}{\partial x^2} = 0
	\end{equation}
	Initial condition is $u(x,0) = \sin(500\pi x)$, boundary conditions are $u(\pm 1,t) = 0$, analytical solution is $u_{\text{exact}}(x,t) = e^{-t} \sin(500\pi x)$, this solution constitutes the most challenging high-frequency spatial pattern.
	
	Using $N=4$ modes, each spatial mode uses $K=50$ Fourier spectral features, total network parameters 1612. From initial conditions, characteristic frequency $w_{char} = 500\pi$ can be determined. We adopt three-level sampling for frequency sampling: 25% low-frequency components linearly distributed in $[1,500\pi]$ range, 50% characteristic frequencies using $\mathcal{N}(500\pi, 50^2)$ Gaussian distribution covering characteristic frequencies, 25% high-frequency components uniformly distributed in $[500\pi,1000\pi]$ range. Temporal network $T_n^{(n)}(t)$ adopts 4-layer fully-connected network with 10 neurons per layer, using Adam optimizer, training 2,000 epochs.

	\begin{table}[htbp]
		\centering
		\caption{Performance comparison between SV-SNN and PINN methods on Heat500$\pi$ problem}
		\label{tab:rank_comparison_heat500pi}
		\resizebox{0.8\textwidth}{!}{%
		\begin{tabular}{lccccccccc}
		\toprule
		Method & \makecell{Total\\Parameters} & \makecell{$r_{\mathcal{I}}^{\text{eff}}$} & IC Loss & \makecell{$r_{\mathcal{B}}^{\text{eff}}$} & BC Loss & \makecell{$r_{\mathcal{F}}^{\text{eff}}$} & PDE Loss & \makecell{ReL2E} & \makecell{MAPE} \\
		\midrule
		SV-SNN & 1,412 & 88 & 6.37$\times$10$^{-5}$ & 3 & 2.08$\times$10$^{-6}$ & 78 & 1.84$\times$10$^{-6}$ & 3.75$\times$10$^{-2}$ & 4.45$\times$10$^{-2}$ \\
		PINN & 58,561 & 2 & 4.94$\times$10$^{-1}$ & 2 & 9.26$\times$10$^{-4}$ & 4 & 6.55$\times$10$^{-4}$ & 9.95$\times$10$^{-1}$ & 1.03 \\
		\bottomrule
		\end{tabular}}
	\end{table}
	
	Table~\ref{tab:rank_comparison_heat500pi} shows SV-SNN's excellent performance in $\kappa = 500\pi$ ultra-high-frequency heat conduction problems. SV-SNN achieves excellent approximation using only 1,412 parameters, compared to PINN's 58,561 parameters, improving parameter efficiency by approximately 98%.
	
\begin{figure}[htbp]
		\centering
		\begin{subfigure}{\textwidth}
			\centering
			\includegraphics[width=0.8\textwidth]{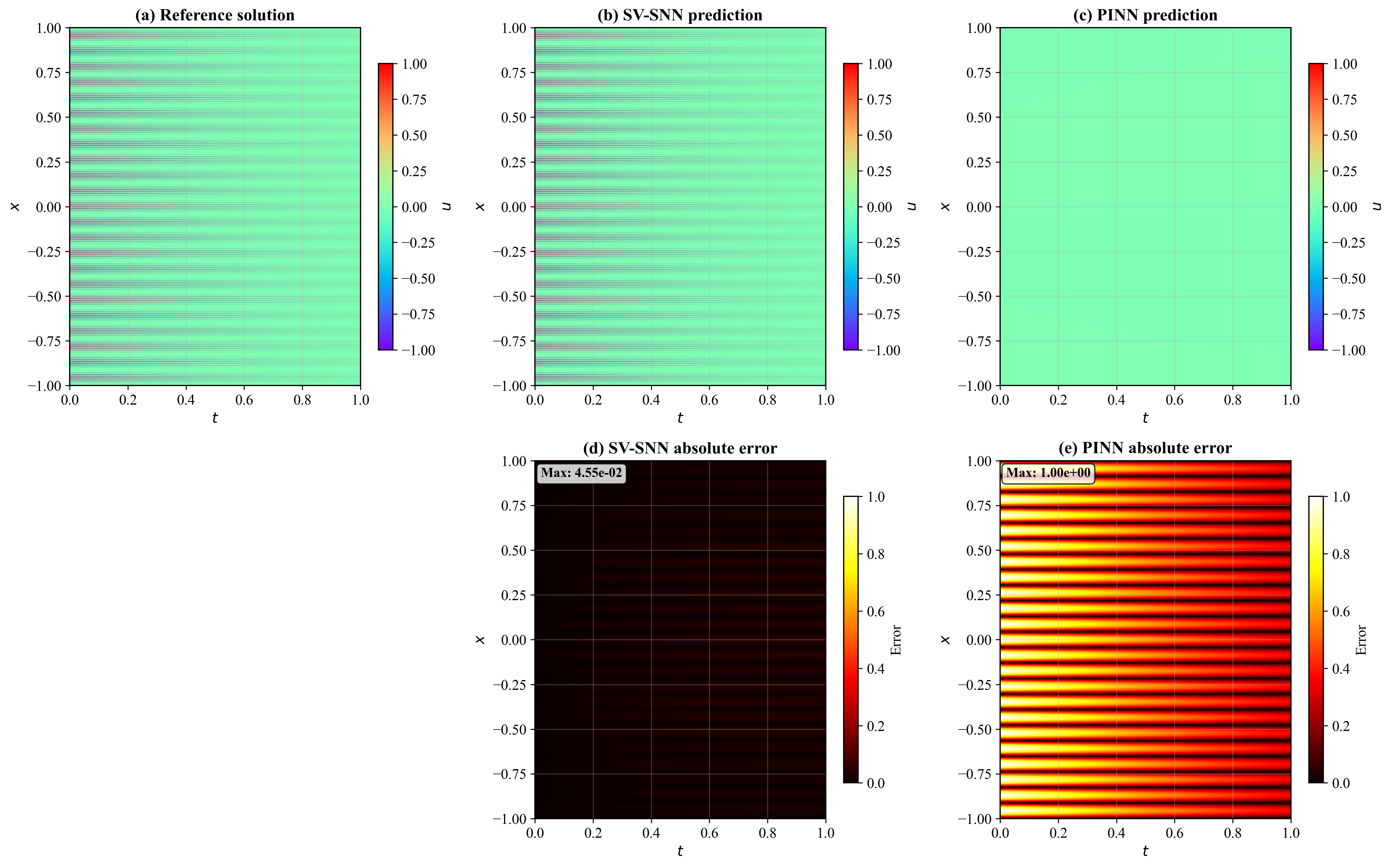}
			\caption{Prediction results and pointwise error distribution of SV-SNN and PINN}
			\label{fig:heat500pi_solutions_errors}
		\end{subfigure}
		
		\vspace{0.5cm}
		
		\begin{subfigure}{\textwidth}
			\centering
			\includegraphics[width=0.8\textwidth]{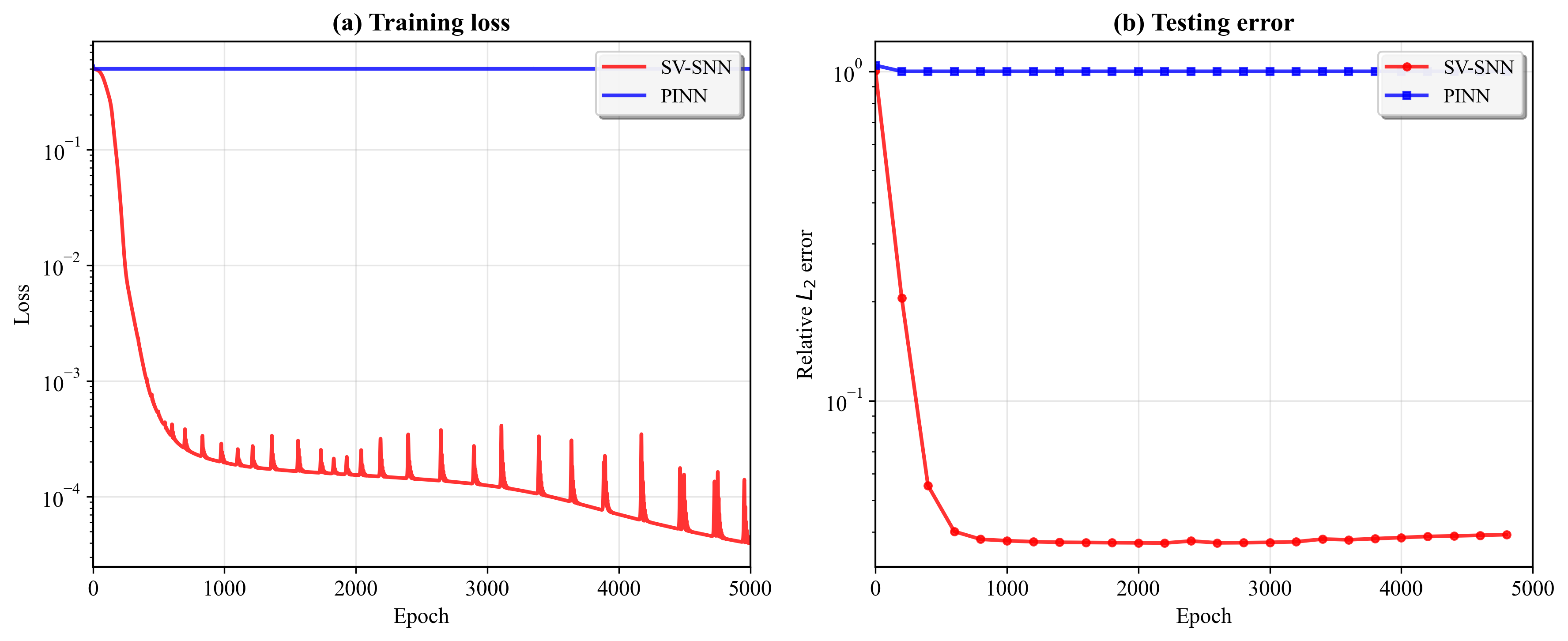}
			\caption{Training dynamics of SV-SNN and PINN, including training loss and test error}
			\label{fig:heat500pi_training_dynamics}
		\end{subfigure}
		
		\vspace{0.5cm}
		
		\begin{subfigure}{\textwidth}
			\centering
			\includegraphics[width=0.8\textwidth]{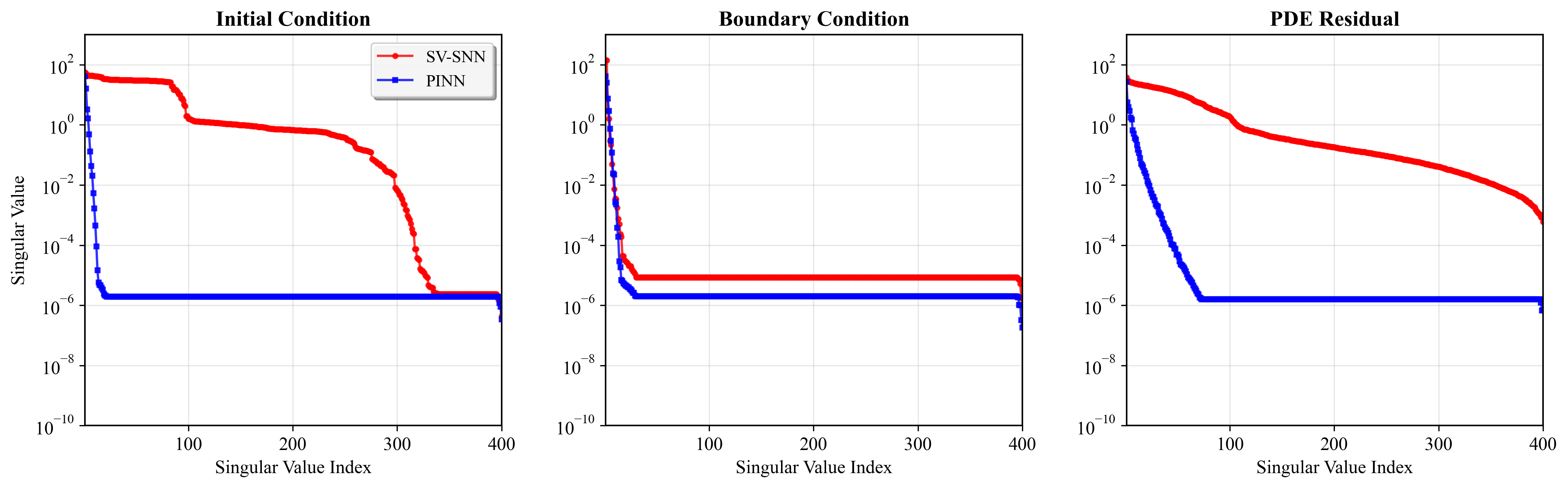}
			\caption{Singular value distributions of Jacobian matrices for SV-SNN and PINN}
			\label{fig:heat500pi_svd_distributions}
		\end{subfigure}
		
		\caption{Heat conduction equation ($\kappa = 500\pi$): Prediction performance, training dynamics, and singular value distributions of SV-SNN and PINN}
		\label{fig:heat500pi_combined}
	\end{figure}

	Effective rank analysis of Jacobian matrices reveals key differences: SV-SNN maintains $r_{\mathcal{I}}^{\text{eff}} = 88$ under initial condition constraints, far exceeding PINN's $r_{\mathcal{I}}^{\text{eff}} = 2$; under differential equation constraints $r_{\mathcal{F}}^{\text{eff}} = 78$ also significantly superior to PINN's $r_{\mathcal{F}}^{\text{eff}} = 4$. Figure~\ref{fig:heat500pi_svd_distributions} singular value spectrum analysis supports effective rank theory: SV-SNN exhibits rich balanced singular value distributions with numerous singular values maintaining significant numerical levels, while PINN shows serious "rapid decay" characteristics with most singular values rapidly decaying to near-zero values.
	
	In numerical error aspects, SV-SNN achieves relative L2 error of $3.75 \times 10^{-2}$ and maximum absolute error of $4.45 \times 10^{-2}$, while PINN's corresponding errors are $9.95 \times 10^{-1}$ and $1.03$, representing accuracy improvement over one order of magnitude. Figure~\ref{fig:heat500pi_solutions_errors} intuitively demonstrates SV-SNN's excellent prediction capability for ultra-high-frequency solution structures, with numerical solutions maintaining good consistency with analytical solutions and uniform low-magnitude error distributions, while PINN prediction results show considerable gaps from true solutions.
	
	Figure~\ref{fig:heat500pi_training_dynamics} shows training dynamic differences: SV-SNN exhibits fast stable convergence from initial stages with test errors decreasing to $10^{-2}$ order of magnitude, while PINN shows obvious convergence difficulties throughout training processes with test errors persistently maintaining high levels near 1. Particularly, initial condition loss reaches $4.94 \times 10^{-1}$, exposing its serious inadequacy in ultra-high-frequency problem handling.

	\subsection{Helmholtz Equations}
	
	\subsubsection{$\kappa$ = 24$\pi$}
	
	To test SV-SNN's performance in two-dimensional spatial high-frequency problems, we consider the two-dimensional Helmholtz equation with high-frequency oscillation modes, defined on spatial region $\Omega = [0,1] \times [0,1]$, with governing equation:
	\begin{equation}
	-\Delta u - \kappa^2 u = f(x,y)
	\end{equation}
	where $\Delta u = \frac{\partial^2 u}{\partial x^2} + \frac{\partial^2 u}{\partial y^2}$ is the two-dimensional Laplacian operator, wavenumber $\kappa = 24\pi$, and we determine the characteristic frequency $w_{char} = 24\pi$ from the wavenumber.	
	Source term is:
	\begin{equation}
	f(x,y) = \kappa^2 \sin(\kappa x) \sin(\kappa y)
	\end{equation}
	Boundary conditions are homogeneous Dirichlet conditions:
	\begin{equation}
	u(x,y) = 0, \quad (x,y) \in \partial\Omega
	\end{equation}
	Analytical solution for this problem is:
	\begin{equation}
	u_{\text{exact}}(x,y) = \sin(\kappa x) \sin(\kappa y)
	\end{equation}
	This solution exhibits high-frequency oscillations in both spatial directions.

	To solve using SV-SNN, we adopt two-dimensional spatial separation architecture:
	\begin{equation}
	u^{\Theta}(x,y) = \sum_{n=1}^{N} c_n X_n^{(F)}(x) Y_n^{(F)}(y)
	\end{equation}
	where $X_n^{(F)}(x)$ and $Y_n^{(F)}(y)$ are adaptive Fourier spectral feature networks for $x$ and $y$ directions respectively:
	\begin{align}
	X_n^{(F)}(x) &= \sum_{k=1}^{32} [a_{n,k}^{(x)} \sin(w_{n,k}^{(x)} x) + b_{n,k}^{(x)} \cos(w_{n,k}^{(x)} x)] + \beta_n^{(x)} \\
	Y_n^{(F)}(y) &= \sum_{k=1}^{32} [a_{n,k}^{(y)} \sin(w_{n,k}^{(y)} y) + b_{n,k}^{(y)} \cos(w_{n,k}^{(y)} y)] + \beta_n^{(y)}
	\end{align}

	Using $N=6$ network modes, each spatial mode uses $K=64$ Fourier spectral features, total parameters 2,322. We also adopt three-level frequency sampling strategy: 25% low-frequency components linearly distributed in $[1,24\pi]$ range, 50% characteristic frequencies using $\mathcal{N}(24\pi, 20^2)$ Gaussian distribution covering characteristic frequencies, 25% high-frequency components uniformly distributed in $[24\pi,48\pi]$ range.

	\begin{figure}[htbp]
		\centering
		\begin{subfigure}{\textwidth}
			\centering
			\includegraphics[width=0.8\textwidth]{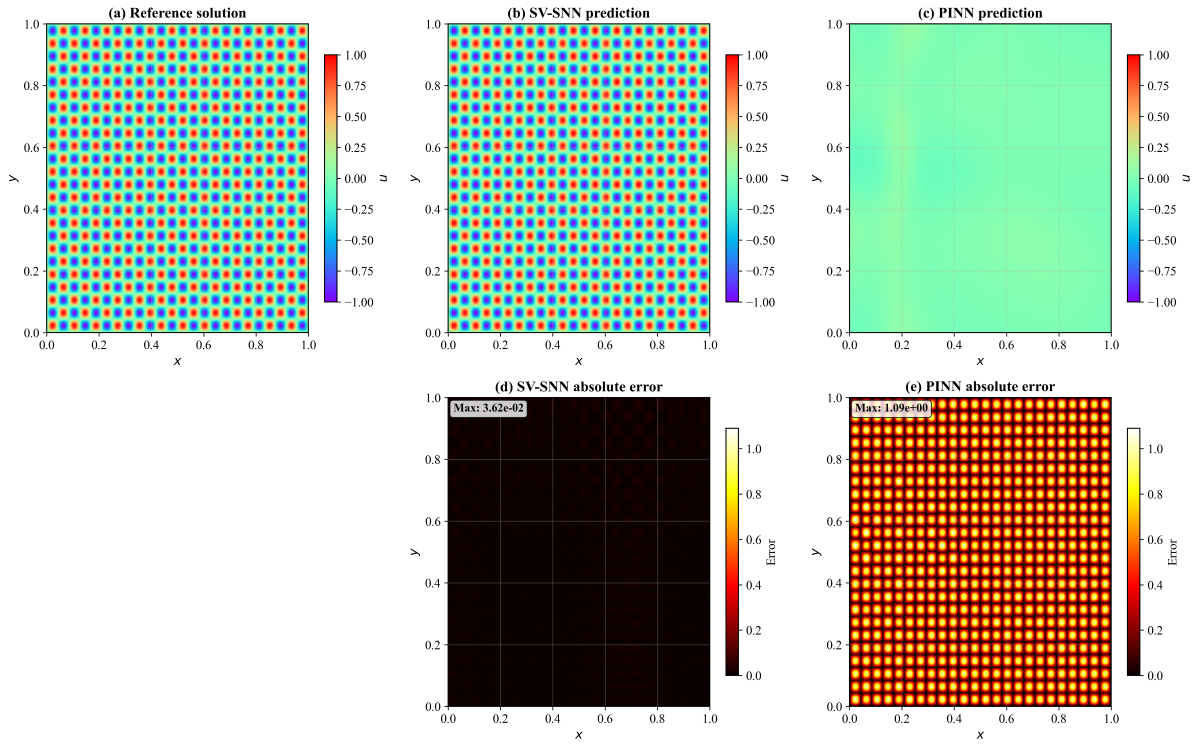}
			\caption{Prediction results and pointwise error distribution of SV-SNN and PINN}
			\label{fig:helmholtz24pi_solutions_errors}
		\end{subfigure}
		
		\vspace{0.5cm}
		
		\begin{subfigure}{\textwidth}
			\centering
			\includegraphics[width=0.8\textwidth]{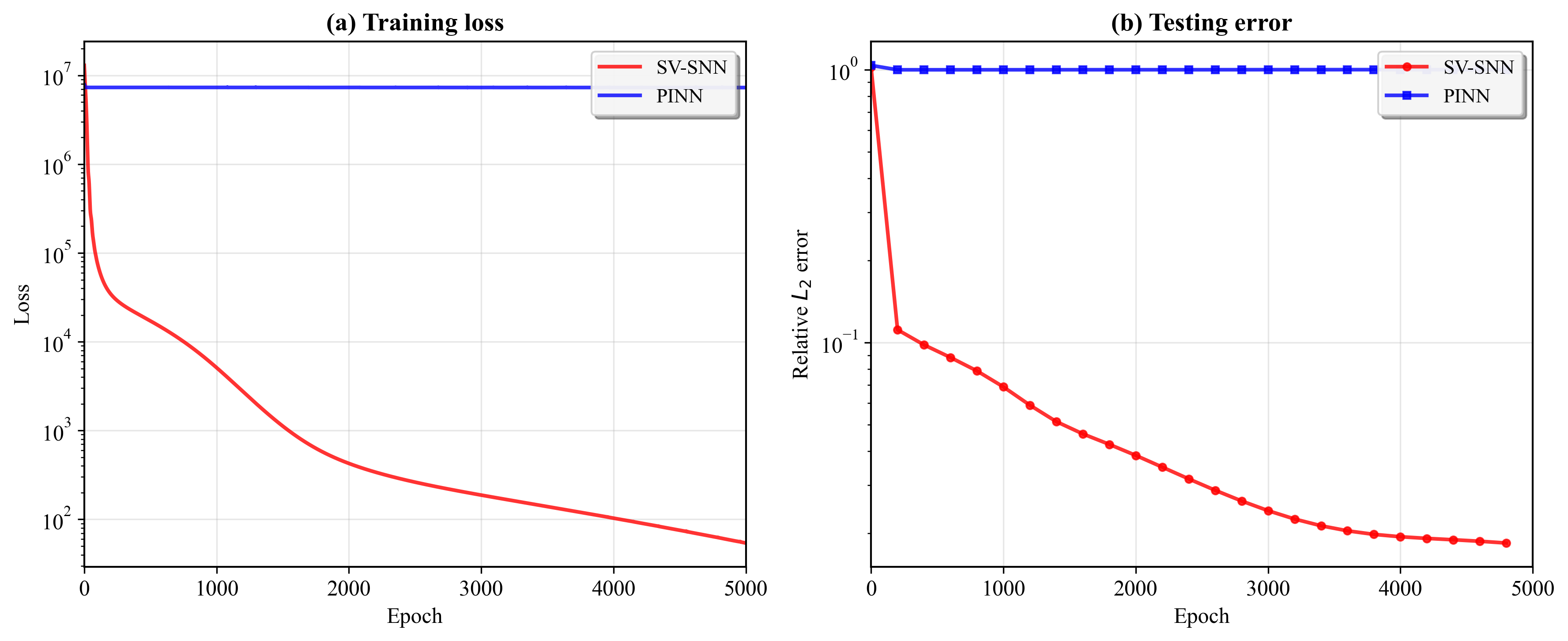}
			\caption{Training dynamics of SV-SNN and PINN, including training loss and test error}
			\label{fig:helmholtz24pi_training_dynamics}
		\end{subfigure}
		
		\vspace{0.5cm}
		
		\begin{subfigure}{\textwidth}
			\centering
			\includegraphics[width=0.8\textwidth]{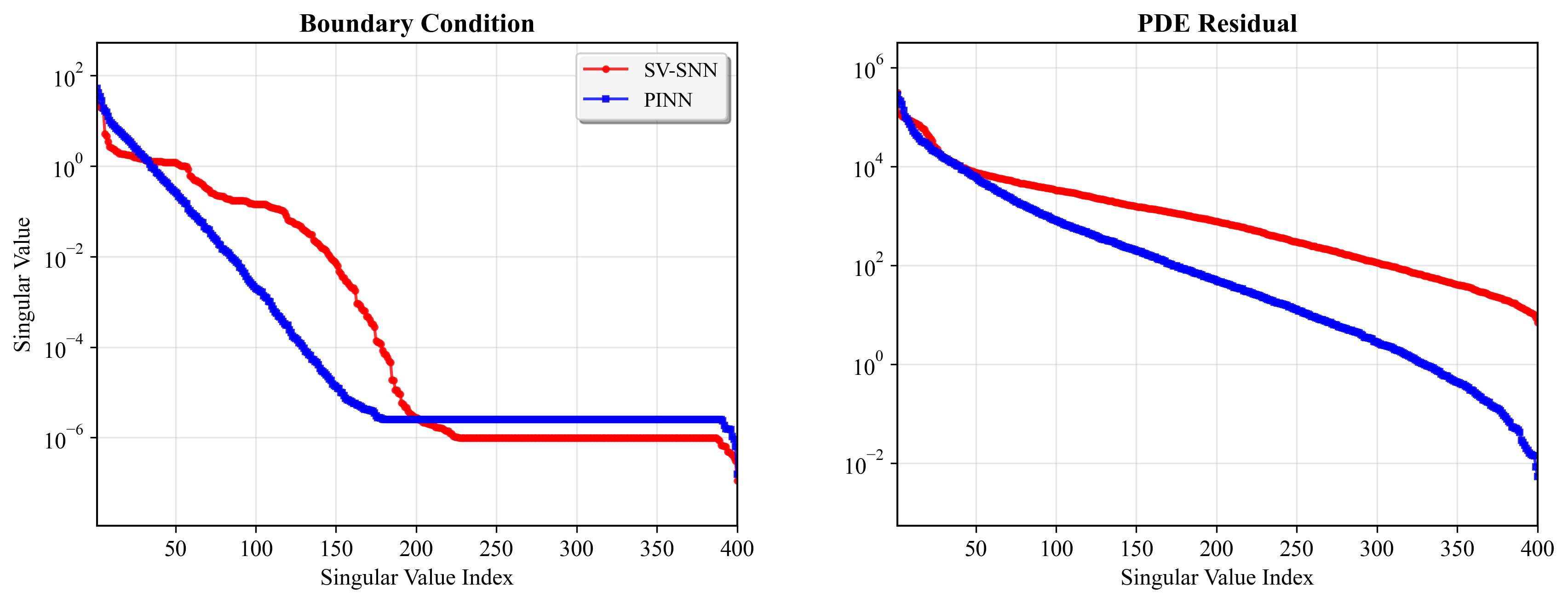}
			\caption{Singular value distributions of Jacobian matrices for SV-SNN and PINN}
			\label{fig:helmholtz24pi_svd_distributions}
		\end{subfigure}
		
		\caption{Two-dimensional Helmholtz equation ($\kappa = 24\pi$): Prediction performance, training dynamics, and singular value distributions of SV-SNN and PINN}
		\label{fig:helmholtz24pi_combined}
	\end{figure}

	\begin{table}[htbp]
		\centering
		\caption{Performance comparison between SV-SNN and PINN methods on two-dimensional Helmholtz equation ($\kappa = 24\pi$) problem}
		\label{tab:rank_comparison_helmholtz24pi}
		\resizebox{0.8\textwidth}{!}{%
		\begin{tabular}{lcccccccc}
		\toprule
		Method & \makecell{Total\\Parameters} & \makecell{$r_{\mathcal{B}}^{\text{eff}}$} & BC Loss & \makecell{$r_{\mathcal{F}}^{\text{eff}}$} & PDE Loss & \makecell{ReL2E} & \makecell{MAPE} \\
		\midrule
		SV-SNN & 2,322 & 43 & 9.30$\times$10$^{-4}$ & 41 & 2.37$\times$10$^{1}$ & 1.33$\times$10$^{-2}$ & 3.62$\times$10$^{-2}$ \\
		PINN & 40,801 & 20 & 3.50$\times$10$^{-3}$ & 29 & 7.18$\times$10$^{6}$ & 1.01$\times$10$^{0}$ & 1.09$\times$10$^{0}$ \\
		\bottomrule
		\end{tabular}}
	\end{table}

	Experimental results in Table~\ref{tab:rank_comparison_helmholtz24pi} fully validate SV-SNN's excellent performance in solving $\kappa = 24\pi$ high-frequency Helmholtz equations. In parameter efficiency, SV-SNN achieves high-quality solving using only 2,322 parameters, compared to PINN's 40,801 parameters, improving efficiency by 94%.
	
	Effective rank analysis of Jacobian matrices reveals essential differences between the two methods: SV-SNN's effective rank under boundary condition constraints $r_{\mathcal{B}}^{\text{eff}} = 43$ significantly exceeds PINN's $r_{\mathcal{B}}^{\text{eff}} = 20$, and effective rank under PDE constraints $r_{\mathcal{F}}^{\text{eff}} = 41$ also clearly superior to PINN's $r_{\mathcal{F}}^{\text{eff}} = 29$. These high effective rank characteristics directly translate to significant improvements in prediction performance: SV-SNN achieves relative L2 error of $1.33 \times 10^{-2}$ and maximum absolute error of $3.62 \times 10^{-2}$, while PINN's corresponding errors are $1.01$ and $1.09$ respectively, representing accuracy improvement of two orders of magnitude. More importantly, SV-SNN's differential equation loss is $2.37 \times 10^{1}$ while PINN reaches $7.18 \times 10^{6}$, this enormous difference indicates high effective rank ensures SV-SNN's effective learning capability for high-frequency characteristics.
	
	Figure~\ref{fig:helmholtz24pi_svd_distributions} singular value distribution validates effective rank theory: SV-SNN maintains rich singular value distributions under various constraint conditions, providing more effective gradient optimization directions, while PINN's rapid singular value decay limits optimization space. Figure~\ref{fig:helmholtz24pi_training_dynamics} shows SV-SNN exhibits fast stable convergence processes with monotonically decreasing test errors, while PINN suffers convergence difficulties due to insufficient gradient information from low effective rank. Figure~\ref{fig:helmholtz24pi_solutions_errors} intuitively demonstrates close correlation between Jacobian matrix effective rank and prediction accuracy: high effective rank enables SV-SNN to accurately reproduce $24\pi$ high-frequency oscillation patterns with predicted solutions highly consistent with analytical solutions and uniform low-magnitude error distributions.

	\subsubsection{Complex Geometry Helmholtz Equation ($\kappa$ = 24$\pi$)}
	
	Furthermore, we test SV-SNN's adaptability in complex geometric domains. Consider two-dimensional Helmholtz equation with internal cylindrical obstacles, defined on complex region $\Omega = [0,1] \times [0,1] \setminus \Omega_c$, where $\Omega_c$ is cylindrical hole region, governing equation consistent with previous Helmholtz equation, but wavenumber $\kappa = 24\pi$, source term $f(x,y) = \kappa^2 \sin(\kappa x) \sin(\kappa y)$.
	
	Cylindrical obstacle configuration: center at $(0.5, 0.5)$, radius $r = 0.15$. 	Boundary conditions are:
	\begin{align}
	u(x,y) &= 0, \quad (x,y) \in \partial\Omega_{\text{ext}} \\
	u(x,y) &= \sin(\kappa x)\sin(\kappa y), \quad (x,y) \in \partial\Omega_c
	\end{align}
	Internal boundary condition:
	\begin{equation}
	u(x,y) = \sin(\kappa x)\sin(\kappa y), \quad (x,y) \in \partial\Omega_c
	\end{equation}
	
	Using two-dimensional spatial separation architecture, network modes $N=6$, each direction uses $K=64$ adaptive Fourier features, SV-SNN total parameters 2,323, frequency sampling strategy same as $\kappa=24\pi$ two-dimensional Helmholtz equation. Training collocation points use random sampling, including 400 external boundary points, 200 cylindrical boundary points, 10,000 PDE collocation points, using Adam optimizer, training 5,000 epochs.
	
	\begin{figure}[htbp]
		\centering
		\begin{subfigure}[b]{\textwidth}
			\centering
			\includegraphics[width=0.8\textwidth]{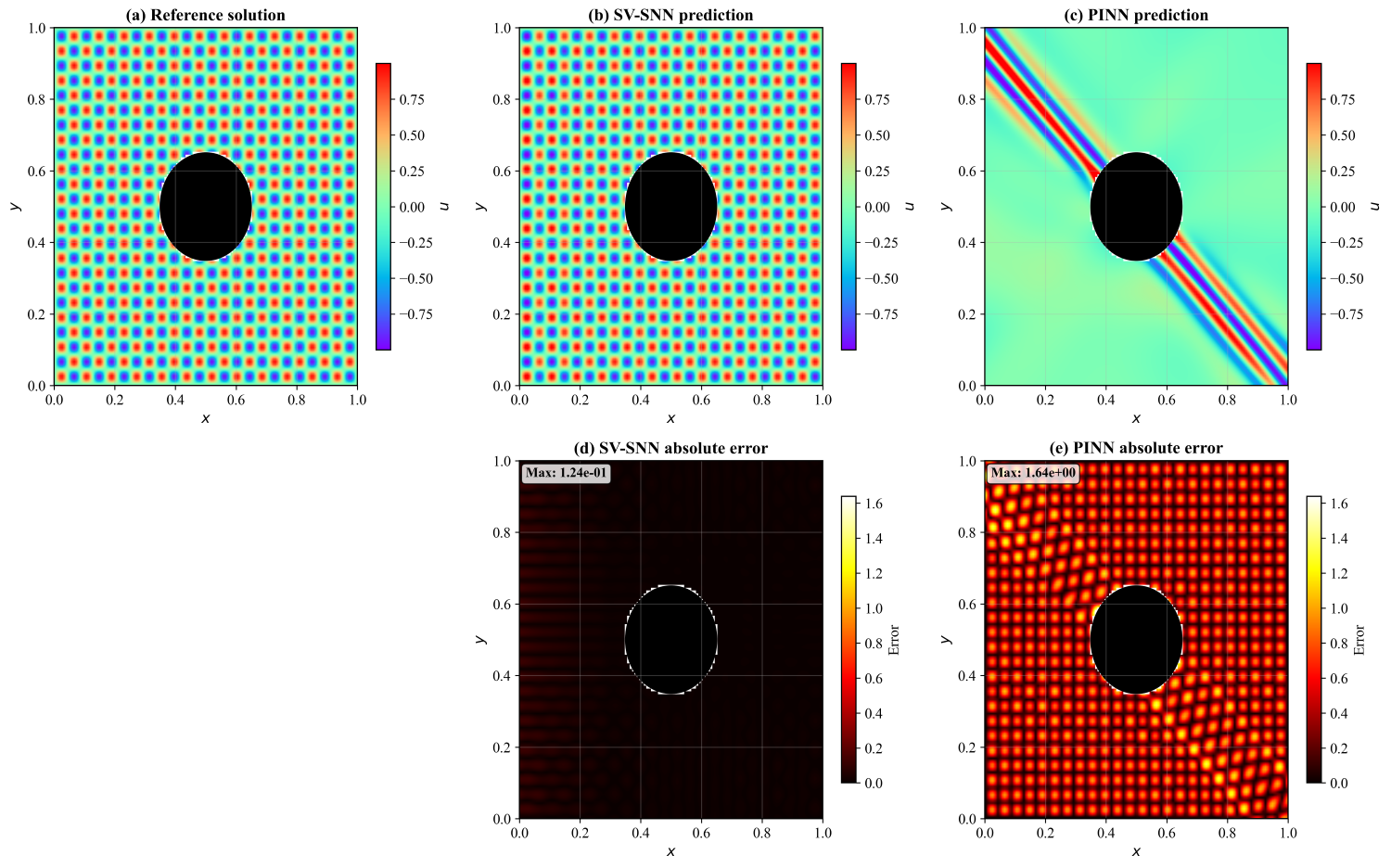}
			\caption{Prediction results and pointwise error distribution of SV-SNN and PINN}
			\label{fig:helmholtz_complex_geometry_solutions_errors}
		\end{subfigure}
		
		\begin{subfigure}[b]{\textwidth}
			\centering
			\includegraphics[width=0.8\textwidth]{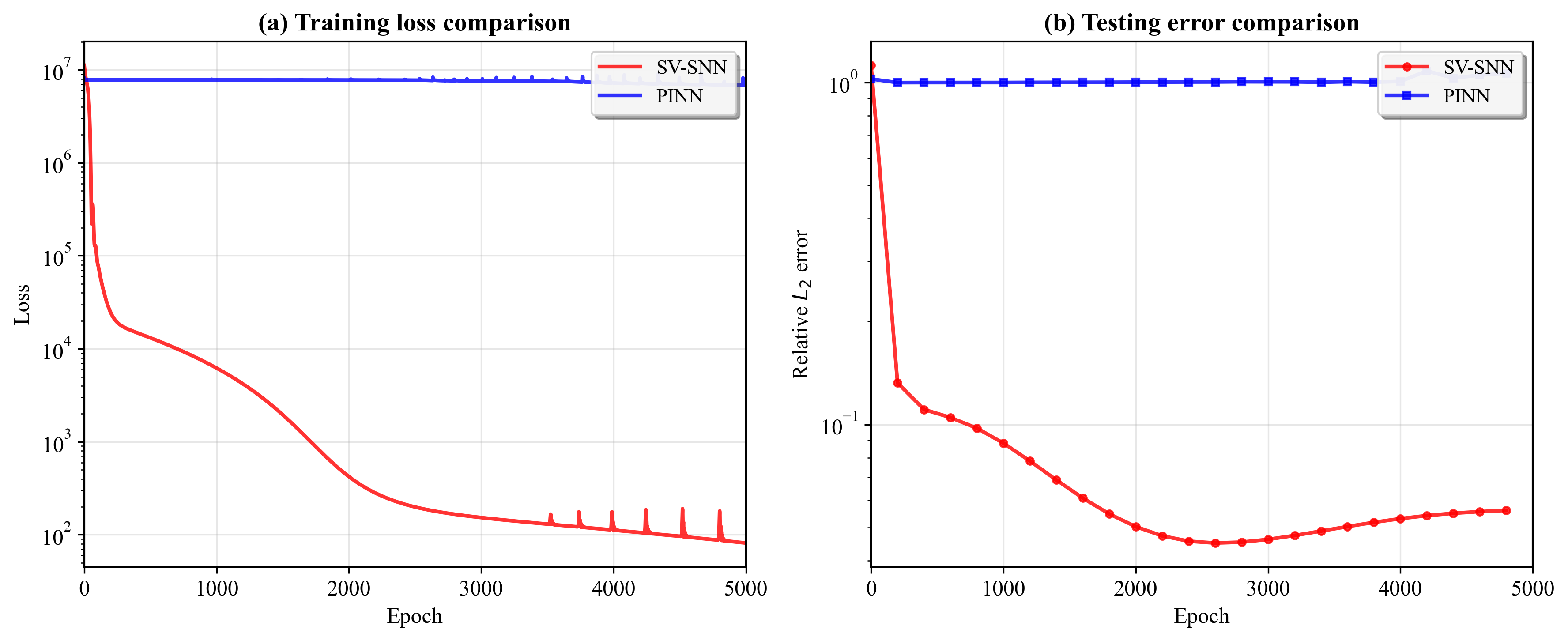}
			\caption{Training dynamics comparison of SV-SNN and PINN}
			\label{fig:helmholtz_complex_geometry_training_dynamics}
		\end{subfigure}
		\caption{Two-dimensional Helmholtz equation with cylindrical obstacles ($\kappa = 24\pi$): Prediction results and training dynamics analysis of SV-SNN and PINN}
		\label{fig:helmholtz_complex_geometry_combined}
	\end{figure}

	\subsubsection{$\kappa$ = 48$\pi$}
	
	For two-dimensional Helmholtz equation with wavenumber $\kappa = 48\pi$, characteristic frequency can be determined as $w_{char} = 48\pi$, source term $f(x,y) = \kappa^2 \sin(\kappa x) \sin(\kappa y)$, boundary conditions are homogeneous Dirichlet conditions $u(x,y) = 0, (x,y) \in \partial\Omega$. Analytical solution is $u_{\text{exact}}(x,y) = \sin(\kappa x) \sin(\kappa y)$.
	
	Using two-dimensional separation architecture, network modes $N=8$, each direction uses $K=64$ adaptive Fourier features, total parameters 3,096. Based on characteristic frequency, our three-level frequency sampling strategy is: 25% low-frequency components linearly distributed in $[1,48\pi]$ range, 50% characteristic frequencies using $\mathcal{N}(48\pi, 30^2)$ Gaussian distribution precisely covering dominant wavenumbers, 25% high-frequency components uniformly distributed in $[48\pi,96\pi]$ range to capture possible high-frequency effects. Training configuration uses 1,024 boundary points and 20,000 PDE collocation points, using Adam optimizer training 5,000 epochs.

	\begin{figure}[htbp]
		\centering
		\begin{subfigure}[b]{\textwidth}
			\centering
			\includegraphics[width=0.8\textwidth]{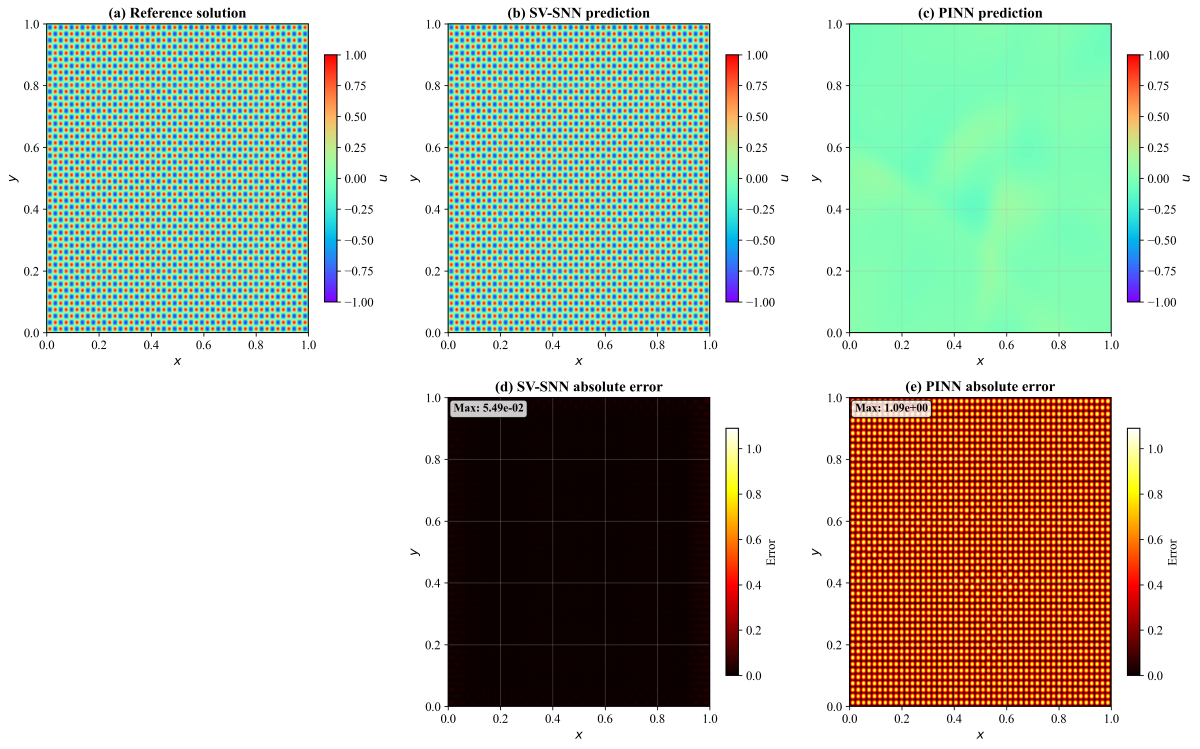}
			\caption{Prediction results and pointwise error distribution of SV-SNN and PINN}
			\label{fig:helmholtz48pi_solutions_errors}
		\end{subfigure}
		
		\begin{subfigure}[b]{\textwidth}
			\centering
			\includegraphics[width=0.8\textwidth]{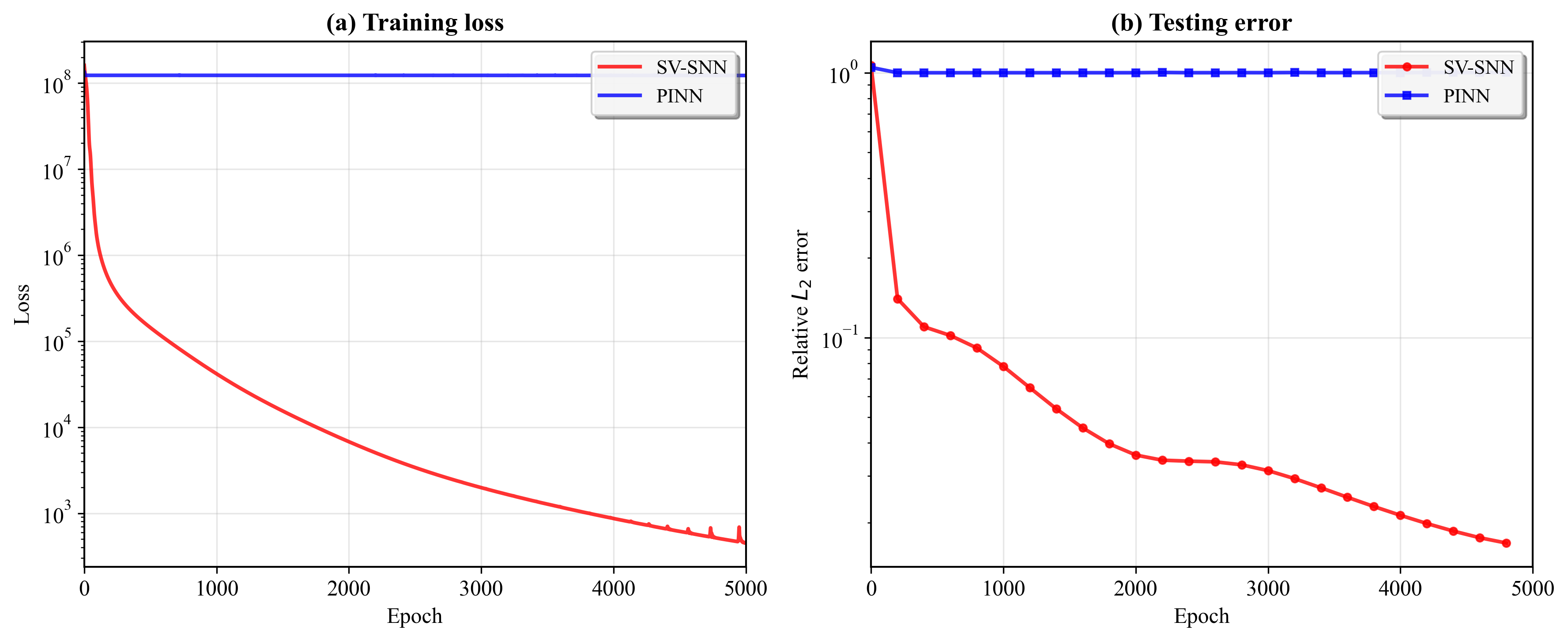}
			\caption{Training dynamics of SV-SNN and PINN, including training loss and test error}
			\label{fig:helmholtz48pi_training_dynamics}
		\end{subfigure}
		
		\begin{subfigure}[b]{\textwidth}
			\centering
			\includegraphics[width=0.8\textwidth]{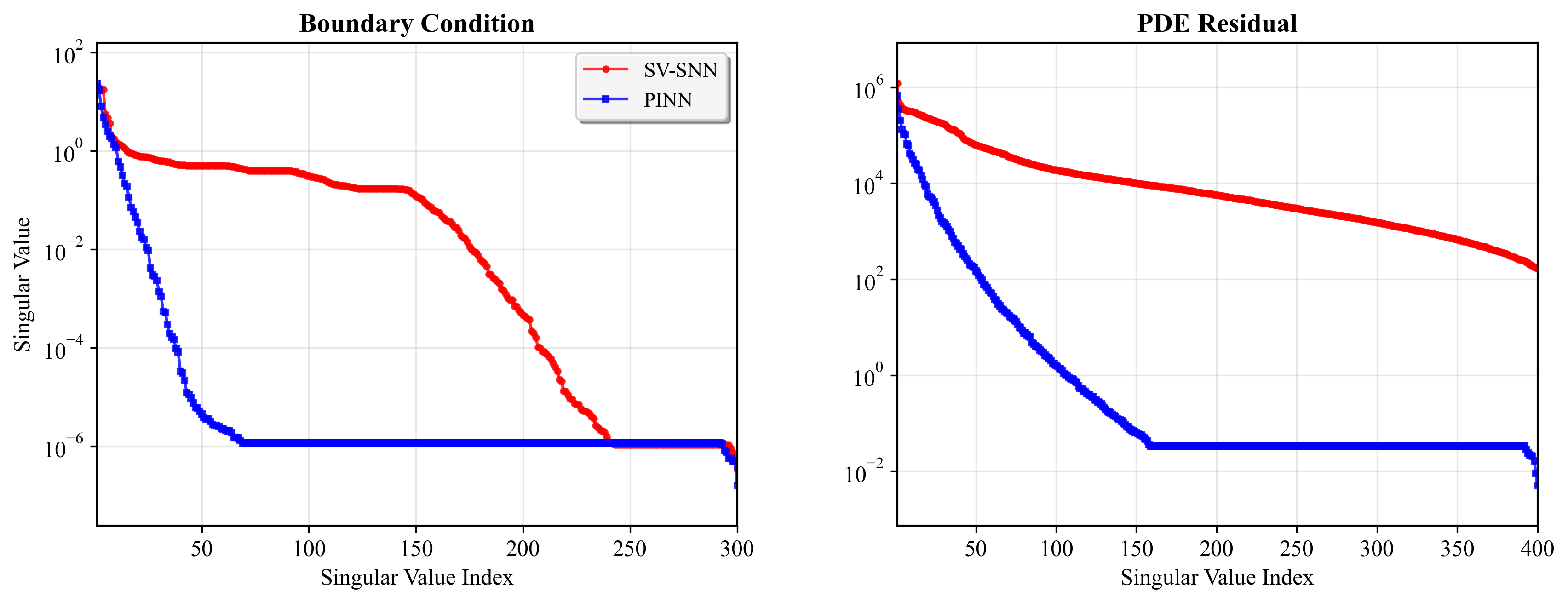}
			\caption{Singular value distributions of Jacobian matrices for SV-SNN and PINN}
			\label{fig:helmholtz48pi_svd_distributions}
		\end{subfigure}
		\caption{Two-dimensional Helmholtz equation ($\kappa = 48\pi$): Prediction performance, training dynamics, and singular value distributions of SV-SNN and PINN}
		\label{fig:helmholtz48pi_comprehensive}
	\end{figure}

	\begin{table}[htbp]
		\centering
		\caption{Performance comparison between SV-SNN and PINN methods on Helmholtz48$\pi$ problem}
		\label{tab:rank_comparison_helmholtz48pi}
		\resizebox{0.8\textwidth}{!}{%
		\begin{tabular}{lcccccccc}
		\toprule
		Method & \makecell{Total\\Parameters} & \makecell{$r_{\mathcal{B}}^{\text{eff}}$} & BC Loss & \makecell{$r_{\mathcal{F}}^{\text{eff}}$} & PDE Loss & \makecell{ReL2E} & \makecell{MAPE} \\
		\midrule
		SV-SNN & 3,096 & 35 & 2.43$\times$10$^{-3}$ & 66 & 6.12$\times$10$^{2}$ & 3.99$\times$10$^{-2}$ & 5.49$\times$10$^{-2}$ \\
		PINN & 58,561 & 7 & 6.37$\times$10$^{-4}$ & 9 & 1.23$\times$10$^{8}$ & 1.00$\times$10$^{0}$ & 1.09$\times$10$^{0}$ \\
		\bottomrule
		\end{tabular}}
	\end{table}

	Figure~\ref{fig:helmholtz48pi_comprehensive} and Table~\ref{tab:rank_comparison_helmholtz48pi} demonstrate SV-SNN's excellent performance on higher frequency wavenumber Helmholtz equation ($\kappa = 48\pi$). Compared to traditional PINN method using 58,561 parameters, SV-SNN requires only 3,096 parameters to achieve significant performance improvement: L2 relative error dramatically decreases from PINN's $1.00 \times 10^{0}$ to $3.99 \times 10^{-2}$, improving approximately 25 times, maximum pointwise absolute error decreases from $1.09 \times 10^{0}$ to $5.49 \times 10^{-2}$, improving approximately 20 times. We observe that SV-SNN has effective rank of 35 under boundary condition constraints and effective rank of 66 under PDE constraints, greatly exceeding PINN's 7 and 9, and SV-SNN's singular values under different constraint conditions are significantly larger than PINN's, indicating SV-SNN can still maintain strong expressive capability and gradient descent optimization capability even under very high-frequency oscillation conditions.

	\subsection{Nonlinear Elliptic Equations}
	
	To demonstrate SV-SNN's capability in handling nonlinear partial differential equations, we consider two-dimensional nonlinear elliptic equations. This class of problems has important applications in nonlinear optics, plasma physics, and materials science. Nonlinear terms make traditional numerical methods face greater challenges, especially when solution functions simultaneously possess high-frequency oscillation characteristics. The problem is defined on spatial region $\Omega = [0,1] \times [0,1]$, with governing equation:
	\begin{equation}
	\Delta u + u^2 = f(x,y)
	\end{equation}
	where $\Delta u = \frac{\partial^2 u}{\partial x^2} + \frac{\partial^2 u}{\partial y^2}$ is the two-dimensional Laplacian operator. Source term contains coupling of multiple frequency components, with specific form:
	\begin{align}
	f(x,y) &= -200(x + y)\cos(10x)\sin(10y) - 20\sin(10x)\sin(10y) \\
	&\quad - 20(x + y)\cos(10x)\cos(10y) + (x + y)^2\cos^2(10x)\sin^2(10y)
	\end{align}
	From source term, characteristic frequency $w_{char} = 10$ can be determined. Exact solution for this test case is:
	\begin{equation}
	u_{\text{exact}}(x,y) = (x + y) \cos(10x) \sin(10y)
	\end{equation}
	This solution has several important characteristics: spatial factor $(x + y)$ provides linearly growing amplitude, $\cos(10x)$ and $\sin(10y)$ produce high-frequency oscillations in $x$ and $y$ directions respectively.
	
	Boundary conditions adopt Dirichlet conditions, imposing exact solution values on all boundaries:
	\begin{equation}
	u(x,y) = u_{\text{exact}}(x,y), \quad (x,y) \in \partial\Omega
	\end{equation}
	Specific boundary values are:
	\begin{align}
	u(0,y) &= y \sin(10y), \quad u(1,y) = (1+y)\cos(10)\sin(10y) \\
	u(x,0) &= x \cos(10x), \quad u(x,1) = (x+1)\cos(10x)\sin(10)
	\end{align}

	Using two-dimensional spatial separation architecture, network modes $N=4$, each direction uses $K=32$ adaptive Fourier features, total parameters 780. We perform three-level sampling based on characteristic frequency, training configuration uses 1,024 boundary points and 10,000 PDE collocation points, using Adam optimizer training 5,000 epochs.

	\begin{figure}[htbp]
		\centering
		\begin{subfigure}{\textwidth}
			\centering
			\includegraphics[width=0.8\textwidth]{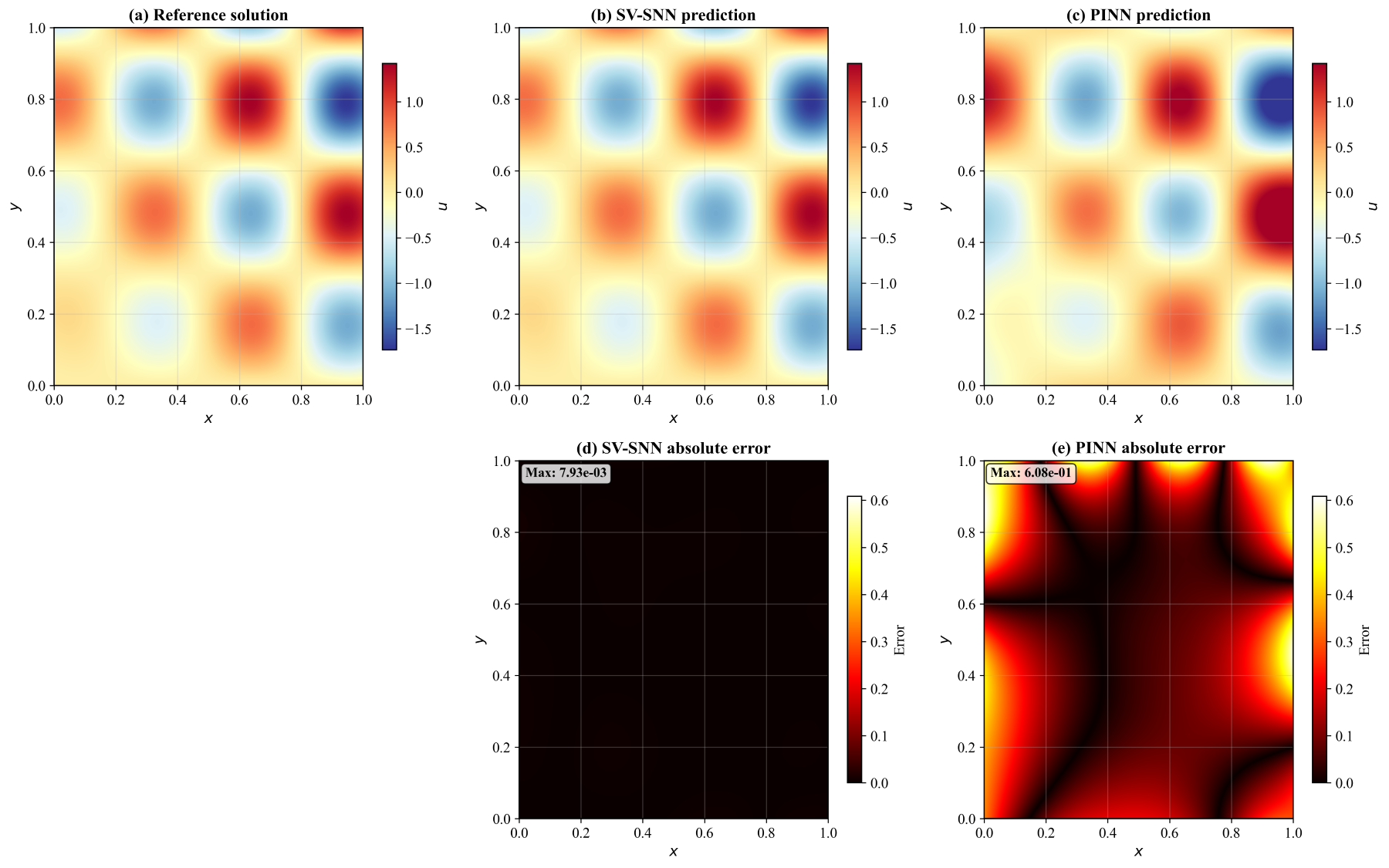}
			\caption{Prediction results and pointwise error distribution of SV-SNN and PINN}
			\label{fig:nonlinear_solutions_errors}
		\end{subfigure}
		
		\vspace{0.5cm}
		
		\begin{subfigure}{\textwidth}
			\centering
			\includegraphics[width=0.8\textwidth]{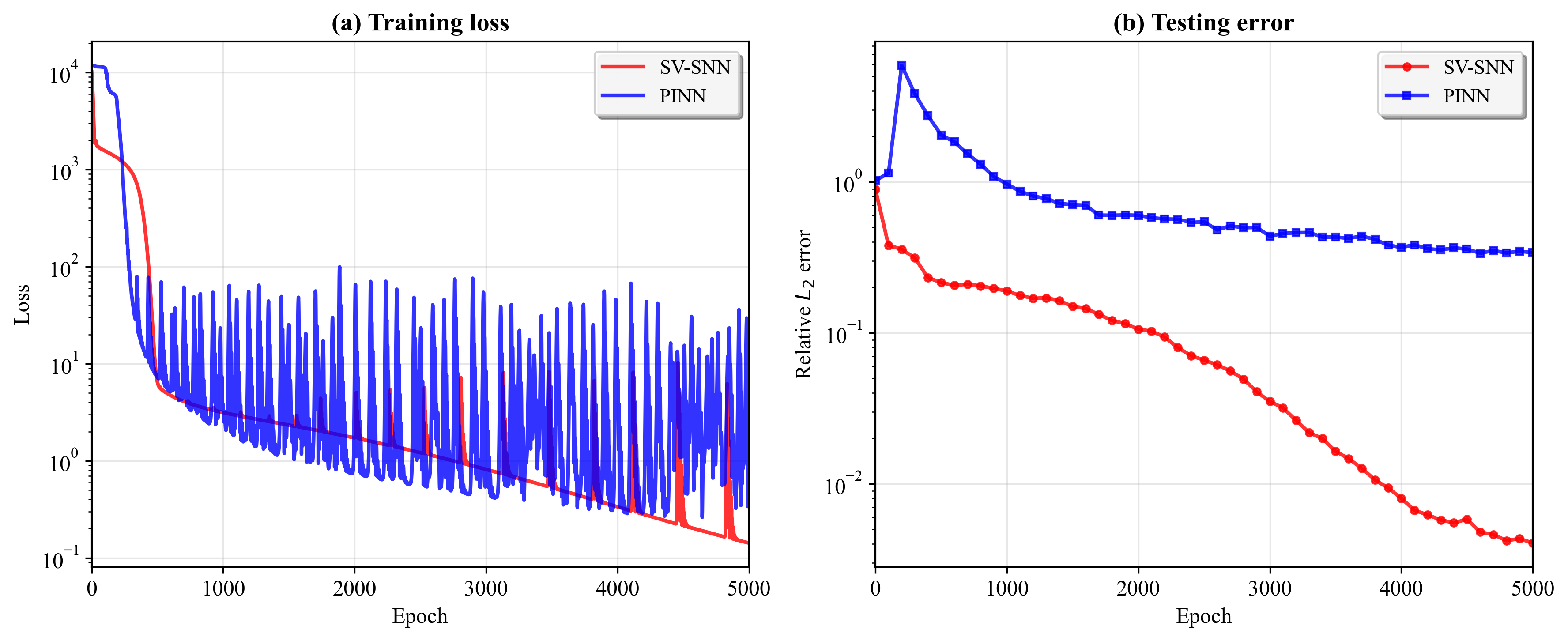}
			\caption{Training dynamics of SV-SNN and PINN, including training loss and test error}
			\label{fig:nonlinear_training_dynamics}
		\end{subfigure}
		
		\vspace{0.5cm}
		
		\begin{subfigure}{\textwidth}
			\centering
			\includegraphics[width=0.8\textwidth]{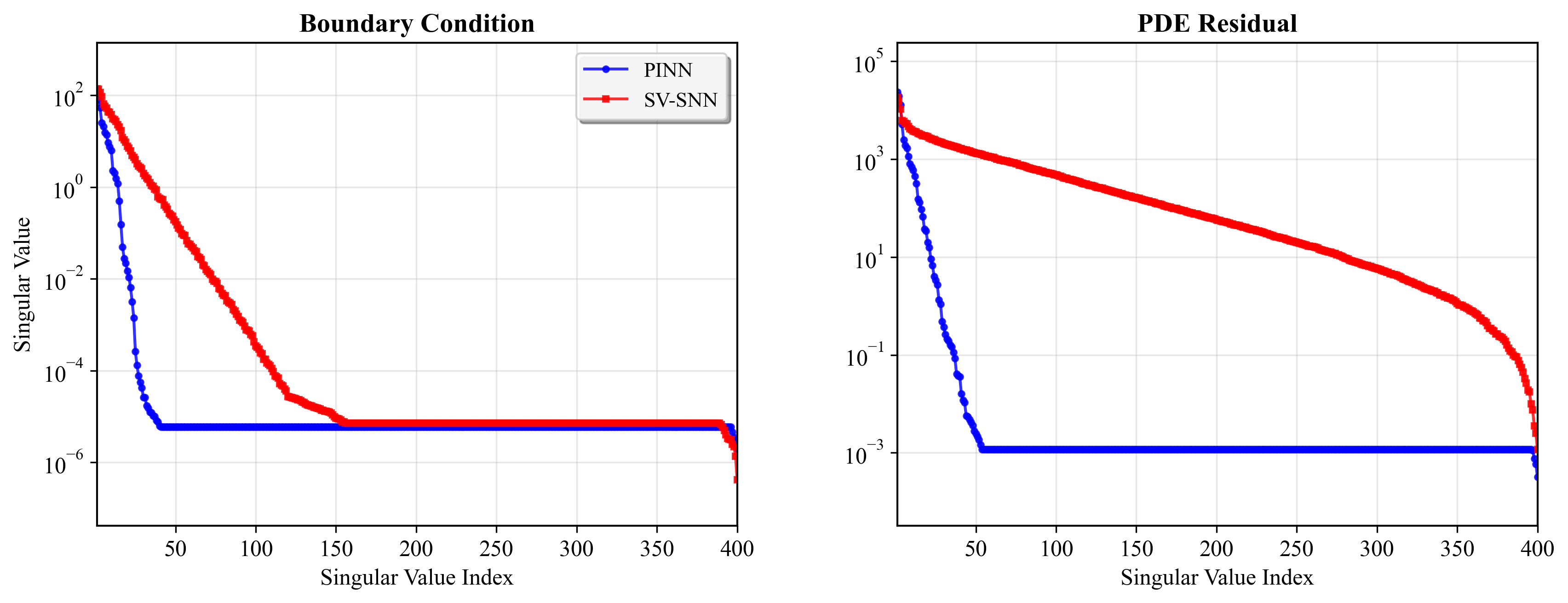}
			\caption{Singular value distributions of Jacobian matrices for SV-SNN and PINN}
			\label{fig:nonlinear_svd_distributions}
		\end{subfigure}
		
		\caption{Two-dimensional nonlinear elliptic equation solving results: Comparison of prediction performance, training dynamics, and singular value distributions between SV-SNN and PINN}
		\label{fig:nonlinear_combined}
	\end{figure}

	\begin{table}[htbp]
		\centering
		\caption{Performance comparison between SV-SNN and PINN on nonlinear elliptic equation problem}
		\label{tab:rank_comparison_nonlinear}
		\resizebox{0.8\textwidth}{!}{%
		\begin{tabular}{lcccccccc}
		\toprule
		Method & \makecell{Total\\Parameters} & \makecell{$r_{\mathcal{B}}^{\text{eff}}$} & BC Loss & \makecell{$r_{\mathcal{F}}^{\text{eff}}$} & PDE Loss & \makecell{ReL2E} & \makecell{MAPE} \\
		\midrule
		SV-SNN & 780 & 16 & 1.15$\times$10$^{-5}$ & 81 & 1.42$\times$10$^{-1}$ & 4.05$\times$10$^{-3}$ & 7.21$\times$10$^{-3}$ \\
		PINN & 40,801 & 7 & 1.28$\times$10$^{-1}$ & 5 & 3.81$\times$10$^{-1}$ & 3.40$\times$10$^{-1}$ & 5.91$\times$10$^{-1}$ \\
		\bottomrule
		\end{tabular}}
	\end{table}

	Figure~\ref{fig:nonlinear_combined} and Table~\ref{tab:rank_comparison_nonlinear} demonstrate SV-SNN's excellent performance on nonlinear elliptic equations. From quantitative results, SV-SNN achieves significant performance improvement using only 780 parameters: L2 relative error dramatically decreases from PINN's $3.40 \times 10^{-1}$ to $4.05 \times 10^{-3}$, improving approximately 84 times; maximum absolute error decreases from $5.91 \times 10^{-1}$ to $7.21 \times 10^{-3}$, improving approximately 82 times. Effective rank analysis shows SV-SNN has effective rank of 16 under boundary condition constraints and effective rank of 81 under PDE constraints, significantly exceeding PINN's 7 and 5, fully validating separated-variable spectral architecture's powerful expressive capability and function approximation characteristics when handling high-frequency nonlinear coupling problems.
	
	\subsection{Complex Geometry Poisson Equations}
	
	To further verify SV-SNN's effectiveness in complex geometric domains, we consider two-dimensional Poisson equations on complex domains with multiple internal holes. This class of problems widely exists in engineering practice, such as porous media flow, electromagnetic field analysis, and structural mechanics. The problem is defined on complex region $\Omega \subset [-1,1] \times [-1,1]$ with holes, with governing equation:
	\begin{equation}
	-\Delta u = f(x,y)
	\end{equation}
	where $\Delta u = \frac{\partial^2 u}{\partial x^2} + \frac{\partial^2 u}{\partial y^2}$ is the two-dimensional Laplacian operator, parameter $\mu = 7\pi$, source term:
	\begin{equation}
	f(x,y) = 2\mu^2 \sin(\mu x) \sin(\mu y)
	\end{equation}
	From source term, characteristic frequency $w_{char} = 7\pi$ can be determined. Analytical solution for this problem is:
	\begin{equation}
	u_{\text{exact}}(x,y) = \sin(\mu x) \sin(\mu y)
	\end{equation}
	Boundary conditions adopt Dirichlet conditions:
	\begin{equation}
	u(x,y) = u_{\text{exact}}(x,y), \quad (x,y) \in \partial\Omega
	\end{equation}
	Complex domain configuration includes external boundary (square $[-1,1]^2$) and multiple internal holes: domain contains three circular holes located at $(-0.5,-0.5)$ (radius 0.1), $(0.5,0.5)$ (radius 0.2), and $(0.5,-0.5)$ (radius 0.2), ellipse equation $16(x+0.5)^2 + 64(y-0.5)^2 = 1$.

	Using two-dimensional spatial separation architecture, network modes $N=8$, each direction uses $K=40$ adaptive Fourier features, SV-SNN total parameters 1,944, frequency sampling uses three-level sampling method, training collocation points use random sampling, including 400 external boundary points, 200 boundary points for each cylinder, 20,000 PDE collocation points, using Adam optimizer, training 5,000 epochs.
	
	\begin{figure}[htbp]
		\centering
		\begin{subfigure}{\textwidth}
			\centering
			\includegraphics[width=0.8\textwidth]{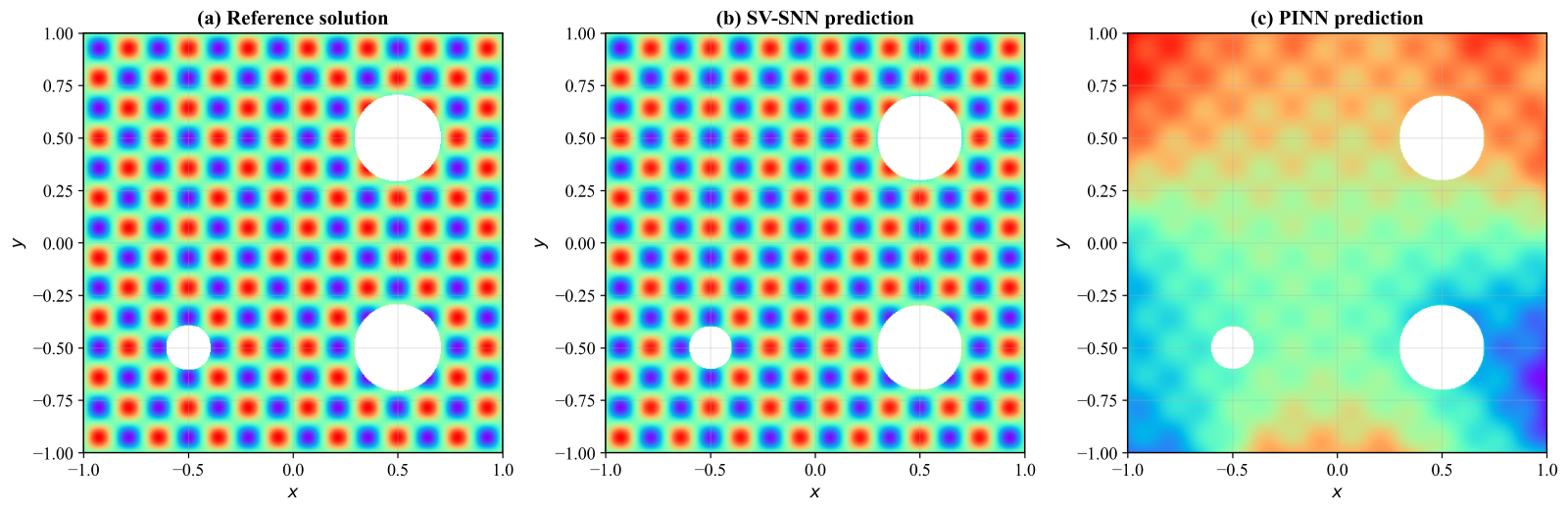}
			\caption{Prediction results and pointwise error distribution of SV-SNN and PINN}
			\label{fig:poisson_complex_solutions_errors}
		\end{subfigure}
		
		\vspace{0.5cm}
		
		\begin{subfigure}{\textwidth}
			\centering
			\includegraphics[width=0.8\textwidth]{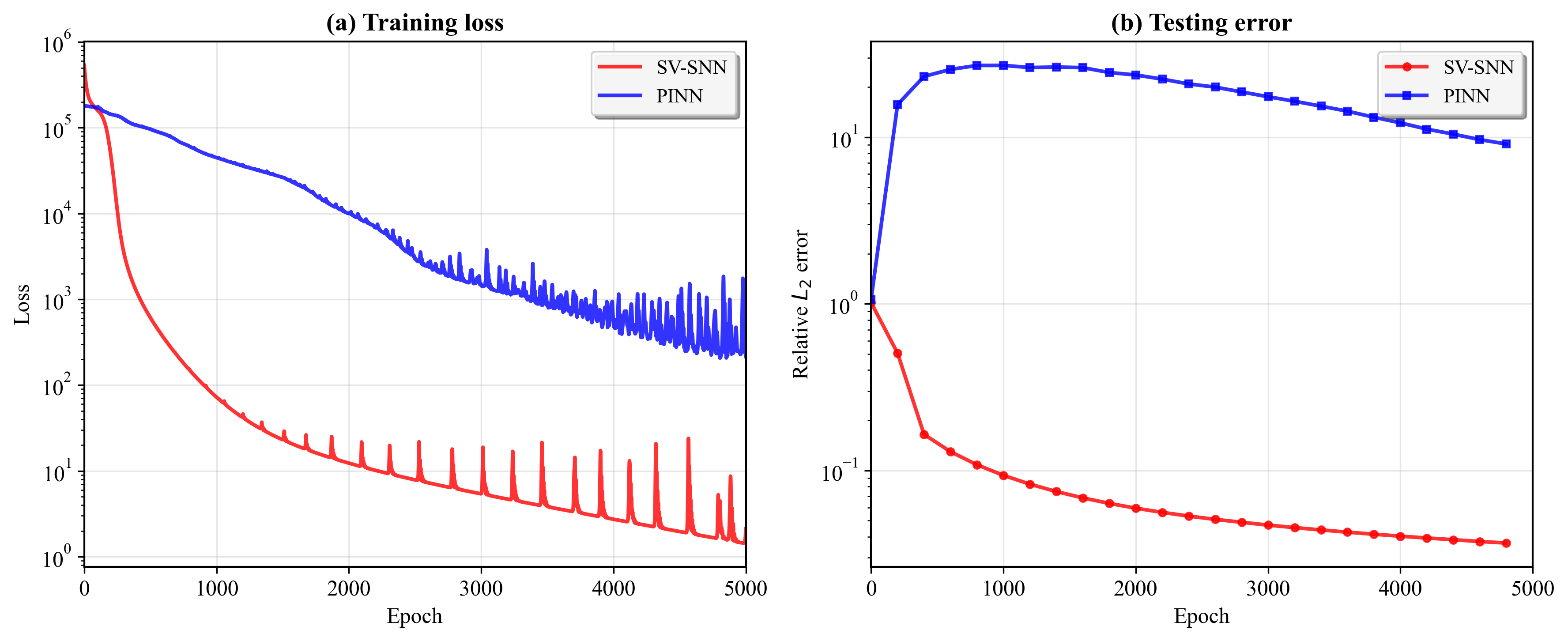}
			\caption{Training convergence curves and singular value distribution comparison}
			\label{fig:poisson_complex_training_dynamics}
		\end{subfigure}
		\caption{Complex geometry Poisson equation: Performance comparison analysis between SV-SNN and PINN}
		\label{fig:poisson_complex_combined}
	\end{figure}
	
	Figure~\ref{fig:poisson_complex_combined}(A) shows SV-SNN's prediction performance on complex geometry Poisson equations. SV-SNN can accurately capture high-frequency oscillation characteristics within complex geometric domains, with prediction results highly matching analytical solutions and prediction error of $3.45 \times 10^{-2}$, maintaining good approximation accuracy even in complex regions near hole boundaries. In contrast, PINN's prediction results show obvious errors throughout the entire region, with prediction error exceeding 1. Figure~\ref{fig:poisson_complex_combined}(B) training dynamic curves indicate SV-SNN exhibits fast and stable convergence characteristics from early training stages, with loss functions rapidly decreasing within the first 2000 epochs and remaining stable, while PINN's training process shows obvious convergence difficulties with very slow loss decrease, difficult to achieve ideal convergence states throughout training processes.

	\subsection{Complex Source Term Poisson Equations}
	
	In this section, we test SV-SNN's performance in handling high-frequency Poisson equations with complex source terms. We consider a two-dimensional Poisson equation with parameter $\mu = 15$, which has nonlinear source terms and high-frequency oscillation characteristics. The problem is defined on spatial region $\Omega = [-1,1] \times [-1,1]$, with governing equation:
	\begin{equation}
	-\Delta u = f(x,y)
	\end{equation}
	where $\Delta u = \frac{\partial^2 u}{\partial x^2} + \frac{\partial^2 u}{\partial y^2}$ is the two-dimensional Laplacian operator. The equation's source term is:
	\begin{align}
	f(x,y) &= 4\mu^2 x^2 \sin(\mu x^2) - 2\mu \cos(\mu x^2) + 4\mu^2 y^2 \sin(\mu y^2) - 2\mu \cos(\mu y^2)
	\end{align}
	This source term contains coupling of multiple frequency components: quadratic terms $x^2$ and $y^2$ provide spatially varying amplitude modulation, $\sin(\mu x^2)$ and $\sin(\mu y^2)$ produce nonlinear high-frequency oscillations, $\cos(\mu x^2)$ and $\cos(\mu y^2)$ provide phase-shifted frequency components. This complex nonlinear source term structure makes the problem challenging to solve, particularly requiring precise capture of solution's rapid changes in high-frequency regions.
	
	From the source term, characteristic frequency $w_{char} = 15$ can be determined. The exact solution for this test case is:
	\begin{equation}
	u_{\text{exact}}(x,y) = \sin(\mu x^2) + \sin(\mu y^2) = \sin(15x^2) + \sin(15y^2)
	\end{equation}
	This solution exhibits complex high-frequency oscillation patterns in space, with frequency characteristics related to spatial position, reaching maximum frequency near domain boundaries.

	Boundary conditions adopt Dirichlet conditions, imposing exact solution values on all boundaries:
	\begin{equation}
	u(x,y) = u_{\text{exact}}(x,y), \quad (x,y) \in \partial\Omega
	\end{equation}
	Specific boundary values on four boundaries are:
	\begin{align}
	u(-1,y) &= \sin(15) + \sin(15y^2), \quad u(1,y) = \sin(15) + \sin(15y^2) \\
	u(x,-1) &= \sin(15x^2) + \sin(15), \quad u(x,1) = \sin(15x^2) + \sin(15)
	\end{align}
	
	Using two-dimensional spatial separation architecture, network modes $N=4$, each direction uses $K=50$ adaptive Fourier features, total parameters 1,212. Based on characteristic frequency, our three-level frequency sampling strategy is: 25% low-frequency components linearly distributed in $[1,15]$ range, 50% characteristic frequencies using $\mathcal{N}(15, 5^2)$ Gaussian distribution precisely covering dominant wavenumbers, 25% high-frequency components uniformly distributed in $[15,30]$ range to capture possible high-frequency effects. Boundary condition points include 1,024 points (uniformly distributed on four boundaries, ensuring accurate capture of high-frequency boundary characteristics), PDE collocation points are 20,000. Using Adam algorithm for optimization, training 40,000 epochs to fully learn high-frequency nonlinear characteristics.
	
	\begin{figure}[htbp]
		\centering
		\begin{subfigure}{\textwidth}
			\centering
			\includegraphics[width=0.8\textwidth]{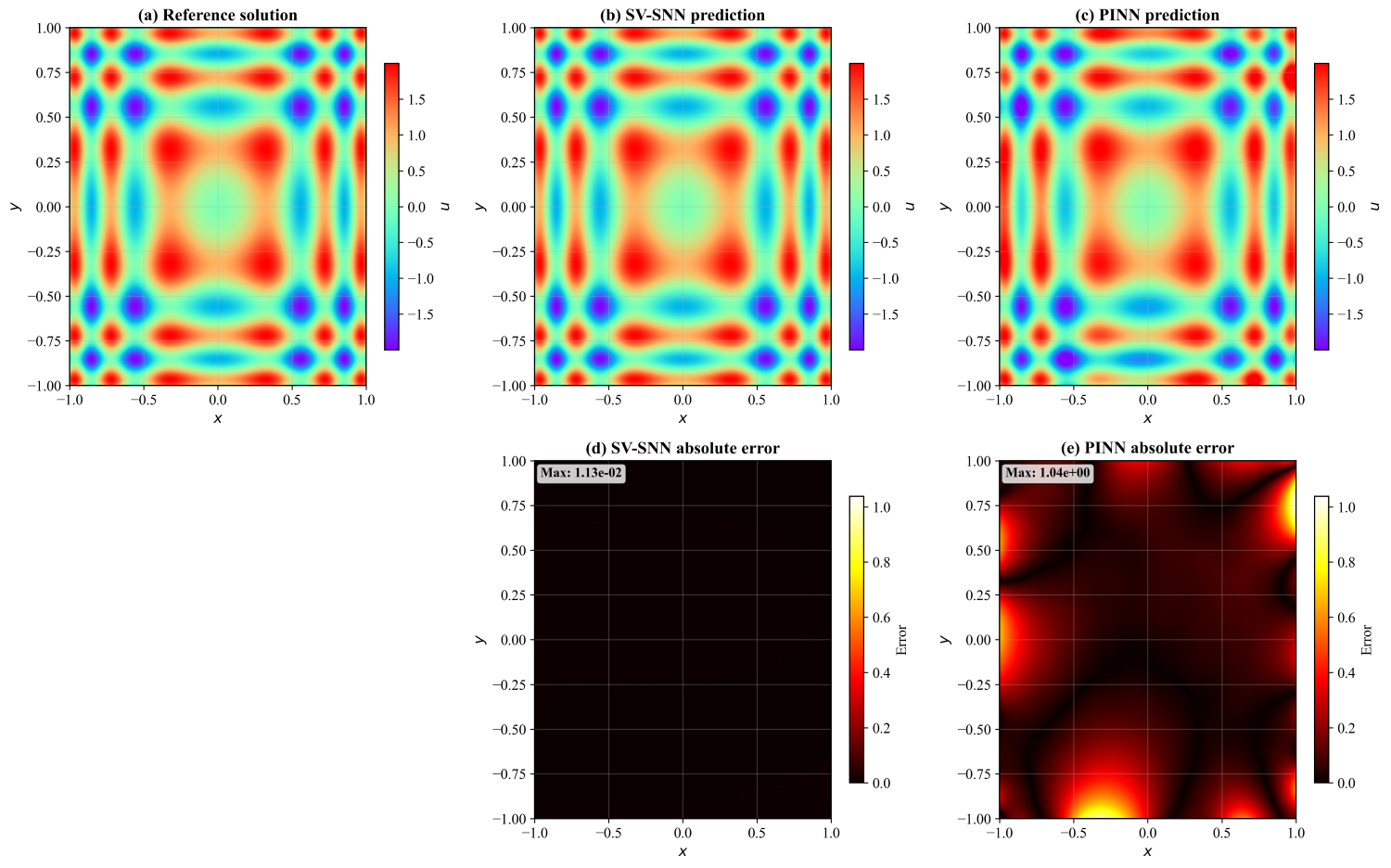}
			\caption{Prediction results and pointwise error distribution of SV-SNN and PINN}
			\label{fig:poisson15_solutions_errors}
		\end{subfigure}
		
		\vspace{0.5cm}
		
		\begin{subfigure}{\textwidth}
			\centering
			\includegraphics[width=0.8\textwidth]{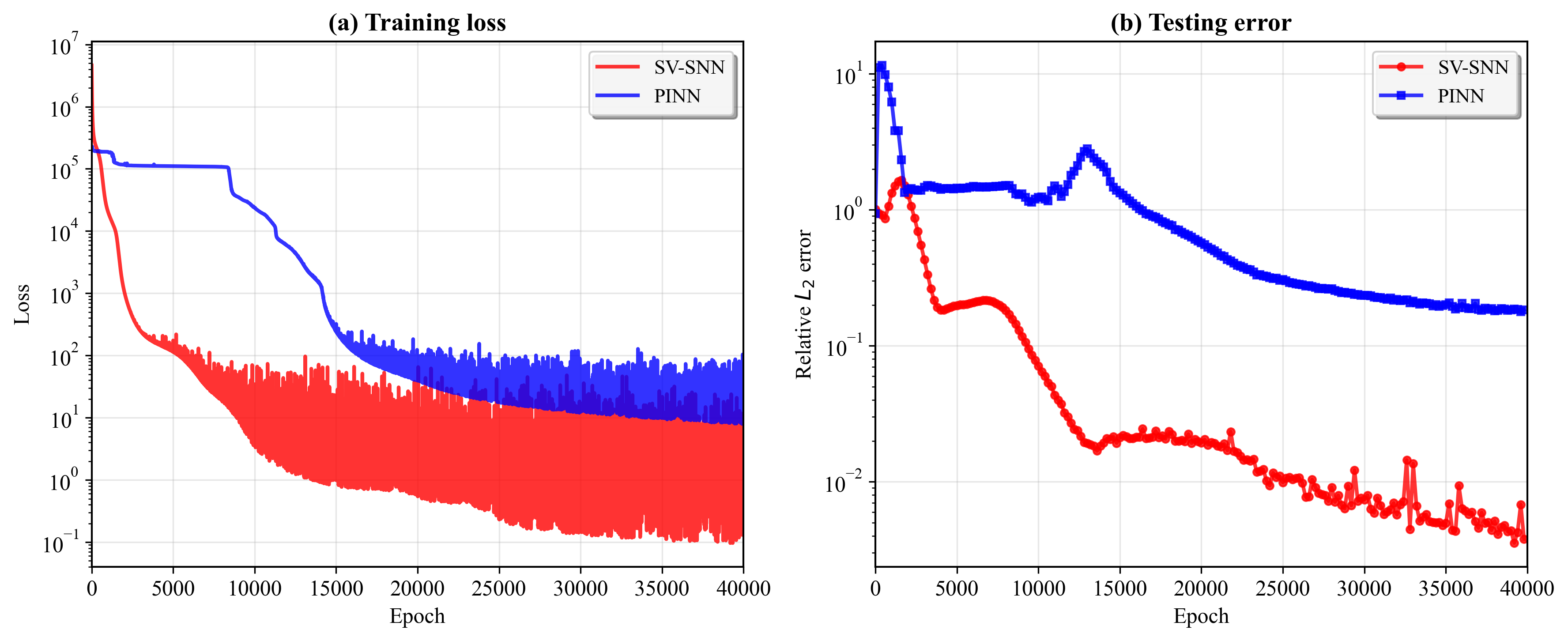}
			\caption{Training dynamics of SV-SNN and PINN, including training loss and test error}
			\label{fig:poisson15_training_dynamics}
		\end{subfigure}
		
		\vspace{0.5cm}
		
		\begin{subfigure}{\textwidth}
			\centering
			\includegraphics[width=0.8\textwidth]{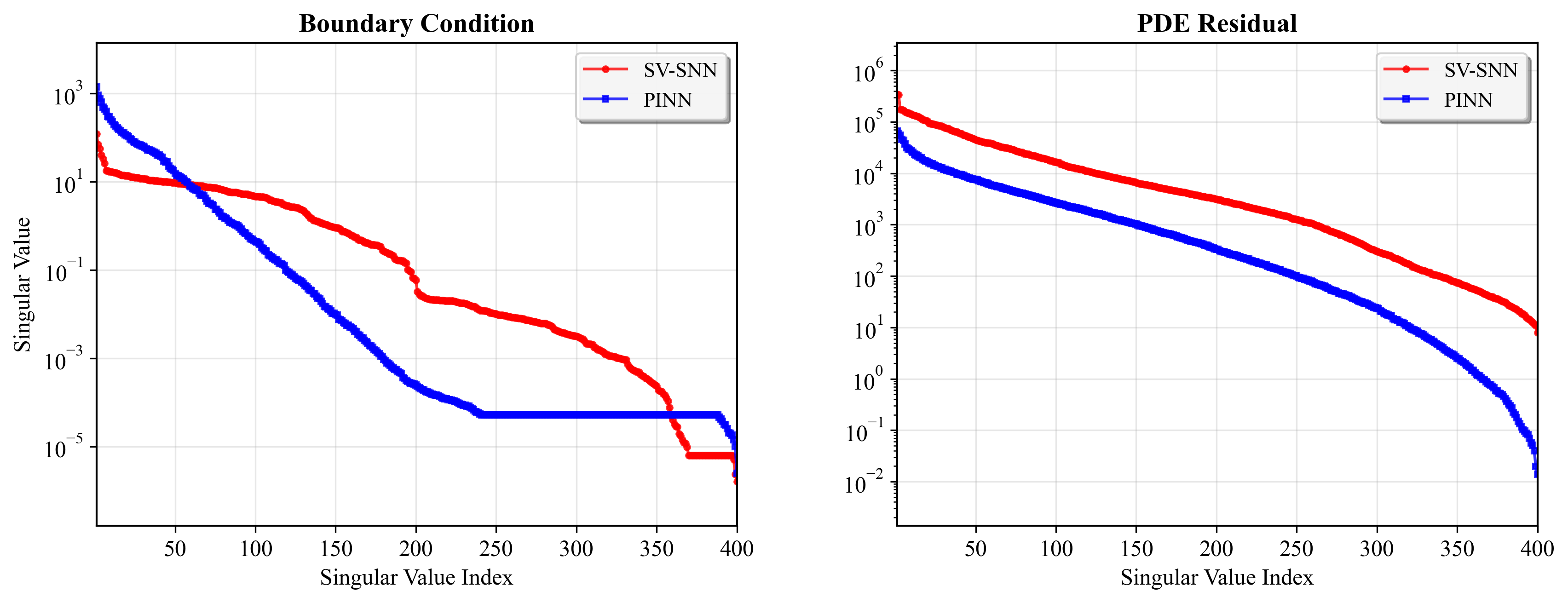}
			\caption{Singular value distributions of Jacobian matrices for SV-SNN and PINN}
			\label{fig:poisson15_svd_distributions}
		\end{subfigure}

		\caption{Complex source term Poisson equation: Prediction performance, training dynamics, and singular value distributions of SV-SNN and PINN}
		\label{fig:poisson15_combined}
	\end{figure}

	\begin{table}[htbp]
		\centering
		\caption{Performance comparison between SV-SNN and PINN on complex source term Poisson equation problem}
		\label{tab:rank_comparison_poisson}
		\resizebox{0.8\textwidth}{!}{%
		\begin{tabular}{lcccccccc}
		\toprule
		Method & \makecell{Total\\Parameters} & \makecell{$r_{\mathcal{B}}^{\text{eff}}$} & BC Loss & \makecell{$r_{\mathcal{F}}^{\text{eff}}$} & PDE Loss & \makecell{ReL2E} & \makecell{MAPE} \\
		\midrule
		SV-SNN & 1,212 & 103 & 6.36$\times$10$^{-3}$ & 98 & 3.29$\times$10$^{0}$ & 4.73$\times$10$^{-3}$ & 1.13$\times$10$^{-2}$ \\
		PINN & 20,601 & 27 & 1.68$\times$10$^{0}$ & 88 & 9.61$\times$10$^{0}$ & 1.78$\times$10$^{-1}$ & 1.04$\times$10$^{0}$ \\
		\bottomrule
		\end{tabular}}
	\end{table}

	Table~\ref{tab:rank_comparison_poisson} shows performance comparison results between SV-SNN and traditional PINN methods when solving high-frequency Poisson equations, fully demonstrating significant advantages of separated-variable spectral neural networks. Despite SV-SNN using only 813 parameters, compared to PINN's 20,601 parameters with approximately 96% reduction, it achieves qualitative leaps in solution accuracy: L2 relative error decreases from PINN's $1.78 \times 10^{-1}$ to $4.73 \times 10^{-3}$, improving nearly two orders of magnitude, maximum pointwise error decreases from $1.04 \times 10^{0}$ to $1.13 \times 10^{-2}$, improving approximately 90 times. More importantly, from effective rank analysis, SV-SNN achieves effective rank of 103 under boundary condition constraints, much higher than PINN's 27, indicating its stronger expressive capability and better function approximation characteristics. This efficient parameter utilization rate and excellent solution accuracy fully validate theoretical advantages and practical effects of Fourier spectral features and separated-variable architecture when handling high-frequency oscillation problems.

	\subsection{Taylor-Green Vortex}
	
	To verify SV-SNN's effectiveness in complex nonlinear fluid dynamics simulation, we consider the classical Taylor-Green vortex problem, which is a benchmark test case for incompressible Navier-Stokes equations. The problem is defined on spatiotemporal domain (two-dimensional space + time) $\Omega = [-\pi,\pi] \times [-\pi,\pi] \times [0,1]$, with governing equations being incompressible Navier-Stokes equation system:
	\begin{align}
	\frac{\partial u}{\partial t} + u \frac{\partial u}{\partial x} + v \frac{\partial u}{\partial y} + \frac{\partial p}{\partial x} - \frac{1}{\text{Re}}\left(\frac{\partial^2 u}{\partial x^2} + \frac{\partial^2 u}{\partial y^2}\right) &= 0 \label{eq:ns_u} \\
	\frac{\partial v}{\partial t} + u \frac{\partial v}{\partial x} + v \frac{\partial v}{\partial y} + \frac{\partial p}{\partial y} - \frac{1}{\text{Re}}\left(\frac{\partial^2 v}{\partial x^2} + \frac{\partial^2 v}{\partial y^2}\right) &= 0 \label{eq:ns_v} \\
	\frac{\partial u}{\partial x} + \frac{\partial v}{\partial y} &= 0 \label{eq:continuity}
	\end{align}
	where $(u,v)$ are velocity components, $p$ is pressure, and $\text{Re}$ is Reynolds number. Boundary conditions are periodic:
	\begin{align}
	u(x+2\pi, y, t) &= u(x, y, t), \quad v(x+2\pi, y, t) = v(x, y, t) \\
	u(x, y+2\pi, t) &= u(x, y, t), \quad v(x, y+2\pi, t) = v(x, y, t)
	\end{align}
	
	Initial conditions are:
	\begin{align}
	u(x,y,0) &= -\cos(\pi x)\sin(\pi y) \\
	v(x,y,0) &= \sin(\pi x)\cos(\pi y) \\
	p(x,y,0) &= -\frac{1}{4}[\cos(2\pi x) + \cos(2\pi y)]
	\end{align}

	We adopt spatiotemporal separation architecture:
	\begin{align}
	u^{\Theta}(x,y,t) &= \sum_{n=1}^{N} c_n^{(u)} X_n^{(F)}(x) Y_n^{(F)}(y) T_n^{(u)}(t) \\
	v^{\Theta}(x,y,t) &= \sum_{n=1}^{N} c_n^{(v)} X_n^{(F)}(x) Y_n^{(F)}(y) T_n^{(v)}(t) \\
	p^{\Theta}(x,y,t) &= \sum_{n=1}^{N} c_n^{(p)} X_n^{(F)}(x) Y_n^{(F)}(y) T_n^{(p)}(t)
	\end{align}
	where spatial parts $X_n^{(F)}(x)$ and $Y_n^{(F)}(y)$ are Fourier features for $x$ and $y$ directions respectively, temporal networks $T_n^{(u)}(t)$, $T_n^{(v)}(t)$, and $T_n^{(p)}(t)$ are standard fully-connected networks with tanh activation functions.

	Spatial directions adopt two-dimensional separation architecture, network modes $N=6$, each direction uses $K=32$ adaptive Fourier features, temporal direction uses fully-connected networks with 4 layers of 10 neurons each, total parameters 2,688. Initial condition sampling is achieved by constructing $100 \times 100$ uniform grids in spatial domain $[-\pi, \pi] \times [-\pi, \pi]$, obtaining 10,000 initial condition points at $t=0$. Boundary condition and PDE residual collocation point sampling both use Latin Hypercube Sampling method, generating 2,000 boundary constraint points and 10,000 internal collocation points in three-dimensional spatiotemporal domain respectively. Using Adam algorithm for optimization, training 5,000 epochs.

	\begin{table}[htbp]
		\centering
		\caption{Performance comparison between SV-SNN and PINN for solving Taylor-Green vortex}
		\label{tab:taylor_green_comparison}
		\resizebox{0.8\textwidth}{!}{%
		\begin{tabular}{lccccc}
		\toprule
		Method & \makecell{Total\\Parameters} & \makecell{Final Loss} & \makecell{$u$ Component\\Relative $l_2$ Error} & \makecell{$v$ Component\\Relative $l_2$ Error} & \makecell{$p$ Component\\Relative $l_2$ Error} \\
		\midrule
		SV-SNN & 2,688 & 4.54$\times$10$^{-5}$ & 1.52$\times$10$^{-3}$ & 1.83$\times$10$^{-3}$ & 3.20$\times$10$^{-3}$ \\
		PINN & 41,103 & 2.93$\times$10$^{-1}$ & 6.85$\times$10$^{-1}$ & 6.84$\times$10$^{-1}$ & 1.67 \\
		\bottomrule
		\end{tabular}}
	\end{table}

\begin{figure}[htbp]
		\centering
		\begin{subfigure}[b]{0.8\textwidth}
			\centering
			\includegraphics[width=\textwidth]{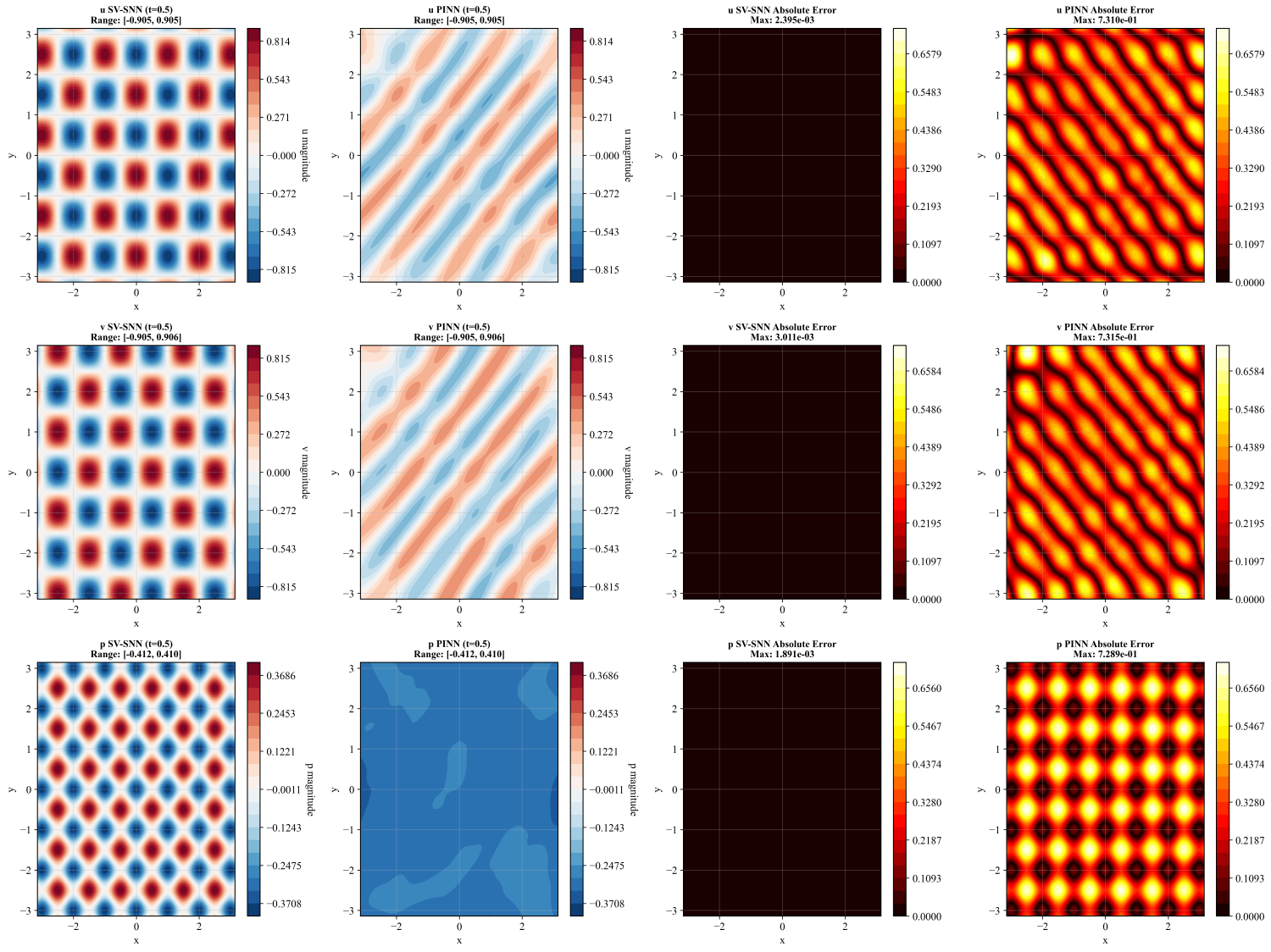}
			\caption{Prediction results and pointwise error distribution of SV-SNN and PINN}
		\end{subfigure}
		
		\begin{subfigure}[b]{0.8\textwidth}
			\centering
			\includegraphics[width=\textwidth]{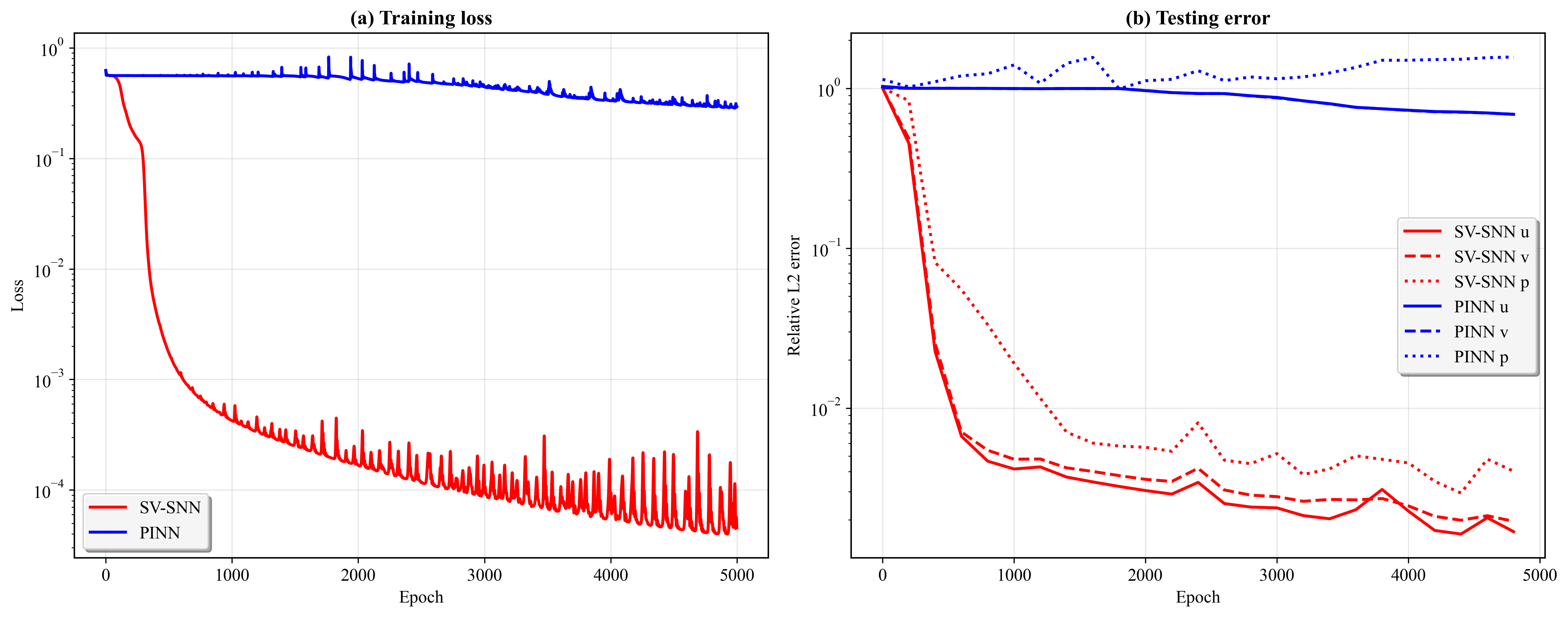}
			\caption{Training dynamics of SV-SNN and PINN, including training loss and test error}
		\end{subfigure}
		\caption{Taylor-Green vortex: Prediction results and training dynamics of SV-SNN and PINN}
		\label{fig:tgns_comparison}
	\end{figure}

	Experimental results in Table~\ref{tab:taylor_green_comparison} comprehensively demonstrate SV-SNN's excellent performance in solving complex nonlinear Taylor-Green vortex problems, with specific prediction effects shown in Figure~\ref{fig:tgns_comparison}. In parameter efficiency, SV-SNN achieves high-precision fluid dynamics numerical simulation using only 2,688 parameters, compared to traditional PINN requiring 41,103 parameters, reducing parameter scale by 93.5%. This significant parameter efficiency improvement mainly benefits from spatiotemporal separation architecture effectively decomposing complexity of high-dimensional nonlinear systems, and adaptive Fourier spectral features' compact representation capability for multi-scale flow field structures. In convergence performance, from final loss function values, SV-SNN reaches $4.54 \times 10^{-5}$, compared to PINN's $2.93 \times 10^{-1}$ with approximately 4 orders of magnitude improvement. This enormous difference indicates SV-SNN can more effectively satisfy physical constraint conditions of Navier-Stokes equation system, including momentum conservation and continuity equation constraints. In approximation accuracy for fluid components, SV-SNN shows significant advantages, with $u$ velocity component relative L2 error of $1.52 \times 10^{-3}$, $v$ velocity component error of $1.83 \times 10^{-3}$, pressure component error of $3.20 \times 10^{-3}$, while PINN's corresponding errors reach $6.85 \times 10^{-1}$, $6.84 \times 10^{-1}$, and $1.67$ respectively, representing accuracy improvement over 2 orders of magnitude. This accuracy advantage mainly results from spatiotemporal separation architecture and adaptive Fourier spectral features effectively capturing periodicity and multi-scale characteristics of vortex structures, with separated-variable representation avoiding optimization difficulties in high-dimensional parameter spaces.

	\subsection{Double-Cylinder Steady Navier-Stokes Equations}
	
	To further verify robustness of separated-variable spectral neural networks in complex geometric domains and multi-obstacle fluid mechanics problems, we consider steady Navier-Stokes equations with two cylindrical obstacles in a circular domain. The problem is defined within circular domain $\Omega = \{(x,y) : x^2 + y^2 \leq 3.0^2\} \setminus (\Omega_1 \cup \Omega_2)$, with two cylindrical obstacles: Cylinder 1 centered at $(-1.0, 0.5)$ with radius $r_1 = 0.3$; Cylinder 2 centered at $(1.0, -0.5)$ with radius $r_2 = 0.3$. Governing equations are steady incompressible Navier-Stokes equation system:
	\begin{align}
	u \frac{\partial u}{\partial x} + v \frac{\partial u}{\partial y} + \frac{1}{\rho}\frac{\partial p}{\partial x} - \frac{\mu}{\rho}\nabla^2 u &= S_x(x,y) \\
	u \frac{\partial v}{\partial x} + v \frac{\partial v}{\partial y} + \frac{1}{\rho}\frac{\partial p}{\partial y} - \frac{\mu}{\rho}\nabla^2 v &= S_y(x,y) \\
	\frac{\partial u}{\partial x} + \frac{\partial v}{\partial y} &= 0
	\end{align}
	where $(u,v)$ are velocity components, $p$ is pressure, $\rho = 1.0$ is density, $\mu = 1.0$ is dynamic viscosity.

	Source terms in this test case are:
	\begin{align}
	S_x(x,y) &= \sin(4x) - 0.25\sin(x - 3y) + \sin(x + y) + 8\sin(2x)\cos(2y) + 0.75\sin(3x - y) \\
	S_y(x,y) &= \sin(4y) - 0.75\sin(x - 3y) - \sin(x + y) - 8\cos(2x)\sin(2y) + 0.25\sin(3x - y)
	\end{align}
	
	Boundary conditions impose exact solution Dirichlet conditions on all boundaries.
	
	Using spatiotemporal separation architecture:
	\begin{align}
	u^{\Theta}(x,y) &= \sum_{n=1}^{N} c_n^{(u)} X_n^{(F)}(x) Y_n^{(F)}(y) \\
	v^{\Theta}(x,y) &= \sum_{n=1}^{N} c_n^{(v)} X_n^{(F)}(x) Y_n^{(F)}(y) \\
	p^{\Theta}(x,y) &= \sum_{n=1}^{N} c_n^{(p)} X_n^{(F)}(x) Y_n^{(F)}(y)
	\end{align}
	where spatial parts $X_n^{(F)}(x)$ and $Y_n^{(F)}(y)$ are Fourier features for $x$ and $y$ directions respectively. Spatial directions adopt two-dimensional separation architecture, network modes $N=4$, each spatial direction uses $K=16$ adaptive Fourier features, total parameters 404. We uniformly sample 400 Dirichlet boundary points on external circular boundary, and uniformly sample 100 no-slip boundary condition points on each cylindrical obstacle boundary. For PDE collocation point sampling, we first perform Latin Hypercube random sampling within the region, then remove points inside obstacles, retaining points in valid computational region, finally obtaining 20,000 PDE collocation points for physical loss constraints, while generating 15,000 test points for error evaluation. Optimization algorithm uses Adam, training iterations 15,000 epochs.

	\begin{figure}[htbp]
		\centering
		\begin{subfigure}[b]{0.8\textwidth}
			\centering
			\includegraphics[width=\textwidth]{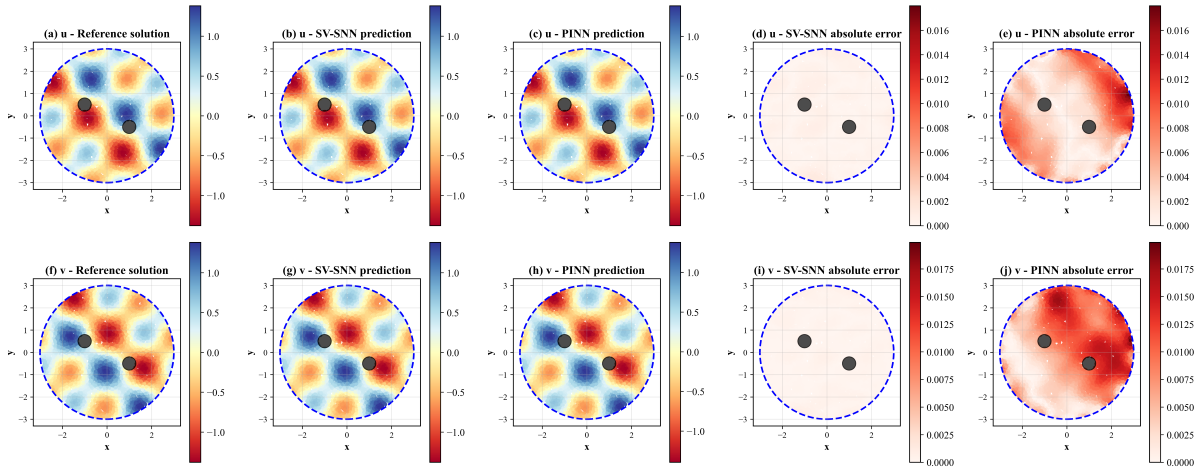}
			\caption{Prediction performance of SV-SNN and PINN: prediction results and pointwise error distribution of velocity fields $u$ and $v$}
		\end{subfigure}
		
		\begin{subfigure}[b]{0.8\textwidth}
			\centering
			\includegraphics[width=\textwidth]{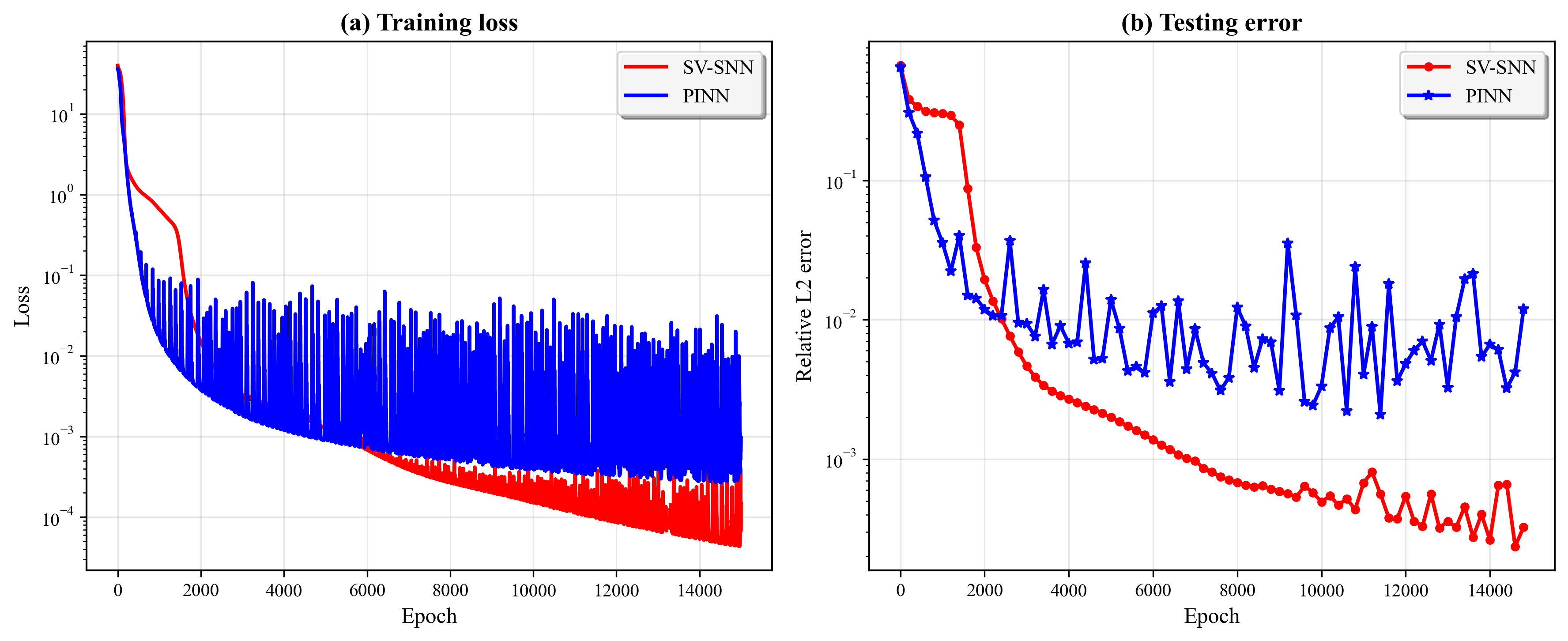}
			\caption{Training dynamics of SV-SNN and PINN: convergence curves of training loss and test error}
		\end{subfigure}
		\caption{Double-cylinder steady Navier-Stokes equations: Prediction results and training dynamics of SV-SNN and PINN}
		\label{fig:navier_stokes_cylinder}
	\end{figure}

	\begin{table}[htbp]
		\centering
		\caption{Performance comparison between SV-SNN and PINN for solving double-cylinder steady Navier-Stokes equations}
		\label{tab:ns_cylinder_comparison}
		\resizebox{0.5\textwidth}{!}{%
		\begin{tabular}{lcccc}
		\toprule
		Method & \makecell{Total\\Parameters} & \makecell{Total Loss} & \makecell{ReL2E $u$} & \makecell{ReL2E $v$} \\
		\midrule
		SV-SNN & 404 & 6.97$\times$10$^{-5}$ & 5.68$\times$10$^{-4}$ & 4.06$\times$10$^{-4}$ \\
		PINN & 7,953 & 9.69$\times$10$^{-4}$ & 9.45$\times$10$^{-3}$ & 1.50$\times$10$^{-2}$ \\
		\bottomrule
		\end{tabular}}
	\end{table}

	Table~\ref{tab:ns_cylinder_comparison} shows performance comparison results between SV-SNN and traditional PINN methods for solving Navier-Stokes equations under complex boundary conditions. From quantitative analysis, SV-SNN significantly outperforms PINN method across multiple key indicators. First, in parameter efficiency, SV-SNN uses only 404 parameters, compared to PINN's 7,953 parameters with 94.9% reduction, demonstrating significant dimensional reduction advantages of separated-variable architecture. Second, in solution accuracy, SV-SNN's total loss is $6.97 \times 10^{-5}$, approximately 14 times lower than PINN's $9.69 \times 10^{-4}$. More importantly, in relative $L^2$ errors of velocity components, SV-SNN's $u$ component error is $5.68 \times 10^{-4}$, approximately 16.6 times better than PINN's $9.45 \times 10^{-3}$; $v$ component error is $4.06 \times 10^{-4}$, approximately 36.9 times better than PINN's $1.50 \times 10^{-2}$. These results fully demonstrate SV-SNN's excellent performance when handling complex geometric domain multi-physics coupling problems, achieving dual optimization of parameter efficiency and solution accuracy.

	\section{Ablation Studies}
	
	To illustrate characteristics of main components in SV-SNN, we design ablation experiments aimed at verifying effective ranges and limitations of key SV-SNN components, including effects of mode numbers on representation capability, effects of frequency sampling strategies on prediction performance, etc. Additionally, we conduct comparative analysis with various advanced methods, demonstrating SV-SNN's advantages in training time, solution accuracy, and other aspects.

	\subsection{SV-SNN Component Ablation Studies}

	\subsubsection{Analysis of Mode Number Effects}

	Network mode number $N$ is an important parameter in SV-SNN, largely determining the network's expressive capability for complex functions. We first test effects of network mode numbers on SV-SNN prediction performance, taking $N \in \{1, 4, 7, 10\}$.
	
	\begin{figure}[htbp]
		\centering
		\begin{subfigure}[b]{1.0\textwidth}
			\centering
			\includegraphics[width=\textwidth]{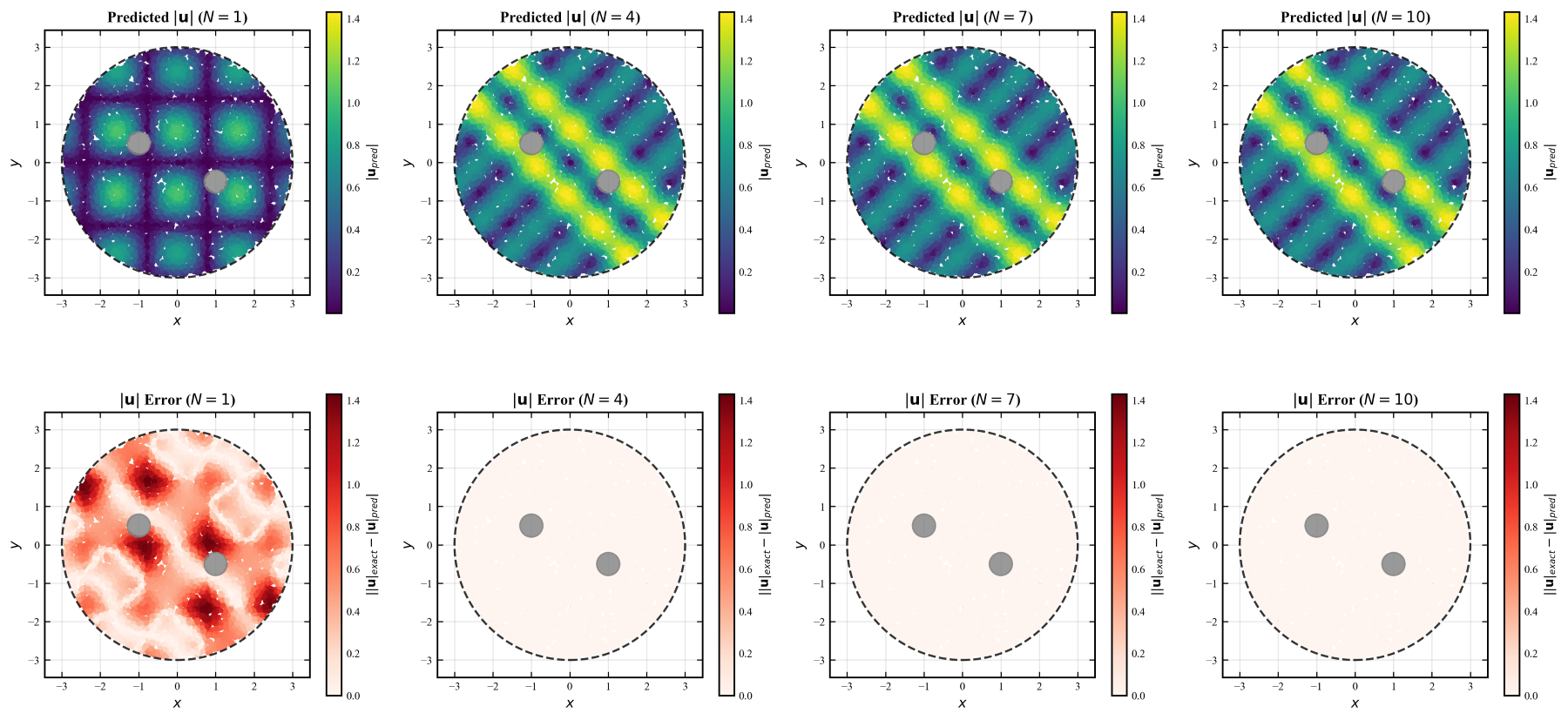}
			\caption{SV-SNN prediction performance under different network modes $N$: prediction results and pointwise error distribution of velocity fields $u$ and $v$}
		\end{subfigure}
		
		\begin{subfigure}[b]{1.0\textwidth}
			\centering
			\includegraphics[width=\textwidth]{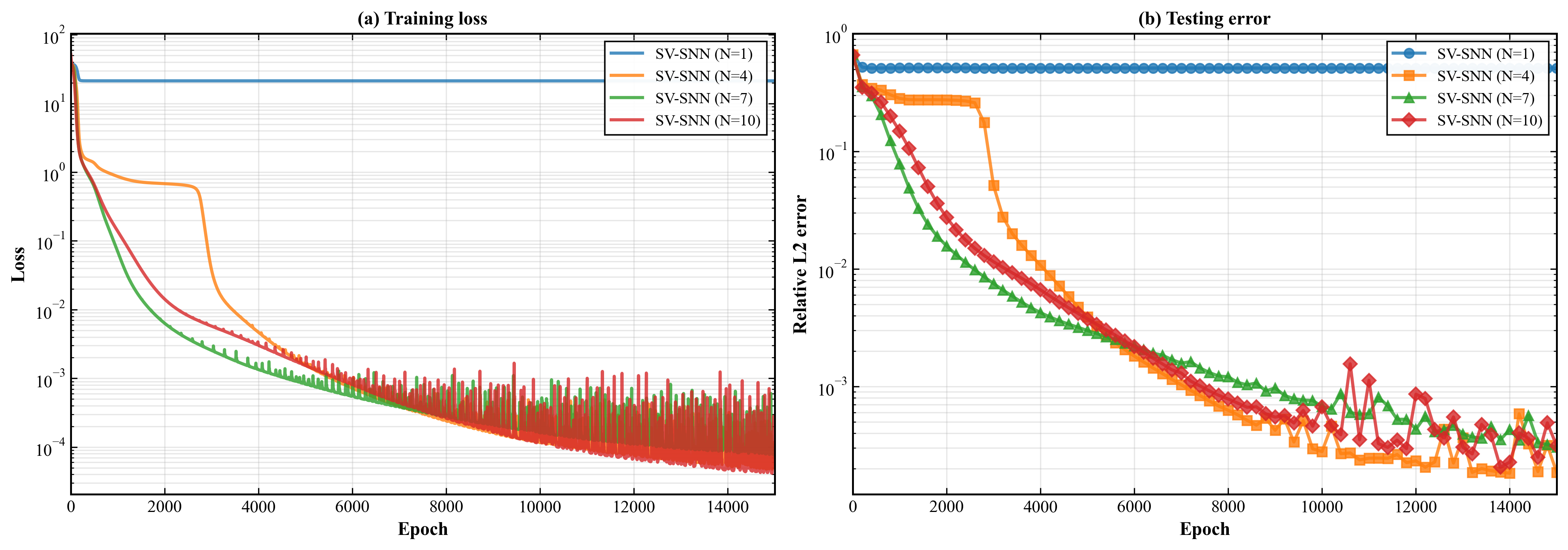}
			\caption{SV-SNN training dynamics under different network modes $N$: convergence curves of training loss and test error}
		\end{subfigure}
		\caption{Double-cylinder steady Navier-Stokes equations: Effects of network mode numbers on SV-SNN solution performance}
		\label{fig:modes_nsp0_analysis}
	\end{figure}

	\begin{table}[htbp]
		\centering
		\caption{Double-cylinder steady Navier-Stokes equations: SV-SNN prediction performance under different network mode numbers}
		\label{tab:modes_comparison_nsp0}
		\resizebox{0.8\textwidth}{!}{%
		\begin{tabular}{lccccc}
		\toprule
		\makecell{N} & \makecell{Total Parameters} & \makecell{Total Loss} & \makecell{ReL2E $u$} & \makecell{ReL2E $v$} & \makecell{Training Time (s)} \\
		\midrule
		1 & 101 & 2.13$\times$10$^{1}$ & 1.00$\times$10$^{0}$ & 5.39$\times$10$^{-1}$ & 212.21 \\
		4 & 404 & 5.06$\times$10$^{-5}$ & 3.61$\times$10$^{-4}$ & 1.96$\times$10$^{-4}$ & 597.54 \\
		7 & 707 & 7.72$\times$10$^{-5}$ & 4.19$\times$10$^{-4}$ & 4.94$\times$10$^{-4}$ & 989.10 \\
		10 & 1010 & 6.68$\times$10$^{-5}$ & 4.81$\times$10$^{-4}$ & 4.58$\times$10$^{-4}$ & 1377.02 \\
		\bottomrule
		\end{tabular}}
	\end{table}

	Figure~\ref{fig:modes_nsp0_analysis} and Table~\ref{tab:modes_comparison_nsp0} show significant effects of network mode numbers on SV-SNN's performance in solving double-cylinder steady Navier-Stokes equations. When mode number $N=1$, SV-SNN shows serious expressive capability insufficiency, with total loss reaching $2.13 \times 10^{1}$, $u$ component relative $L^2$ error as high as $1.00 \times 10^{0}$, and $v$ component error of $5.39 \times 10^{-1}$, indicating single mode cannot effectively represent complex flow field spatial structures. When mode number increases to $N=4$, network performance shows significant improvement, with total loss plummeting to $5.06 \times 10^{-5}$, approximately 6 orders of magnitude reduction compared to $N=1$, and $u$ and $v$ component relative errors decreasing to $3.61 \times 10^{-4}$ and $1.96 \times 10^{-4}$ respectively, indicating 4 modes can well capture main flow characteristics of the problem. Further increasing mode numbers to $N=7$ and $N=10$, solution accuracy does not significantly improve, with total losses of $7.72 \times 10^{-5}$ and $6.68 \times 10^{-5}$ respectively, relative errors maintaining same order of magnitude. From computational efficiency perspective, training time shows approximately linear growth with increasing mode numbers: from 212.21 seconds for $N=1$ to 1377.02 seconds for $N=10$, approximately 6.5 times increase, indicating trade-off relationships between accuracy and efficiency. For this test case, $N=4$ is optimal choice, ensuring relatively high solution accuracy while maintaining reasonable computational cost, demonstrating SV-SNN's good adaptability in mode number selection.

	\subsubsection{Analysis of Characteristic Frequency Effects}

	We define characteristic frequencies based on problem characteristics, then perform three-level frequency sampling based on characteristic frequencies. Since characteristic frequency-related sampling function is Gaussian normal distribution with mean as characteristic frequency and variance $\sigma^2$, characteristic frequency $w_{\text{char}}$ and variance $\sigma^2$ are two other key parameters of SV-SNN. First, we analyze effects of characteristic frequencies on SV-SNN solution performance, using two-dimensional Helmholtz equation ($\kappa=24\pi$) as example, taking different characteristic frequencies $w_{\text{char}} \in \{18\pi, 20\pi, 22\pi, 25\pi\}$, then using SV-SNN for solving and recording training time, relative $L^2$ error, and maximum absolute pointwise error.
	
	From results in Table~\ref{tab:frequency_analysis}, characteristic frequency selection has certain effects on SV-SNN solution performance. When characteristic frequency $w_{\text{char}} = 20\pi$, method achieves optimal performance with relative $L^2$ error decreasing to $2.56 \times 10^{-4}$ and maximum absolute pointwise error of $3.90 \times 10^{-4}$, indicating matching degree between characteristic frequency and problem intrinsic frequency is key factor affecting network performance. When characteristic frequency deviates from optimal value, such as $w_{\text{char}} = 18\pi$ and $w_{\text{char}} = 25\pi$, relative errors increase to $4.31 \times 10^{-4}$ and $4.21 \times 10^{-4}$ respectively. Notably, training time remains relatively stable across all tested characteristic frequency ranges, between 94.7-96.2 seconds, indicating characteristic frequency changes mainly affect convergence accuracy rather than convergence speed. This finding provides important guidance for characteristic frequency selection in practical applications: characteristic frequencies should be reasonably set according to problem physical characteristics, making them as close as possible to dominant frequency characteristics of target problems to achieve optimal solution performance.

	\begin{table}[htbp]
		\centering
		\caption{SV-SNN performance comparison under different characteristic frequencies (variance $\sigma^2 = 20^2$)}
		\label{tab:frequency_analysis}
		\resizebox{0.5\textwidth}{!}{%
		\begin{tabular}{lccc}
		\toprule
		\makecell{$w_{\text{char}}$} & \makecell{ReL2E} & \makecell{MAPE} & \makecell{Training Time (s)} \\
		\midrule
		$18\pi$ & 4.31$\times$10$^{-4}$ & 5.45$\times$10$^{-4}$ & 94.7 \\
		$20\pi$ & 2.56$\times$10$^{-4}$ & 3.90$\times$10$^{-4}$ & 95.3 \\
		$22\pi$ & 3.84$\times$10$^{-4}$ & 4.26$\times$10$^{-4}$ & 96.2 \\
		$25\pi$ & 4.21$\times$10$^{-4}$ & 4.73$\times$10$^{-4}$ & 95.8 \\
		\bottomrule
		\end{tabular}}
	\end{table}

	\begin{table}[htbp]
		\centering
		\caption{SV-SNN performance comparison under different variances (characteristic frequency $w_{\text{char}} = 15\pi$)}
		\label{tab:variance_analysis}
		\resizebox{0.5\textwidth}{!}{%
		\begin{tabular}{lccc}
		\toprule
		\makecell{$\sigma^2$} & \makecell{ReL2E} & \makecell{MAPE} & \makecell{Training Time (s)} \\
		\midrule
		$15^2$ & 6.89$\times$10$^{-4}$ & 7.31$\times$10$^{-4}$ & 96.4 \\
		$18^2$ & 3.68$\times$10$^{-4}$ & 4.05$\times$10$^{-4}$ & 95.1 \\
		$22^2$ & 5.47$\times$10$^{-4}$ & 5.78$\times$10$^{-4}$ & 96.7 \\
		$24^2$ & 7.73$\times$10$^{-4}$ & 7.12$\times$10$^{-4}$ & 96.8 \\
		\bottomrule
		\end{tabular}}
	\end{table}

	Table~\ref{tab:variance_analysis} further analyzes effects of variance parameter $\sigma^2$ on SV-SNN solution performance, showing variance parameter selection also has important effects on network performance. Under fixed characteristic frequency $w_{\text{char}} = 15\pi$, when variance $\sigma^2 = 18^2$, SV-SNN achieves optimal performance with relative $L^2$ error decreasing to $3.68 \times 10^{-4}$ and maximum absolute pointwise error of $4.05 \times 10^{-4}$, indicating appropriate variance settings enable adaptive Fourier features to effectively cover problem frequency spectrum ranges. When variance is too small ($\sigma^2 = 15^2$), frequency sampling range is limited, relative error increases to $6.89 \times 10^{-4}$; when variance is too large ($\sigma^2 = 22^2$ and $\sigma^2 = 24^2$), frequency distribution becomes too dispersed, leading to reduced key frequency components, with relative errors rising to $5.47 \times 10^{-4}$ and $7.73 \times 10^{-4}$ respectively. Similar to characteristic frequency analysis, training time under different variance settings remains stable, between 95.1-96.8 seconds, indicating variance parameters mainly affect convergence quality rather than training efficiency. This result indicates that in practical applications, variance parameters need reasonable balancing according to problem frequency characteristics, ensuring sufficient frequency coverage range while avoiding performance degradation caused by excessive dispersion, thus achieving optimal SV-SNN performance.

	\subsection{Comparison Experiments with Existing Methods}

	To fully demonstrate SV-SNN's advantages in solving high-frequency problems, we compare SV-SNN with multiple advanced physics-informed neural network methods.

	\begin{table}[h]
		\centering
		\caption{Comparison of SV-SNN with existing state-of-the-art methods for solving the two-dimensional 
		Helmholtz equation with $\kappa = 24\pi$. Training times for PINN-based methods are reported after 50,000 epochs, 
		while SV-SNN results are obtained after 5,000 epochs. Error metrics (AvgReL2E and StdReL2E) are obtained from 10 
		different runs with varied seeds.}
		\resizebox{0.8\textwidth}{!}{%
		\begin{tabular}{lccccc}
		\hline
		\textbf{Methods} & \textbf{XPINN} & \textbf{FBPINN} & \textbf{FourierPINN} & \textbf{BsPINN} & \textbf{SV-SNN} \\
		\hline
		Training time (s) & 7964.54 & 14256.50 & 9273.12 & 13126.60 & 132.54 \\
		AvgReL2E & $9.44 \times 10^{-1}$ & $6.79 \times 10^{-1}$ & $3.06 \times 10^{-1}$ & $1.74 \times 10^{-1}$ & $1.27 \times 10^{-2}$ \\
		StdReL2E & $1.62 \times 10^{-1}$ & $3.28 \times 10^{-1}$ & $8.17 \times 10^{-2}$ & $5.69 \times 10^{-2}$ & $2.05 \times 10^{-2}$ \\
		\hline
		\end{tabular}}
		\label{tab:pinn_comparison}
	\end{table}
	
\begin{figure}[htbp]
		\centering
		\includegraphics[width=1.0\textwidth]{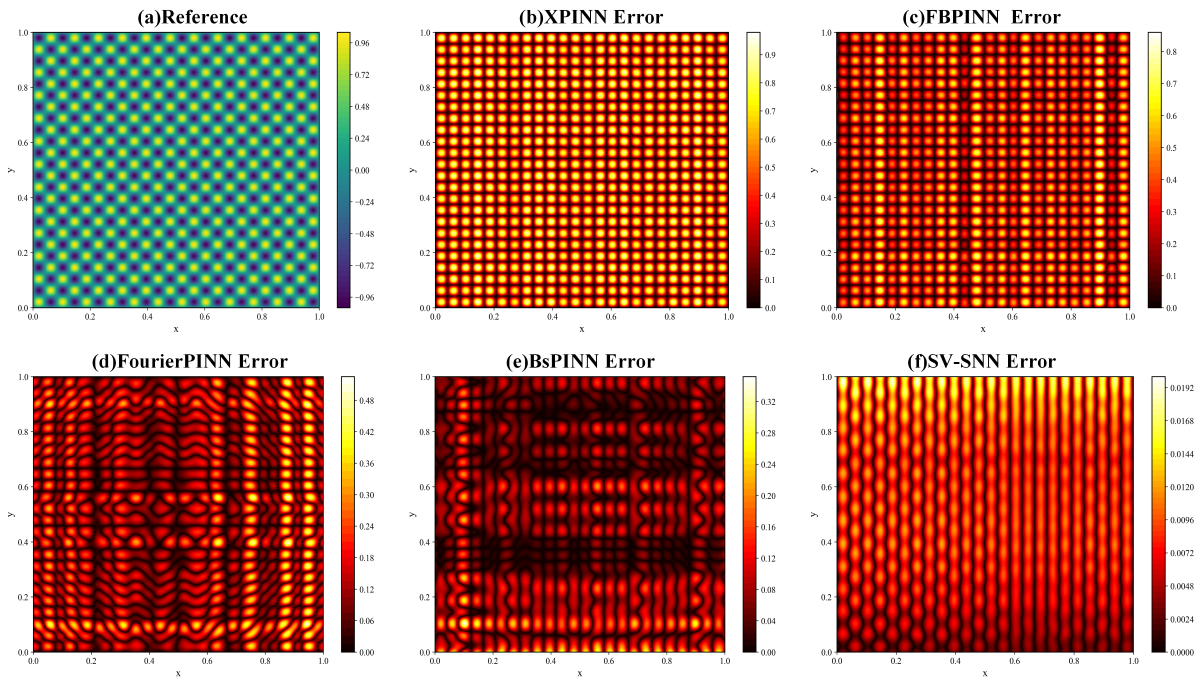}
		\caption{Performance comparison between SV-SNN and existing advanced methods for solving two-dimensional high-frequency Helmholtz equation ($\kappa = 24\pi$).}
		\label{fig:hel24pi_methods_compare}
	\end{figure}

	Table~\ref{tab:pinn_comparison} shows performance comparison results between SV-SNN and various advanced PINN methods when solving two-dimensional high-frequency Helmholtz equation ($\kappa = 24\pi$), with specific prediction effects shown in Figure~\ref{fig:hel24pi_methods_compare}. SV-SNN demonstrates significant advantages in both solution accuracy and computational efficiency: SV-SNN requires only 5,000 training epochs to achieve higher accuracy with training time of only 132.54 seconds, approximately 60 times faster than the fastest XPINN method; in solution accuracy, SV-SNN's average L2 relative error is $1.27 \times 10^{-2}$, more than one order of magnitude better than best-performing BsPINN, with standard deviation of only $2.05 \times 10^{-2}$, indicating good method stability. These results fully validate separated-variable spectral neural networks' excellent performance when handling high-frequency oscillation problems, providing efficient and reliable new approaches for neural network solving of high-frequency partial differential equations.

	\begin{table}[h]
		\centering
		\caption{Comparison of SV-SNN with existing state-of-the-art methods for solving the two-dimensional Helmholtz 
		equation with $\kappa = 48\pi$. Training times for PINN-based methods are reported after 50,000 epochs, 
		while SV-SNN results are obtained after 5,000 epochs. Error metrics (AvgReL2E and StdReL2E) are obtained 
		from 10 different runs with varied seeds.}
		\resizebox{0.8\textwidth}{!}{%
		\begin{tabular}{lccccc}
		\hline
		\textbf{Methods} & \textbf{XPINN} & \textbf{FBPINN} & \textbf{FourierPINN} & \textbf{BsPINN} & \textbf{SV-SNN} \\
		\hline
		Training time (s) & 8742.18 & 15834.72 & 10156.89 & 14389.34 & 148.92 \\
		AvgReL2E & $1.85 \times 10^{0}$ & $1.34 \times 10^{0}$ & $7.23 \times 10^{-1}$ & $4.18 \times 10^{-1}$ & $3.94 \times 10^{-2}$ \\
		StdReL2E & $3.47 \times 10^{-1}$ & $5.92 \times 10^{-1}$ & $1.89 \times 10^{-1}$ & $1.26 \times 10^{-1}$ & $4.73 \times 10^{-2}$ \\
		\hline
		\end{tabular}}
		\label{tab:pinn_comparison_48pi}
	\end{table}

	When wavenumber further increases to $\kappa = 48\pi$, problem difficulty significantly increases with more intense high-frequency oscillation characteristics. Table~\ref{tab:pinn_comparison_48pi} shows performance of various methods under these more challenging conditions. Results indicate that as frequency increases, traditional PINN methods' performance further deteriorates: XPINN and FBPINN average L2 relative errors reach $1.85$ and $1.34$ respectively, while SV-SNN still maintains high accuracy of $3.94 \times 10^{-2}$, still one order of magnitude better than best-performing BsPINN ($4.18 \times 10^{-1}$). In computational efficiency, SV-SNN training time only increases to 148.92 seconds, while other methods show different degrees of training time increases. These results further validate SV-SNN's robustness and superiority when handling high-frequency problems, indicating the method can still maintain stable high-performance when facing more complex high-frequency oscillation challenges.

	\section{Conclusions}
	
	In this paper, we introduce Separated-Variable Spectral Neural Networks, a novel class of physics-informed neural networks specifically designed for high-frequency partial differential equations. This method employs conceptual frameworks analogous to separation of variables and spectral methods in classical numerical approaches, while incorporating the mesh-free characteristics and flexibility of neural networks. Through singular value decomposition of Jacobian matrices, we introduce the concept of effective rank for analyzing parameter space dimensions that play crucial roles during gradient descent training, and leverage singular value distribution curves to analyze spectral bias phenomena. Our effective rank analysis framework provides valuable guidance for designing new efficient physics-informed neural network architectures.
	
	We believe SV-SNN opens new possibilities for solving high-frequency PDEs using neural networks. We advocate for deeper integration of fundamental concepts and important principles from numerical methods with neural networks to design efficient and high-precision neural PDE solvers. Taking orthogonal basis functions as an example, spectral methods encompass multiple types of orthogonal basis functions (such as Fourier bases, Chebyshev bases, and Legendre bases). These basis functions naturally possess orthogonal characteristics, making them well-suited for constructing full-rank neural network modules, thereby avoiding neural network parameter space collapse caused by low-rank issues.
	
	%Bibliography
	\bibliographystyle{unsrt} 
	 
	\bibliography{references}  
	%\appendix
	
\begin{algorithm}[H]
	\caption{Separated-Variable Spectral Neural Network (SV-SNN) Training Algorithm}
	\label{alg:sv_snn_training}
	\begin{algorithmic}[1]
	\Require Partial differential equation $\mathcal{F}[u] = 0$, initial condition $\mathcal{I}[u](\bm{x},0) = g_0(\bm{x})$, boundary condition $\mathcal{B}[u] = g_B(\bm{x},t)$
	\Require Spatial domain $\Omega \subset \mathbb{R}^d$, temporal domain $[0,T]$, network modes $N$, frequency numbers $K$, trainable parameter set $\Theta$
	\Ensure Trained network parameters $\Theta^*$
	\State \textbf{Initialization:}
	\State Design three-level sampling strategy based on characteristic frequency $\omega_{\text{char}}$:
	\begin{align*}
	\omega_{n,k}^{(j)} &= \begin{cases}
	\text{Linear distribution}[\omega_{\min}, \omega_{\text{char}}], & k \in [1, K/4] \\
	\text{Gaussian distribution}\mathcal{N}(\omega_{\text{char}}, \sigma^2), & k \in [K/4+1, 3K/4] \\
	\text{Uniform distribution}[\omega_{\text{char}}, \omega_{\max}], & k \in [3K/4+1, K]
	\end{cases}
	\end{align*}
	\State Initialize spatial Fourier spectral feature network parameters $\{a_{n,k}^{(j)}, b_{n,k}^{(j)},w_{n,k}^{(j)},\beta_n^{(j)}\}$, mode coefficients $\{c_n\}$, temporal network parameters $\{\Theta_n^{(t)}\}$
	\State Generate training data: initial condition points $\{(\bm{x}_{\text{IC}}^i, 0)\}_{i=1}^{N_{\text{IC}}}$, boundary condition points $\{(\bm{x}_{\text{BC}}^j, t_{\text{BC}}^j)\}_{j=1}^{N_{\text{BC}}}$, PDE collocation points $\{(\bm{x}_{\text{PDE}}^k, t_{\text{PDE}}^k)\}_{k=1}^{N_{\text{PDE}}}$
	\State \textbf{Network construction:}
	\For{$n = 1$ to $N$}
		\State Construct spatial Fourier features: $\Phi_n^{(j)}(x_j) = \sum_{k=1}^{K} [a_{n,k}^{(j)} \sin(w_{n,k}^{(j)} x_j) + b_{n,k}^{(j)} \cos(w_{n,k}^{(j)} x_j)] + \beta_n^{(j)}$
		\State Construct temporal neural network: $T_n(t) = f_n^{(NN)}(t; \Theta_n^{(t)})$
	\EndFor
	\State Spatiotemporal spectral neural network solution: $u^{\Theta}(\bm{x}, t) = \sum_{n=1}^{N} c_n \prod_{j=1}^{d} \Phi_n^{(j)}(x_j) T_n(t)$
	\State \textbf{Training loop:}
	\For{epoch $= 1$ to $E$}
		\State \textbf{Loss function computation:}
		\State Initial condition loss: $\mathcal{L}_{\text{IC}}(\Theta) = \frac{1}{N_{\text{IC}}} \sum_{i=1}^{N_{\text{IC}}} |\mathcal{I}[u^{\Theta}](\bm{x}_{\text{IC}}^i, 0) - g_0(\bm{x}_{\text{IC}}^i)|^2$
		\State Boundary condition loss: $\mathcal{L}_{\text{BC}}(\Theta) = \frac{1}{N_{\text{BC}}} \sum_{j=1}^{N_{\text{BC}}} |\mathcal{B}[u^{\Theta}](\bm{x}_{\text{BC}}^j, t_{\text{BC}}^j) - g_B(\bm{x}_{\text{BC}}^j, t_{\text{BC}}^j)|^2$
		\State \textbf{PDE residual computation (hybrid differentiation):}
		\State For all PDE collocation points $\{(\bm{x}_{\text{PDE}}^k, t_{\text{PDE}}^k)\}_{k=1}^{N_{\text{PDE}}}$ parallel computation:
		\State Analytical spatial derivatives: $\frac{\partial u^{\Theta}}{\partial x_l} = \sum_{n=1}^{N} c_n \prod_{m \neq l} \Phi_n^{(m)}(x_m) \frac{\partial \Phi_n^{(l)}}{\partial x_l}(x_l) T_n(t)$
		\State Automatic temporal derivatives: $\frac{\partial u^{\Theta}}{\partial t} = \sum_{n=1}^{N} c_n \prod_{j=1}^{d} \Phi_n^{(j)}(x_j) \frac{\partial T_n}{\partial t}(t)$
		\State Compute PDE residual vector for all collocation points: $\boldsymbol{r} = [\mathcal{F}[u^{\Theta}](\bm{x}_{\text{PDE}}^1, t_{\text{PDE}}^1), \ldots, \mathcal{F}[u^{\Theta}](\bm{x}_{\text{PDE}}^{N_{\text{PDE}}}, t_{\text{PDE}}^{N_{\text{PDE}}})]^T$
		\State PDE loss: $\mathcal{L}_{\text{PDE}}(\Theta) = \frac{1}{N_{\text{PDE}}} \sum_{k=1}^{N_{\text{PDE}}} |\mathcal{F}[u^{\Theta}](\bm{x}_{\text{PDE}}^k, t_{\text{PDE}}^k)|^2$
		\State Total loss: $\mathcal{L}(\Theta) = \lambda_{\text{IC}} \mathcal{L}_{\text{IC}}(\Theta) + \lambda_{\text{BC}} \mathcal{L}_{\text{BC}}(\Theta) + \lambda_{\text{PDE}} \mathcal{L}_{\text{PDE}}(\Theta)$
		\State \textbf{Parameter update:} Use Adam optimizer to update all parameters $\Theta$
	\EndFor
	\Return Optimized network parameters $\Theta^*$
	\end{algorithmic}
\end{algorithm}
	
\end{document}